\documentclass[letterpaper, 10 pt, conference]{ieeeconf}  

\IEEEoverridecommandlockouts                              

\usepackage{multicol}

\usepackage{float}
\usepackage{siunitx}
\usepackage{amsmath}
\usepackage{amssymb}
\usepackage{amsfonts}
\usepackage{breqn}
\usepackage{xcolor}

\usepackage{caption}
\usepackage{subcaption}

\usepackage[style=ieee]{biblatex}
\usepackage{booktabs}

\usepackage{nicefrac}

\usepackage{algpseudocode}
\usepackage[ruled,vlined]{algorithm2e}
\usepackage{graphicx}
\usepackage{subfiles}
\usepackage{wrapfig}

\usepackage[english]{babel}
\usepackage{hyperref}
\addto\captionsenglish{%
}
\addto\extrasenglish{%
}

\definecolor{darkbrown}{rgb}{0.4, 0.26, 0.13}

\newcommand{\rev}[1]{{#1}}
\newcommand{\revtwo}[1]{{#1}}

\newcommand{\bmat}[1]{\begin{bmatrix} #1 \end{bmatrix}}

\newcommand{\paramdim}[0]{M}
\newcommand{\params}[0]{\theta}

\newcommand{\numparticles}[0]{N}

\newcommand{\statevec}[0]{\mathbf{s}}

\newcommand{\observationdim}[0]{N}
\newcommand{\observationvec}[0]{\mathbf{x}}

\newcommand{\trajectory}[0]{\mathcal{X}}
\newcommand{\trajectoryset}[0]{D_\trajectory}

\newcommand{\simu}[0]{^{\text{sim}}}
\newcommand{\real}[0]{^{\text{real}}}
\newcommand{\fsim}{f_{\text{sim}}}
\newcommand{\fobs}{f_{\text{obs}}}
\newcommand{\fstep}{f_{\text{step}}}
\newcommand{\pobs}{p_{\text{obs}}}

\newcommand{\exparams}{\bar{\params}}

\newcommand{\jointpos}{\mathbf{q}}
\newcommand{\jointvel}{\mathbf{\dot{q}}}
\newcommand{\jointacc}{\mathbf{\ddot{q}}}
\newcommand{\jointtorque}{\tau}
\newcommand{\externalforce}{\mathbf{f}^{\text{ext}}}
\newcommand{\timestep}{\Delta t}

\newcommand{\pdef}[0]{p_{\text{def}}}

\newcommand{\plim}[0]{p_{\text{lim}}}

\title{\LARGE \bf
Probabilistic Inference of Simulation Parameters \\via Parallel Differentiable Simulation
}

\author{
Eric Heiden${}^{1}$, Christopher E. Denniston${}^{1}$, David Millard${}^{1}$, Fabio Ramos$^{2}$, Gaurav S. Sukhatme$^{1,3}$
\thanks{$^{1}$Department of Computer Science, University of Southern California, Los Angeles, USA
        {\tt\small \{heiden, cdennist, dmillard, gaurav\}@usc.edu}}%
\thanks{$^{2}$NVIDIA, Seattle, USA
        {\tt\small ftozetoramos@nvidia.com}}%
\thanks{$^{3}$G.S. Sukhatme holds concurrent appointments as a Professor at USC and as an Amazon Scholar. This paper describes work performed at USC and is not associated with Amazon.}%
\thanks{This work was supported by a Google PhD Fellowship and a NASA Space Technology Research Fellowship, grant number 80NSSC19K1182.}
}

\begin{document}

\setlength{\textfloatsep}{0.5em}

\maketitle
\thispagestyle{empty}
\pagestyle{empty}

\begin{abstract}
Reproducing real world dynamics in simulation is critical for the development of new control and perception methods. This task typically involves the estimation of simulation parameter distributions from observed rollouts through an inverse inference problem characterized by multi-modality and skewed distributions.   
We address this challenging problem through a novel Bayesian inference approach that approximates a posterior distribution over simulation parameters given real sensor measurements. By extending the commonly used Gaussian likelihood model for trajectories via the multiple-shooting formulation, our gradient-based particle inference algorithm, Stein Variational Gradient Descent, is able to identify highly nonlinear, underactuated systems. We leverage GPU code generation and differentiable simulation to evaluate the likelihood and its gradient for many particles in parallel.
Our algorithm infers nonparametric distributions over simulation parameters more accurately than comparable baselines and handles constraints over parameters efficiently through gradient-based optimization. We evaluate estimation performance on several physical experiments. On an underactuated mechanism where a 7-DOF robot arm excites an object with an unknown mass configuration, we demonstrate how the inference technique can identify symmetries between the parameters and provide highly accurate predictions.\\
Website: {\footnotesize \url{https://uscresl.github.io/prob-diff-sim}}
\end{abstract}

\begin{refsection}

\begin{figure}[h!]
    \centering
    \newcommand{\figheight}{3.5cm}
    \hspace*{-1em}
    \includegraphics[height=\figheight]{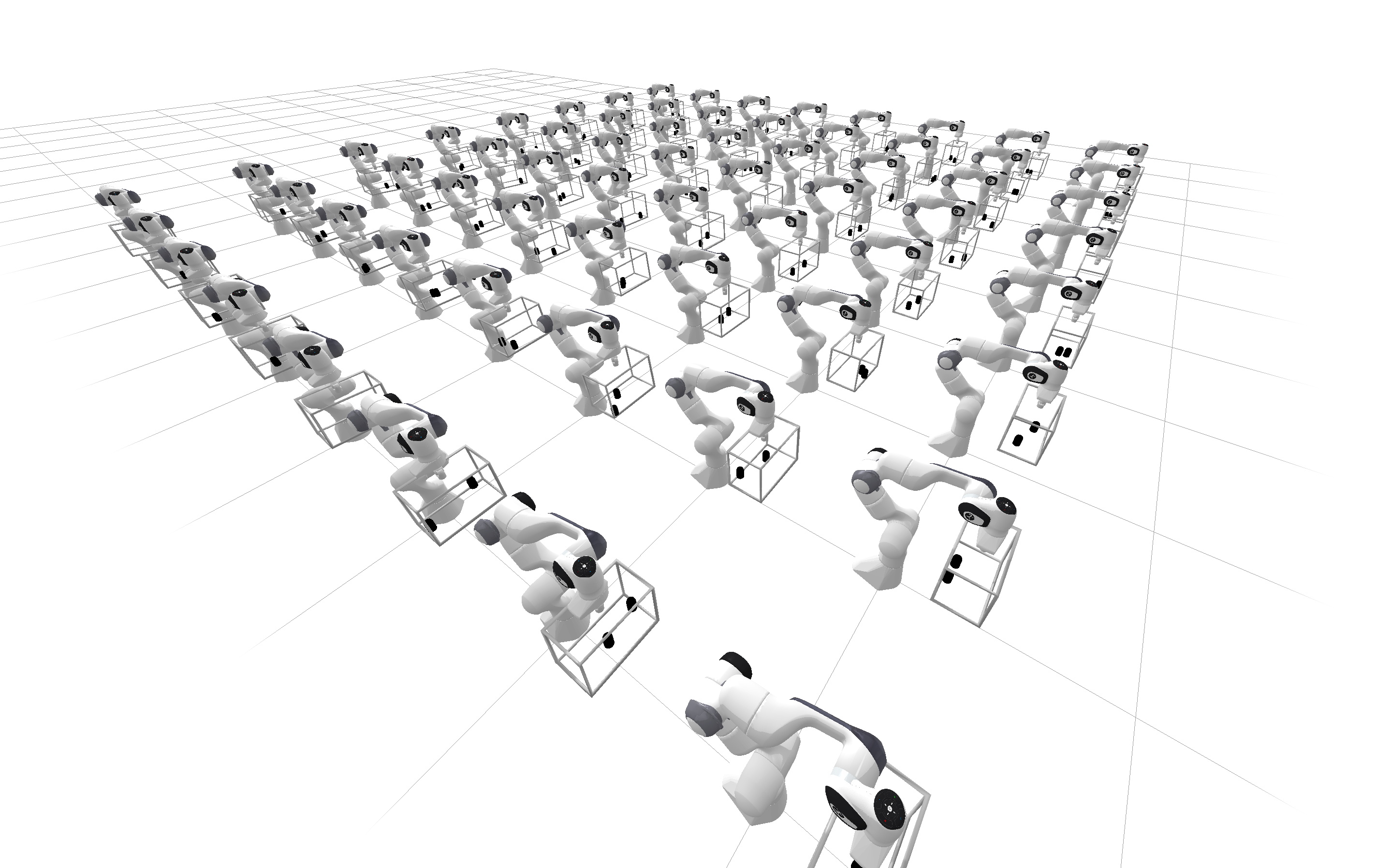}
    \includegraphics[height=\figheight]{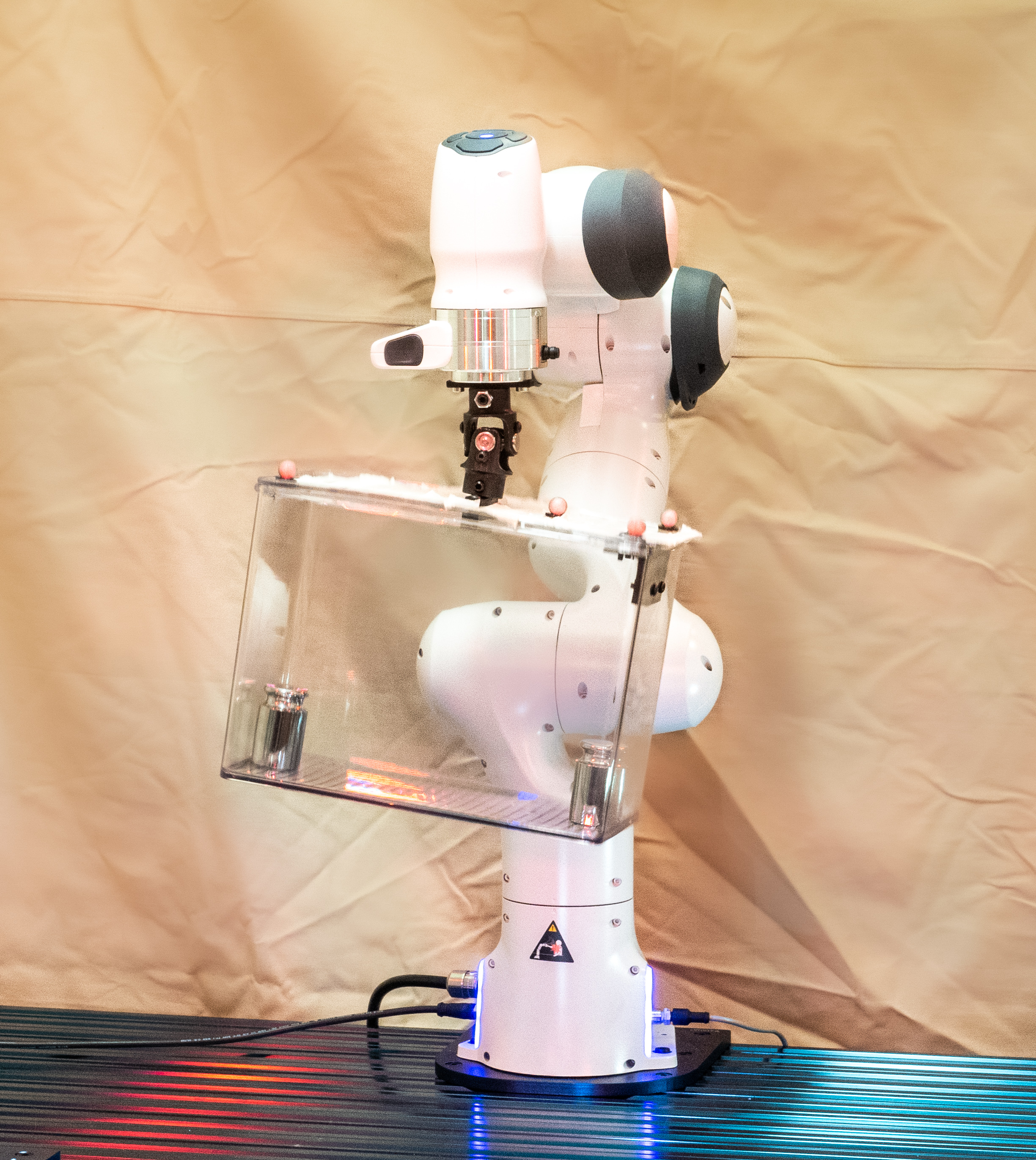}
    \includegraphics[height=\figheight]{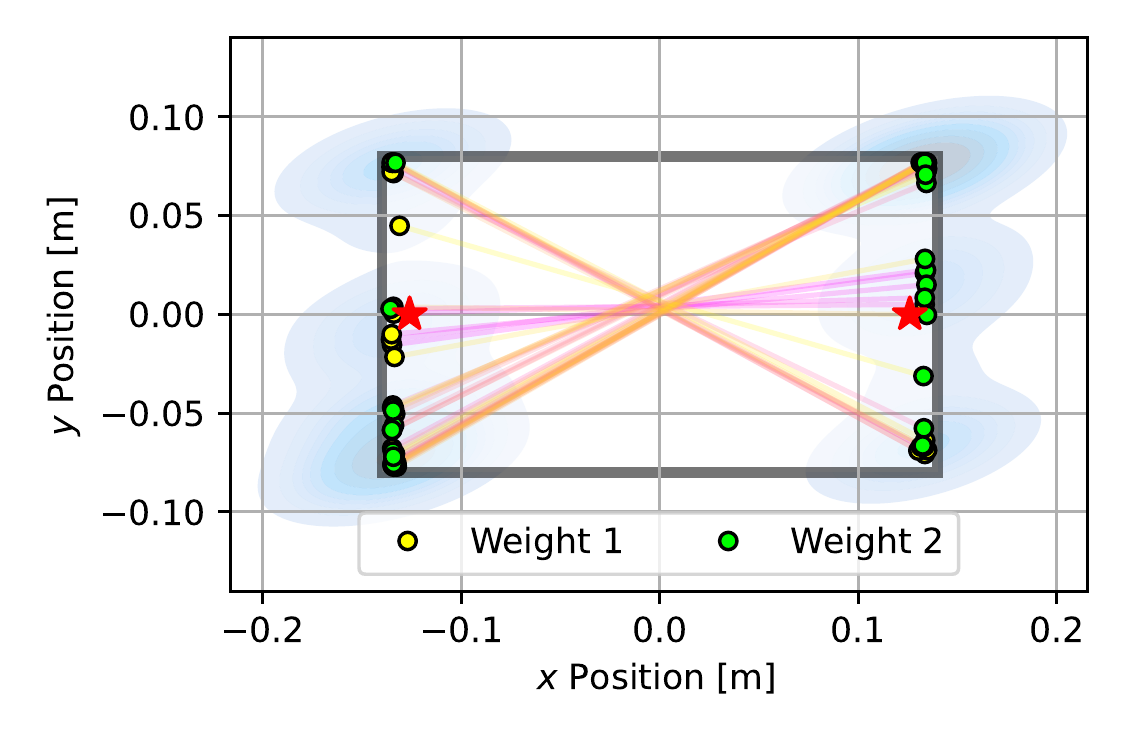}
    \vspace*{-0.5em}
    \caption{Panda robot arm shaking a box with two weights in it at random locations in our parallel differentiable simulator (left), physical robot experiment (center), and the inferred particle distribution using our proposed method over the 2D positions of the two weights inside the box (right).\vspace*{0.5em}}
    \label{fig:panda-sim2real}
\end{figure}

\section{Introduction}
\label{sec:intro}
Simulators for robotic systems allow for rapid prototyping and development of algorithms and systems~\cite{koenig_design_2004}, as well as the ability to quickly and cheaply generate training data for reinforcement learning agents and other control algorithms~\cite{andrychowicz_learning_2020}.
In order for these models to be useful, the simulator must accurately predict the outcomes of real-world interactions.
This is accomplished through both accurately modeling the dynamics of the environment as well as correctly identifying the parameters of such models. In this work, we focus on the latter problem of parameter inference.

Optimization-based approaches have been applied in the past to find simulation parameters that best match the measured trajectories~\cite{chebotar_closing_2019, ramos_bayessim_2019}.
However, in many systems we encounter in the real world, the dynamics are highly nonlinear, resulting in optimization landscapes fraught with poor local optima where such algorithms get stuck.
Global optimization approaches, such as population-based methods, have been applied~\cite{heiden2021neuralsim} but are sample inefficient and cannot quantify uncertainty over the predicted parameters.

In this work we follow a probabilistic inference approach and estimate belief distributions over the the most likely simulation parameters given the noisy trajectories of observations from the real system. The relationship between the trajectories and the underlying simulation parameters can be highly nonlinear, hampering commonly used inference algorithms. To tackle this, we introduce a multiple-shooting formulation to the parameter estimation process which drastically improves convergence to high-quality solutions.
Leveraging GPU-based parallelism of a differentiable simulator allows us to efficiently compute likelihoods and evaluate its gradients over many particles simultaneously. Based on Stein Variational Gradient Descent (SVGD), our gradient-based nonparametric inference method allows us to optimize parameters while respecting constraints on parameter limits and continuity between shooting windows. Through various experiments we demonstrate the improved accuracy and convergence of our approach.


Our contributions are as follows: first, we \revtwo{reformulate the commonly used Gaussian likelihood function through the multiple-shooting strategy to allow for the tractable estimation of simulation parameters from long noisy trajectories.}
Second, we propose a constrained optimization algorithm for nonparametric variational inference with constraints on parameter limits and shooting defects.
Third, we leverage a fully differentiable simulator and GPU parallelism to automatically calculate gradients for many particles in parallel.
Finally, we validate our system on a simulation parameter estimation problem from real-world data and show that our calculated posteriors are more accurate than \rev{comparable algorithms, as well as likelihood-free methods}.

\section{Related Work}


System identification methods for robotics use a dynamics model with often linearly dependent parameters in classical time or frequency domain~\cite{vandanjon1995spectrum}, and solve for these parameters via least-squares methods~\cite{kozlowski_modelling_1998}. 
Such estimation approaches have been applied, for example, to the identification of inertial parameters of robot arms~\cite{khosla1985sysid,atkeson1986inertial,caccavale_identification_1994,vandanjon1995spectrum} with time-dependent gear friction~\cite{grotjahn_friction_2001}, or parameters of contact models~\cite{verscheure2010contact,fazeli2018identifiability}. 
Parameter estimation has been studied to determine a minimum set of identifiable inertial parameters~\cite{gautier1990minset} and finding exciting trajectories which maximize identifiability~\cite{antonelli_systematic_1999,gautier_exciting_1991}. 
More recently, least-squares approaches have been applied to estimate parameters of nonlinear models, such as the constitutive equations of material models~\cite{mahnken2017identification,hahn2019real2sim,narang2020tactile}, and contact models~\cite{kolev2015sysid,lelidec2021diffsim}.
In this work, we do not assume a particular type of system to identify, but propose a method for general-purpose differentiable simulators that may combine multiple models whose parameters can influence the dynamics in highly nonlinear ways.

Bayesian methods seek to infer probability distributions over simulation parameters, and have been applied to infer the parameters of dynamical systems in robotic tasks~\cite{wang_markov_2011,Muratore2020BayesianDR,Tan2018SimtoRealLA} and complex dynamical systems~\cite{ninness_bayesian_2010}. Our approach is a Bayesian inference algorithm which allows us to include priors to find posterior distributions over simulation parameters.
The advantages \revtwo{of} Bayesian inference approaches have been shown to be useful in the areas of uncertainty quantification~\cite{ninness_bayesian_2010,eykhoff_chapter_1981}, system noise quantification~\cite{ting_bayesian_2011} and \revtwo{model parameter inference~\cite{qian2003mc,cranmer2020sbi}}.

Our method is designed for differentiable simulators which have been developed recently for various areas of modeling, such as articulated rigid body dynamics~\cite{peres2018lcp,hu2020difftaichi,geilinger2020add,Qiao2020Scalable,heiden2021neuralsim,lelidec2021diffsim}, deformables~\cite{hu2019chainqueen,hu2020difftaichi,murthy2021gradsim,heiden2021disect,Huang2021PlasticineLabAS} and cloth simulation~\cite{liang2019diffcloth,hu2020difftaichi,Qiao2020Scalable}, as well as sensor simulation~\cite{NimierDavidVicini2019Mitsuba2,heiden2020lidar}. Certain physical simulations (e.g. fracture mechanics) may not be analytically differentiable, so that surrogate gradients may be necessary~\cite{han2018stein}.




Without assuming access to the system equations, likelihood-free inference approaches, such as approximate Bayesian computation (ABC), have been applied to the inference of complex phenomena~\cite{toni2008abc,papamakarios2016epsilon,ramos_bayessim_2019,hsu2019likelihoodfree,matl_inferring_2020,matl_stressd_2020}. While such approaches do not rely on a simulator to compute the likelihood, our experiments in \autoref{sec:likelihoodfree} indicate that the approximated posteriors inferred by likelihood-free methods are less accurate \revtwo{for high-dimensional parameter distributions} while requiring significantly more simulation roll-outs as training data.

Domain adaptation techniques have been proposed that close the loop between parameter estimation from real observation and improving policies learned from simulators~\cite{chebotar_closing_2019, ramos_bayessim_2019, mehta2020adaptiveda, du_auto-tuned_2021}.
Achieving an accurate simulation is typically not the final objective of these methods. Instead, the performance of the learned policy derived from the calibrated simulator is evaluated which does not automatically imply that the simulation is accurate~\revtwo{\cite{lambert2020mismatch}}. 
In this work, we focus solely on the calibration of the simulator where its parameters need to be inferred.
\vspace*{-0.5em}
\section{Formulation}
\label{sec:formulation}
\vspace*{-0.5em}

In this work we address the parameter estimation problem via the Bayesian inference methodology.
The posterior $p(\params|\trajectoryset)$ over simulation parameters $\params\in\mathbb{R}^\paramdim$ and a set of \rev{trajectories $\trajectoryset$} is calculated using Bayes' rule:
\vspace*{-0.5em}
\begin{align*}
    p(\params|\trajectoryset) \propto p(\trajectoryset|\params)p(\params),\\[-2em]
\end{align*}
where $p(\trajectoryset|\params)$ is the likelihood distribution and $p(\params)$ is the prior distribution over the simulation parameters.
We aim to approximate the distribution over true parameters $p(\params\real)$, which in general may be intractable to compute.
These true parameters $\params\real$ generate a set of trajectories $\trajectoryset\real$ which may contain some observation noise.
We assume that these parameters are unchanging during the trajectory and represent physical parameters, such as friction coefficients or link masses.

We assume that each trajectory is a Hidden Markov Model (HMM)~\cite{russell_artificial_2009} which has some known initial state, $\statevec_0$, and some hidden states $\statevec_t$, $t \in [1..T]$.
These hidden states induce an observation model $p_{\text{obs}}(\observationvec_t | \statevec_t)$.
In \rev{the} simulator, we map simulation states to observations via a \rev{deterministic} observation function $\fobs: \statevec \mapsto \observationvec$.

The transition probability $p(\statevec_{t} | \statevec_{t-1}, \params)$ of the HMM cannot be directly observed but, in the case of a physics engine, can be approximated by sampling from a distribution of simulation parameters and applying a deterministic simulation step function $\fstep: (\statevec, t, \params) \mapsto \statevec$. In \autoref{sec:dynamics}, we describe our implementation of $\fstep$, the discrete dynamics simulation function.
The function $\fsim$ rolls out multiple steps via $\fstep$ to produce a trajectory of $T$ states given a parameter vector $\params$ and an initial state $\statevec_0$: $\fsim(\params, \statevec_0) = [\statevec]_{t=1}^{T}$. To compute measurements from such state vectors, we use the observation function $\fobs$: $\trajectory = \fobs([\statevec]_{t=1}^{T})$.
Finally, we obtain a set of \rev{simulated} trajectories $\trajectoryset\simu = [\fobs (\fsim(\params, \statevec_0\real))]$ for each initial state $\statevec_0\real$ from the trajectories in $\trajectoryset\real$.
The initial state $\statevec_0\real$ from \revtwo{an} observed trajectory may be acquired via state estimation techniques, \revtwo{e.g.} methods that use inverse kinematics to infer joint positions from tracking measurements.



We aim to minimize the Kullback-Leibler (KL) divergence between the trajectories generated from forward simulating our learned parameter particles and the ground-truth trajectories
, while taking into account the priors over simulation parameters:
\begin{align*}
d_{\text{KL}} \left[p(\trajectoryset\simu | \params\simu)p(\params\simu)  \parallel  p(\trajectoryset\real | \params\real)p(\params\real)\right].
\end{align*}
We choose the \emph{exclusive} KL divergence instead of the opposite direction since it was shown for our choice of particle-based inference algorithm in \cite{liu2017svgdgradientflow} that the particles exactly approximate the target measure when an infinitely dimensional functional optimization process minimizes this form of KL divergence.



\vspace{-0.5em}
\section{Approach}
\label{sec:approach}



\subsection{Stein Variational Gradient Descent}
\label{sec:svgd}
A common challenge in statistics and machine learning is the approximation of intractable posterior distributions. In the domain of robotics simulators, the inverse problem of inferring high-dimensional simulation parameters from trajectory observations is often nonlinear and non-unique as there are potentially a number of parameters values that equally well produce simulated roll-outs similar to the real dynamical behavior of the system. This often results in non-Gaussian, multi-modal parameter estimation problems.
Markov Chain Monte-Carlo (MCMC) methods are known to be able to find the true distribution, but require an enormous amount of samples to converge, which is exacerbated in high-dimensional parameter spaces.
Variational inference approaches, on the other hand, approximate the target distribution by a simpler, tractable distribution, which often does not capture the full posterior over parameters accurately~\cite{blei_variational_2017}.

We present a solution based on the Stein Variational Gradient Descent (SVGD) algorithm~\cite{liu_stein_2019} that approximates the posterior distribution $p(\params | \trajectoryset) = \frac{p(\trajectoryset|\params)p(\params)}{\int p(\trajectoryset|\params)p(\params) d \params}$ by a set of particles $q(\params | \trajectoryset) = \frac {1}{N} \sum_{i=1}^N \delta(\params_i - \params)$ where $\delta(\cdot)$ is the Dirac delta function, and makes use of differentiable likelihood and prior functions to be efficient. SVGD avoids the computation of the intractable marginal likelihood in the denominator by only requiring the computation of $\nabla_\params \log p(\params | \trajectoryset) = \frac{\nabla_\params p(\params | \trajectoryset)} {p(\params | \trajectoryset)}$ which is independent of the normalization constant.
The particles are adjusted according to the steepest descent direction to reduce the KL divergence in a reproducing kernel Hilbert space (RKHS) between the current set of particles representing $q(\params | \trajectoryset)$ and the target $p(\params | \trajectoryset)$.

As derived in~\cite{liu_stein_2019}, the set of particles $\{\params_i\}_{i=1}^\numparticles$ is updated by the following function:
\begin{align}\label{eq:svgd_update}
    \params_i &\gets \params_i + \epsilon \phi(\params_i), \\ \nonumber
    \phi(\cdot) &= \frac{1}{\numparticles} \sum_{j=1}^\numparticles \left[k(\params_j,\params)\nabla_{\params_j}\!\log  p(\trajectoryset|\params_j)p(\params_j)
    \!+\! \nabla_{\params_j}k(\params_j,\params)\right]\!,
\end{align}
where $k(\cdot,\cdot): \mathbb{R}^M\times\mathbb{R}^M\to\mathbb{R}$ is a positive definite kernel and $\epsilon$ is the step size.
In this work, we use the radial basis function kernel, \rev{which is a common choice for SVGD~\cite{liu_stein_2019} due to its smoothness and infinite differentiability}. To tune the kernel bandwidth, we adopt the median heuristic, \rev{which has been shown to provide robust performance on large learning problems~\cite{garreau2018median}}.

SVGD requires that the likelihood function be differentiable in order to calculate \autoref{eq:svgd_update}, which in turn requires that $\fsim$ and $\fobs$ be differentiable.
To enable this, we use a fully-differentiable simulator, observation function and likelihood function, and leverage automatic differentiation with code generation to generate CUDA kernel code~\cite{sanders_cuda_2010}.
Because of this CUDA code generation, we are able to calculate $\nabla_{\params_j}\log  p(\trajectoryset|\params)p(\params)$ for each particle in parallel on the GPU.

\vspace{-0.5em}
\subsection{Likelihood Model for Trajectories}
\begin{figure}
    \centering
    \newcommand{\figheight}{3.4cm}
    \begin{subfigure}[b]{0.48\columnwidth}
        \centering
        \includegraphics[height=\figheight,trim=1cm 0 0 0]{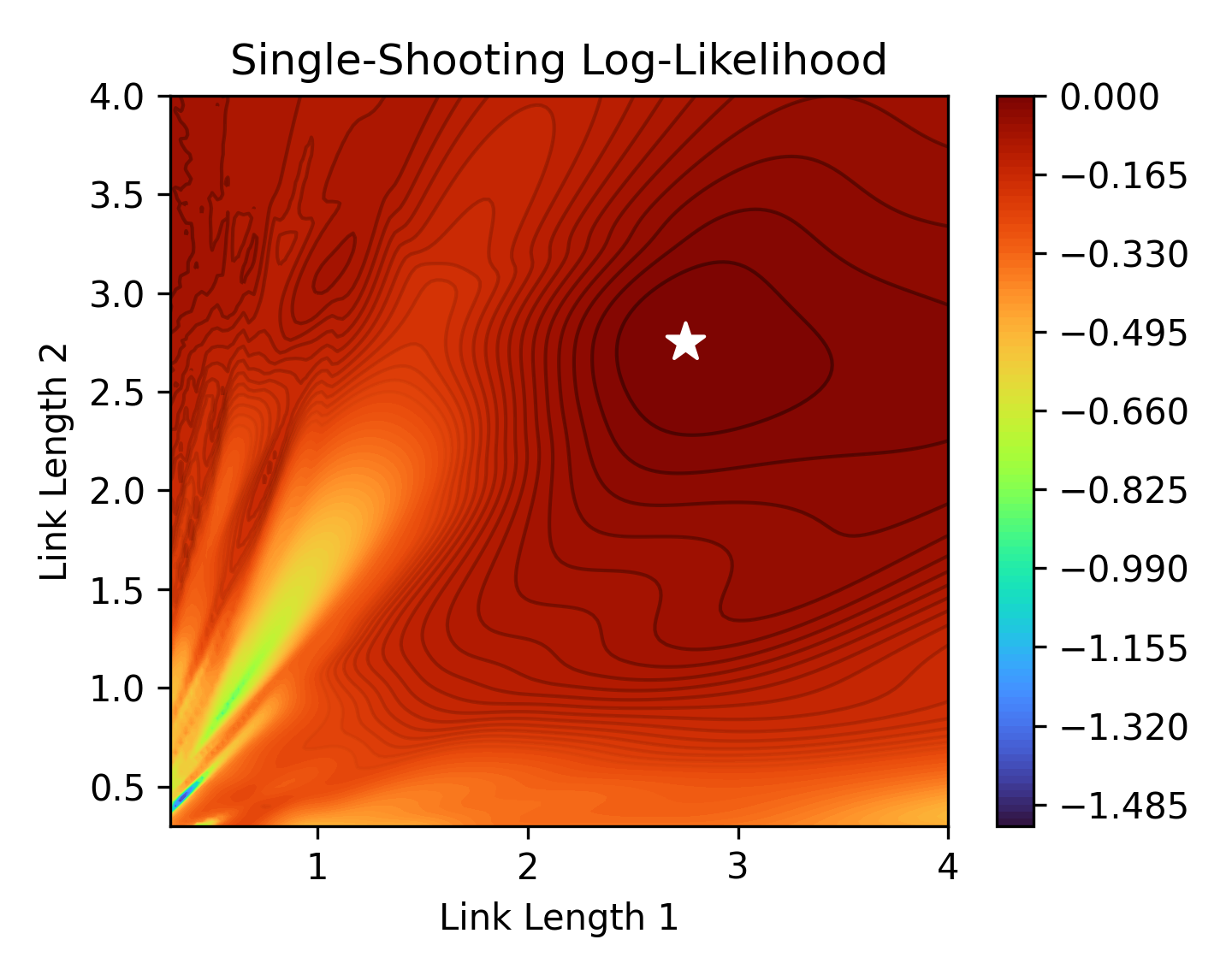}
        \caption{Single Shooting}
        \label{fig:double-pendulum-ll-ss}
    \end{subfigure}
    \hfill
    \begin{subfigure}[b]{0.48\columnwidth}
        \centering
        \includegraphics[height=\figheight,trim=1cm 0 0 0]{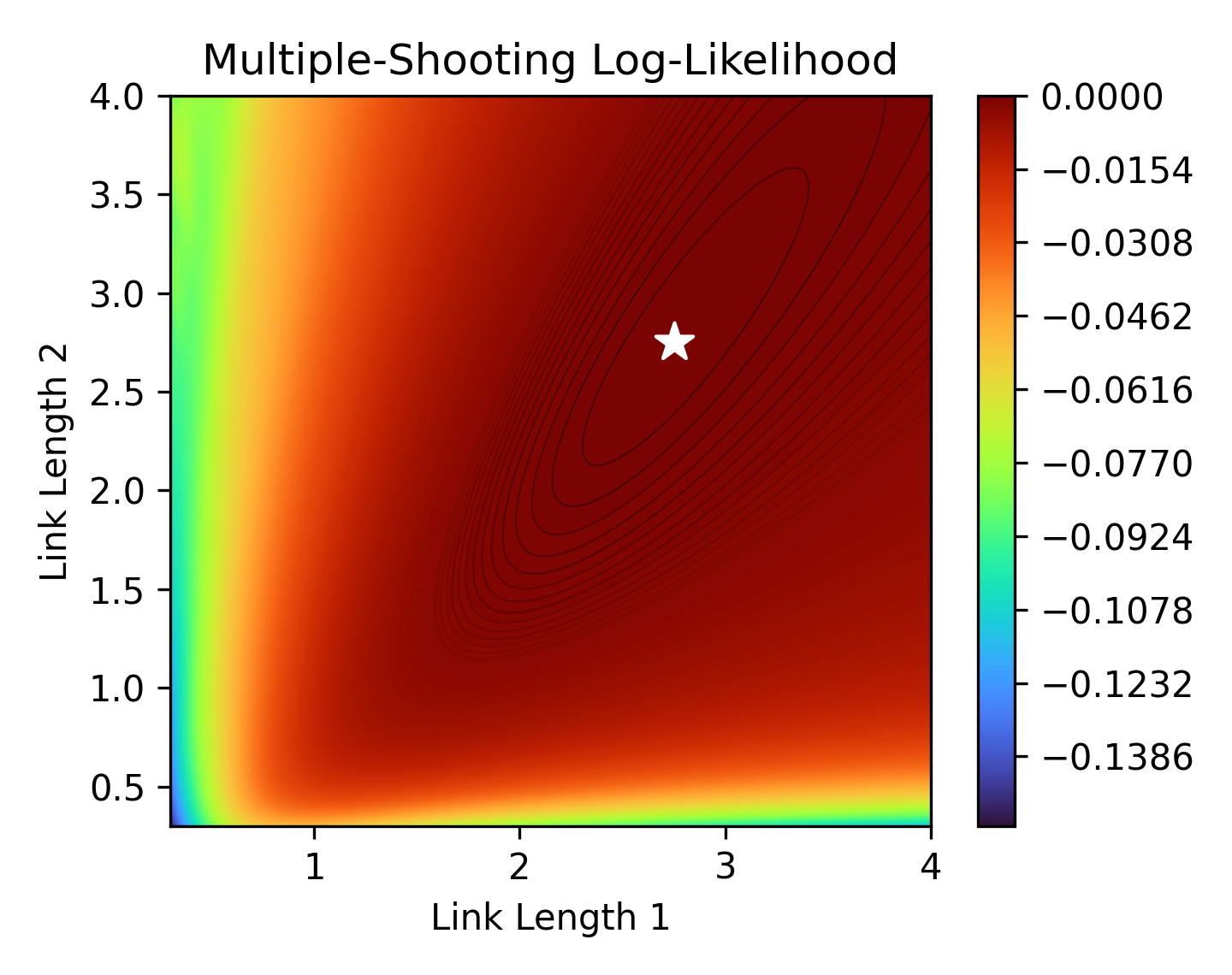}
        \caption{Multiple Shooting}
        \label{fig:double-pendulum-ll-ms}
    \end{subfigure}\\
    \begin{subfigure}[b]{0.9\columnwidth}
        \centering
        \includegraphics[height=\figheight]{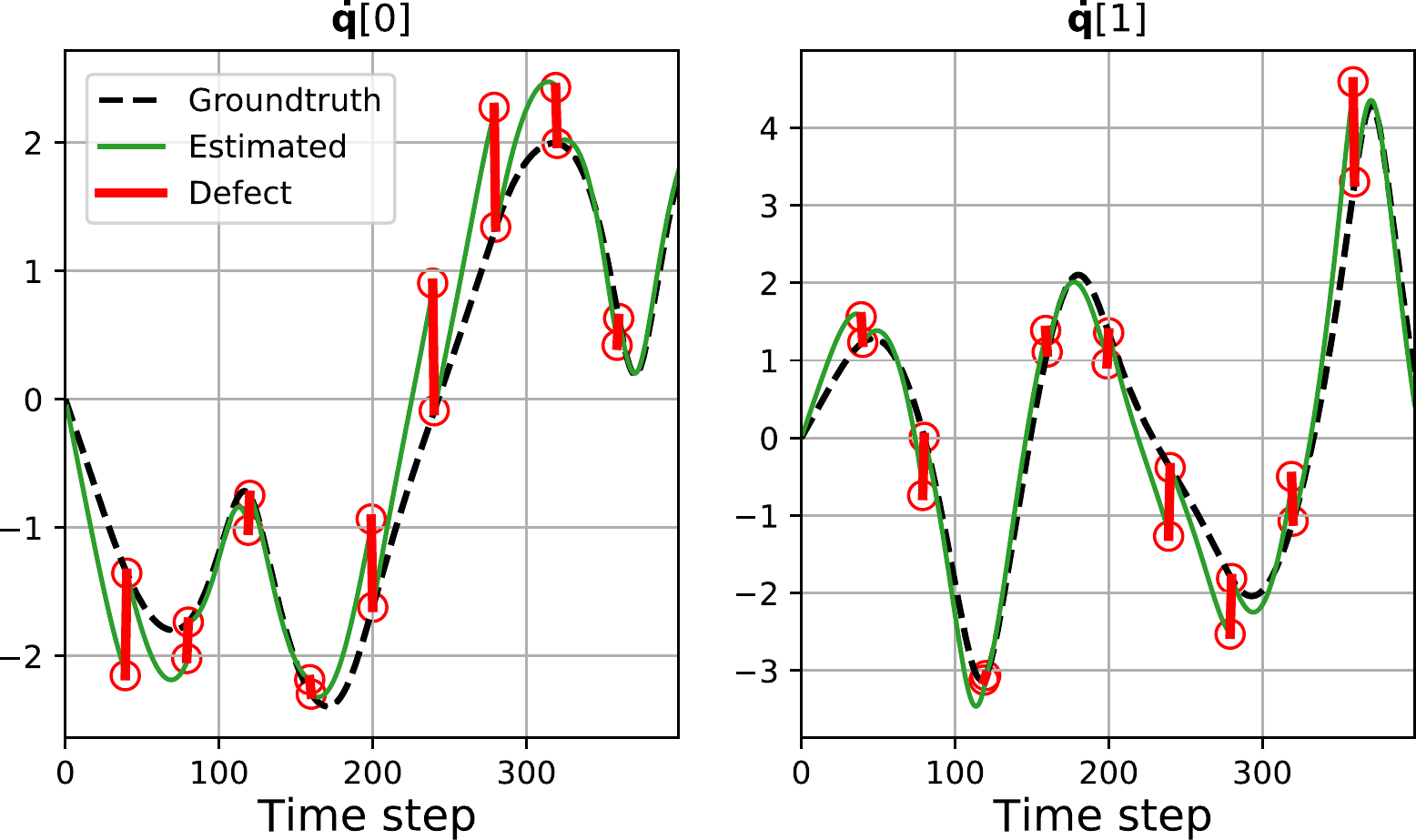}
        \caption{Shooting Windows and Defects}
        \label{fig:double-pendulum-ms-defects}
    \end{subfigure}
    \caption{The two heatmaps plots on the left show the landscape of the log-likelihood function for an inference problem where the two link lengths of a double pendulum are estimated (ground-truth parameters indicated by a white star). In (a), the likelihood is evaluated over a 400-step trajectory; in (b), the trajectory is split into 10 shooting windows and the likelihood is computed via~\autoref{eq:ms_likelihood}. In (c), the shooting intervals and defects are visualized for an exemplary parameter guess.\vspace*{0.5em}}
    \label{fig:double-pendulum-ms}
\end{figure}

We define $p (\trajectory\simu | \trajectory\real)$ as the probability of an individual trajectory $\trajectory\simu$ simulated from parameters $\theta$ with respect to a matched ground-truth trajectory $\trajectory\real$.
Following the HMM assumption from \autoref{sec:formulation}, we treat $p (\trajectory\simu | \trajectory\real)$ as the product of probabilities over observations, shown in \autoref{eq:combinedlikelihood}.
This assumption is justified because the next state $\statevec_{t+1}$ is fully determined by the position $\mathbf{q}_t$ and velocity $\mathbf{\dot{q}}_t$ of the current state $\statevec_t$ in articulated rigid body dynamics (see \autoref{sec:dynamics}).
To directly compare observations we use a Gaussian likelihood:
\begin{align}\label{eq:combinedlikelihood}
    p(\trajectory\real | \trajectory\simu) 
    &= \prod_{t \in T} p(\observationvec_t\real, \observationvec_t\simu) \\ \nonumber
    &= \prod_{t \in T}  \mathcal{N}(\observationvec_t\real | \observationvec_t\simu,  \sigma_{\text{obs}}^2).
\end{align}

This \revtwo{likelihood model for observations is} used to compute the likelihood for a trajectory
\begin{align}\label{eq:ss_likelihood}
    p_{ss}(\trajectory\real | \params) 
    &= p(\trajectory\real | \fobs (\fsim (\params, \statevec_0\real))) \\ \nonumber
    &= p(\trajectory\real | \trajectory\simu),
\end{align}
where $\statevec_0\real$ is (an estimate of) the first state of $\trajectory\real$.
This formulation is known as a \emph{single-shooting} estimation problem, where a trajectory is compared against another only by varying the initial conditions of the modeled system.
To evaluate the likelihood for a collection of ground-truth trajectories, $\trajectoryset\real$, we use an equally weighted Gaussian Mixture Model likelihood:
\begin{equation}\label{eq:traj_likelihood}
    \pobs (\trajectoryset\real | \params) = \sum_{\trajectory\real \in \trajectoryset\real} p_{ss}(\trajectory\real | \params).  
\end{equation}

\vspace*{-1em}
\subsection{Multiple Shooting}
\label{sec:multiple-shooting}

Estimating parameters from long trajectories can prove difficult in the face of observation noise, and for systems in which small changes in initial conditions produce large changes in trajectories, potentially resulting in poor local optima~\cite{aydogmus_modified_2020}.
We adopt the \emph{multiple-shooting} method which significantly improves the convergence and stability of the estimation process.
Multiple shooting has been applied in system identification problems~\cite{bock_recent_1983} and biochemistry~\cite{peifer_parameter_2007}.

Multiple shooting divides up the trajectory into $n_s$ shooting windows over which the likelihood is computed. To be able to simulate such shooting windows, we require their start states $\statevec^s$ to be available for $\fsim$ to generate a shooting window trajectory. Since we only assume access to the first true state $\statevec_0\real$ from the real trajectory $\trajectory\real$, we augment the parameter vector $\params$ by the start states of the shooting windows, which we refer to as \emph{shooting variables} $\statevec_{t}^s$ (for $t=h,2h,\dots,n_s\cdot h$).
We define an augmented parameter vector as $\bar{\params} = \bmat{\params & \statevec^s_h & \dots & \statevec^s_{n_s \cdot h}}$. A shooting window of $h$ time steps starting from time $t$ is then simulated via $\trajectory_t = \fobs(\fsim(\params, \statevec^s_{t})[0\!:\!h])$, where $[i\!:\!j]$ denotes a selection operation of the sub-array between the indices $i$ and $j$. Analogous to \autoref{eq:combinedlikelihood}, we evaluate the likelihood $p(\trajectory\real[t\!:\!t\!+\!h]~|~\trajectory_t\simu)$ for a single shooting window as a product of state-wise likelihoods.

To impose continuity between the shooting windows, \emph{defect constraints} are imposed as a Gaussian likelihood term between the last simulated state $\statevec_t$ from the previous shooting window at time $t$ and the shooting variable $\statevec_t^s$ at time $t$:
\begin{align}
    \label{eq:likelihood-defect}
    \pdef(\statevec_t^s, \statevec_t) = \mathcal{N}(\statevec_t^s | \statevec_t,\sigma^2_{\text{def}}) \qquad  t\in [h, 2h, \dots]
\end{align}
where $\sigma^2_{\text{def}}$ is a small variance \rev{so that the MCMC samplers adhere to the constraint}.
Including the defect likelihood allows the extension of the likelihood defined in \autoref{eq:ss_likelihood} to a multiple-shooting scenario:
\begin{align}
    \label{eq:ms_likelihood}
    p_{ms}(\trajectory\real | \exparams) &= \prod_{t\in  H} \pdef (\statevec^s_t , \statevec_t) ~ p(\trajectory\real[t\!:\!t\!+\!h]~|~\trajectory_t\simu), \\[-0.8em] \nonumber
    \statevec_0^s &= \statevec_0\real \qquad H = [0, h, 2h, \dots]
\end{align}
As for the single-shooting case, \autoref{eq:traj_likelihood} with $p_{ms}$ as the trajectory-wise likelihood function gives the likelihood $\pobs$ for a set of trajectories.

In \autoref{fig:double-pendulum-ms}, we provide a parameter estimation example where the two link lengths of a \rev{double} pendulum must be inferred.
The single-shooting likelihood from \autoref{eq:ss_likelihood} (shown in \autoref{fig:double-pendulum-ll-ss}) exhibits a rugged landscape where many estimation algorithms will require numerous samples to escape from poor local optima, or a much finer choice of parameter prior to limit the search space. The multiple-shooting likelihood (shown in \autoref{fig:double-pendulum-ms}) is significantly smoother and therefore easier to optimize.

\subsection{Parameter Limits as a Uniform Prior}
\label{sec:prior}

Simulators may not be able to handle simulating trajectories from any given parameter and it is often useful to enforce some limits on the parameters.
To model this, we define a uniform prior distribution on the parameter settings $
    \plim (\params) = \prod_{i=1}^\paramdim U(\params_i | \params_{\text{min}_i}, \params_{\text{max}_i})
$,
where $\params_{\text{min}_i}, \params_{\text{max}_i}$ denote the upper and lower limits of parameter dimension $i$.

\subsection{Constrained Optimization for SVGD}
\label{sec:mmdm}

Directly optimizing SVGD for the unnormalized posterior on long trajectories, with a uniform prior, can be difficult for gradient based optimizers. The uniform prior has discontinuities at the extremities, effectively imposing constraints, which produce non-differentiable regions in the parameter space domain. We propose an alternative solution to deal with this problem and treat SVGD as a constrained optimization on $p_{\text{obs}}$ with $\pdef$ and $\plim$ as constraints.

A popular method of constrained optimization is the Modified Differential Multiplier Method (MDMM)~\cite{platt1988constrained} which augments the cost function to penalize constraint violations. In contrast to the basic penalty method, MDMM uses Lagrange multipliers in place of constant penalty coefficients that are updated automatically during the optimization:
\begin{align}\vspace*{-0.5em}
    \operatorname{maximize} & \qquad \log p_{\text{obs}}( \trajectoryset=\trajectoryset\real \ |\  \params ) \\ \nonumber
    \text{s.t.}             & \qquad g(\exparams) = 0.
\end{align}
MDMM formulates this setup into an unconstrained minimization problem by introducing Lagrange multipliers $\lambda$ (initialized with zero):
\begin{align}
    \label{eq:mdmm-loss}
    \mathcal{L}_{c}(\exparams, \lambda) = -\log p_{\text{obs}}(\trajectoryset=\trajectoryset\real | \params) + \lambda g(\exparams) + \frac{c}{2}[g(\exparams)]^2,
\end{align}
where $c>0$ is a constant damping factor that improves convergence in gradient descent algorithms.
To accommodate these Lagrange multipliers per-particle we again extend the parameter set ($\exparams$) introduced in \autoref{sec:multiple-shooting} to store the multiple shooting variables and Lagrange multipliers, $\exparams = (\params, \statevec^s, \lambda_{\text{def}}, \lambda_{\text{lim}})$.
The following update equations for $\params$ and $\lambda$ are used to minimize \autoref{eq:mdmm-loss}:
\begin{align*}
    \dot{\params} & = \frac{\partial \log p_{\text{obs}}(\trajectoryset\real | \params)}{\partial \params} - \lambda \frac{\partial g(\exparams)}{\partial \params} -cg(\exparams) \frac{\partial g(\exparams)}{\partial \params} \\ \nonumber
    \dot{\lambda} &= g(\exparams)
\end{align*}

We include parameter limit priors from \autoref{sec:prior} as \rev{the} following equality constraints (where $\operatorname{clamp}(x,a,b)$ clips the value $x$ to the interval $[a,b]$): $
    g_{\text{lim}}(\params) = \operatorname{clamp}(\params, \params_{\text{min}}, \params_{\text{max}}) - \params.
$
Other constraints are the defect constraints from
\autoref{eq:likelihood-defect} which are included as equality constraints as well:
$g_{\text{def}}(\params, \statevec^s_t)
    = \log \pdef (\statevec^s_t, \statevec_t)
    = \| \statevec^s_t - \statevec_t \|^2 / \sigma^2_{\text{def}}$.

The overall procedure of our Constrained SVGD (CSVGD) algorithm is given in Algorithm~1.\vspace*{-0.8em}

\begin{algorithm}
    \label{alg:csvgd}
    \SetAlgoLined
    \textbf{Inputs:} differentiable simulator $\fsim: (\params, \rev{\statevec_0}) \mapsto [\statevec]$, observation function $\fobs: \statevec\mapsto\observationvec$,
    start states $\statevec_0^i$ for each ground-truth trajectory $\trajectory_i\real\in\trajectoryset\real$,
    learning rate scheduler (e.g. Adam), kernel choice (e.g. RBF)\\
    \For{$i = 1 \dots \operatorname{max\_iterations}$}{
    Roll out simulated observations $\trajectoryset\simu = [\fobs(\fsim(\params,\statevec_0\real)) ~ \forall \trajectory\real \in \trajectoryset\real]$ \\
    Compute $\phi(\params)$ via \autoref{eq:svgd_update} and $\log p_{\text{obs}}( \trajectoryset=\trajectoryset\real \ |\  \params )$\\
    Update $\params$ via $\dot{\params} = \phi(\params)
        - \lambda_{\text{lim}}\frac{\partial g_{\text{lim}}}{\partial\params}
        - cg_{\text{lim}}\frac{\partial g_{\text{lim}}}{\partial\params}
        - \lambda_{\text{def}}\frac{\partial g_{\text{def}}}{\partial\params}
        - cg_{\text{def}}\frac{\partial g_{\text{def}}}{\partial\params}$ \\
    Update $\lambda_{\text{lim}}, \lambda_{\text{def}}, {\statevec^s}_t$ via
    $
        \dot{\lambda}_{\text{lim}} = g_{\text{lim}}(\params),
        \dot{\lambda}_{\text{def}} = g_{\text{def}}(\params),
        \dot{\statevec^s}_t = g_{\text{def}}(\params, \statevec^s_t)
    $ for $t\in[h,2h,\dots]$
    }
    \textbf{return} $\params$
    \caption{Constrained SVGD}
\end{algorithm}



\subsection{Performance Metrics for Particle Distributions}
\label{sec:metrics}
In most cases we do not know the underlying ground-truth parameter distribution $p(\params\real)$, and only have access to a finite set of ground-truth trajectories. Therefore, we measure the discrepancy between trajectories rolled out from the estimated parameter distribution, $\trajectoryset\simu$, and the reference trajectories, $\trajectoryset\real$.

One measure, the KL divergence, is the expected value of the log likelihood ratio between two distributions.
Although the KL divergence cannot be calculated from samples of continuous distributions, methods have been developed to estimate it from particle distributions using the $k$-nearest neighbors distance~\cite{wang_divergence_2009}.
The KL divergence is non-symmetric and this estimate can be poor in two situations.
The first is estimating $d_{\text{KL}}(\trajectoryset\simu \parallel \trajectoryset\real)$ when the particles are all very close to one trajectory, causing a low estimated divergence, yet the posterior is of poor quality.
The opposite can happen when the particles in the posterior are overly spread out when estimating $d_{\text{KL}}(\trajectoryset\real \parallel \trajectoryset\simu)$.

We additionally measure the maximum mean discrepancy (MMD)~\cite{gretton2012mmd}, which is a metric used to determine if two sets of samples are drawn from the same distribution by calculating the square of the distance between the embedding of the two distributions in a RKHS.

\vspace*{-0.5em}
\section{Experiments}
\vspace*{-0.5em}
\label{sec:experiments}
We compare our method against commonly used parameter estimation baselines. As an algorithm comparable to our particle-based approach, we use the Cross Entropy Method (CEM)~\cite{rubinstein1997optimization}.
In addition, we evaluate the Markov chain Monte-Carlo techniques Emcee~\cite{foreman2013emcee}, Stochastic Gradient Langevin Dynamics (SGLD)~\cite{welling_bayesian_2011}, and the No-U-Turn-Sampler (NUTS)~\cite{hoffman_no-u-turn_2011}, which is an adaptive variant of the gradient-based Hamiltonian MC algorithm. For Emcee, we use a parallel sampler that implements the ``stretch move'' ensemble method~\cite{goodman_ensemble_2010}. For BayesSim, we report results from the best performing instantiation using a mixture density random Fourier features model (MDRFF) with Matérn kernel in \autoref{tab:system-accuracy}.
We present further details and extended results from our experiments in the appendix.

\vspace*{-0.5em}
\subsection{Parameter Estimation Accuracy}
\begin{figure}
    \newcommand{\figheight}{2.5cm}
    \newcommand{\subfigwidth}{.3\columnwidth}
    \begin{subfigure}[b]{\subfigwidth}
        \centering
        \includegraphics[height=\figheight,trim=0 0.2cm 5.8cm 0.8cm,clip]{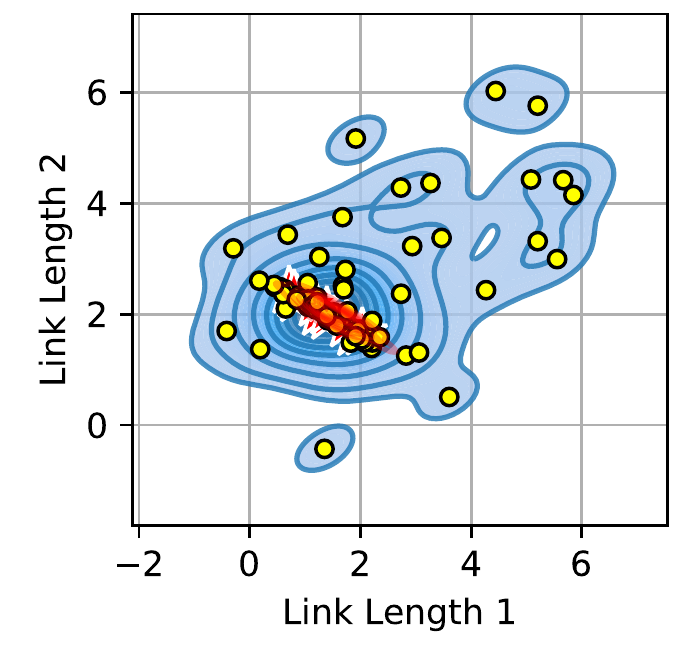}\vspace*{-0.5em}
        \caption{Emcee}
        \label{fig:exp-param-noise-posterior-emcee}
    \end{subfigure}
    \begin{subfigure}[b]{\subfigwidth}
        \centering
        \includegraphics[height=\figheight,trim=0 0.2cm 5.8cm 0.8cm,clip]{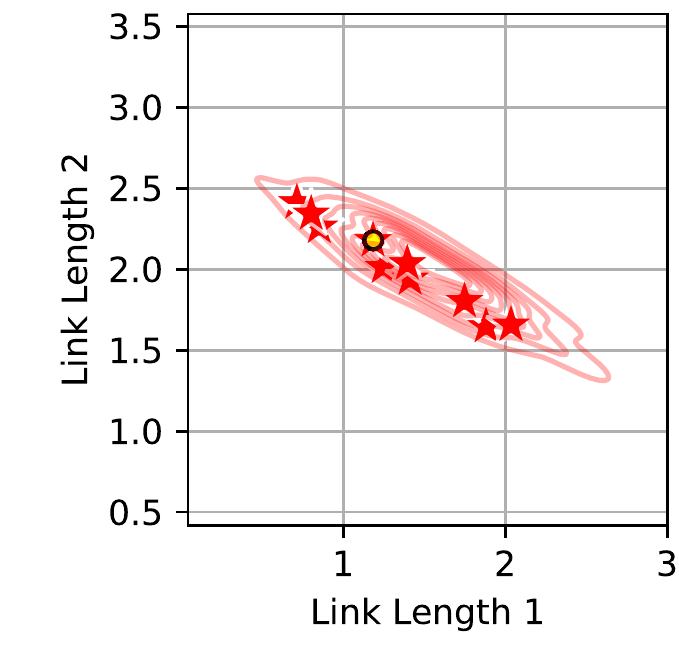}\vspace*{-0.5em}
        \caption{CEM}
        \label{fig:exp-param-noise-posterior-cem}
    \end{subfigure}
    \begin{subfigure}[b]{\subfigwidth}
        \centering
        \includegraphics[height=\figheight,trim=0 0.2cm 5.8cm 0.8cm,clip]{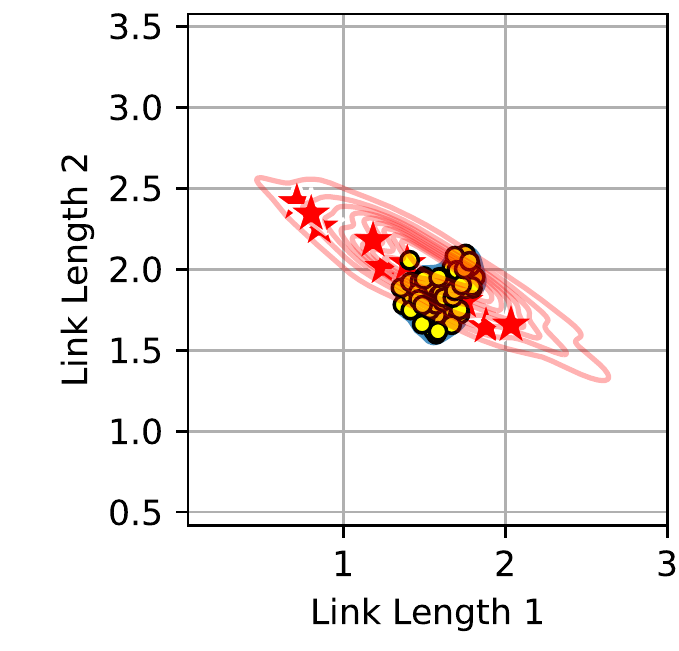}\vspace*{-0.5em}
        \caption{SGLD}
        \label{fig:exp-param-noise-posterior-sgld}
    \end{subfigure}
    \begin{subfigure}[b]{\subfigwidth}
        \centering
        \includegraphics[height=\figheight,trim=0 0.2cm 5.8cm 0.8cm,clip]{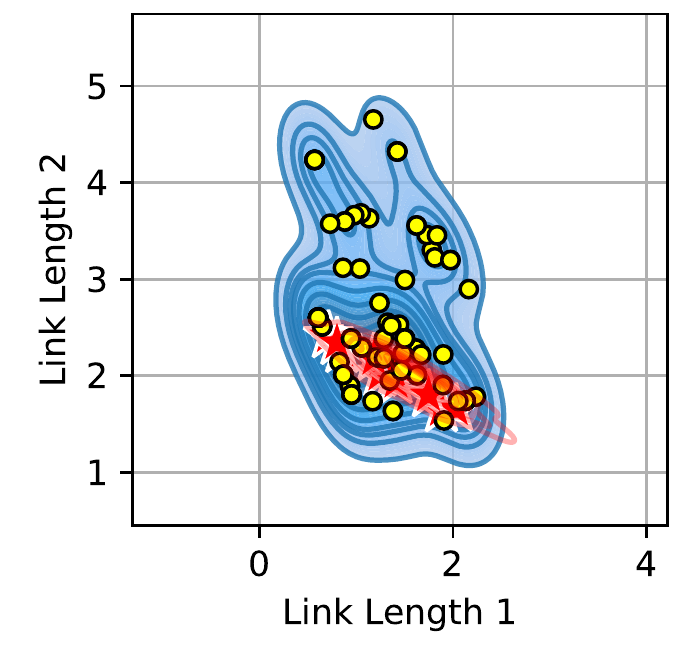}\vspace*{-0.5em}
        \caption{NUTS}
        \label{fig:exp-param-noise-posterior-nuts}
    \end{subfigure}
    \begin{subfigure}[b]{0.2\textwidth}
        \centering
        \includegraphics[height=\figheight,trim=0 0.2cm 0 0.8cm,clip]{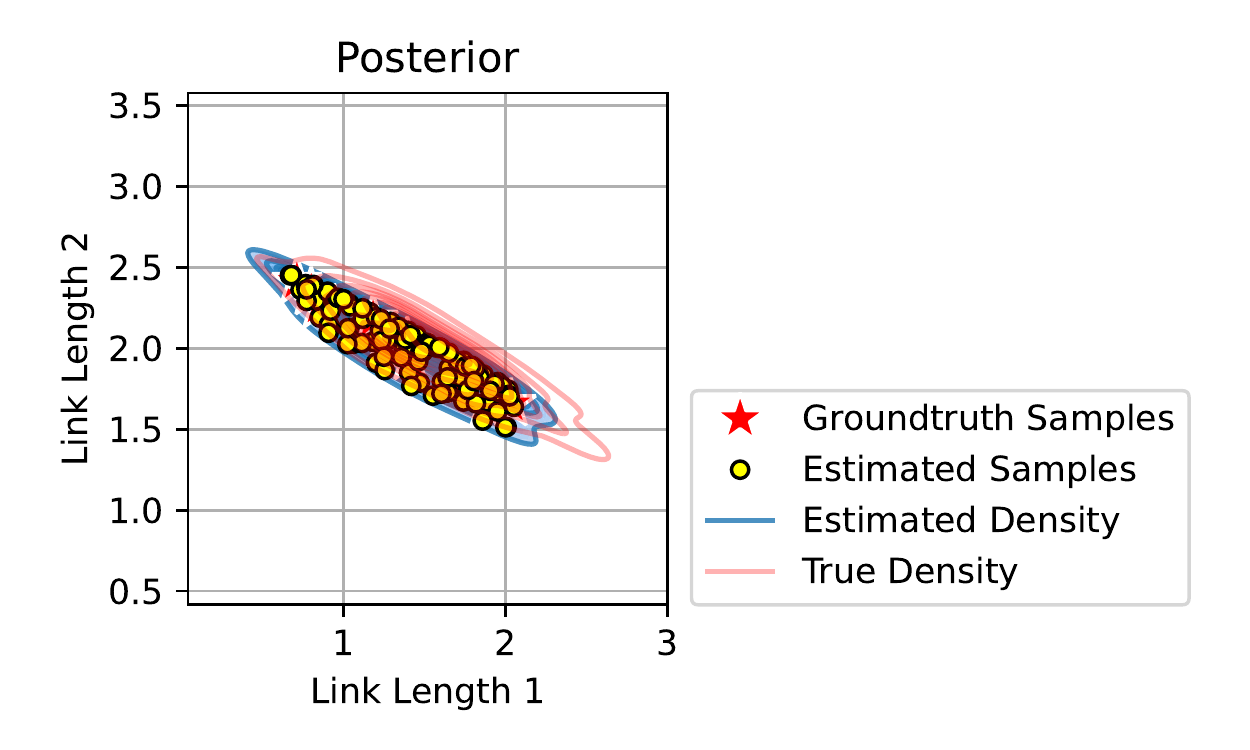}\vspace*{-0.5em}
        \caption{SVGD}
        \label{fig:exp-param-noise-posterior-svgd}
    \end{subfigure}
    \caption{Estimated posterior distributions from synthetic data generated from a known multivariate Gaussian distribution (\autoref{sec:exp-param-noise}). The 100 last samples were drawn from the Markov chains sampled by Emcee, SGLD, and NUTS, while CEM and SVGD used 100 particles.}
    \label{fig:exp-param-noise-posterior}
\end{figure}

\begin{figure}
    \centering
    \newcommand{\figheight}{2.8cm}
    \begin{subfigure}[b]{0.23\textwidth}
        \centering
        \includegraphics[height=\figheight]{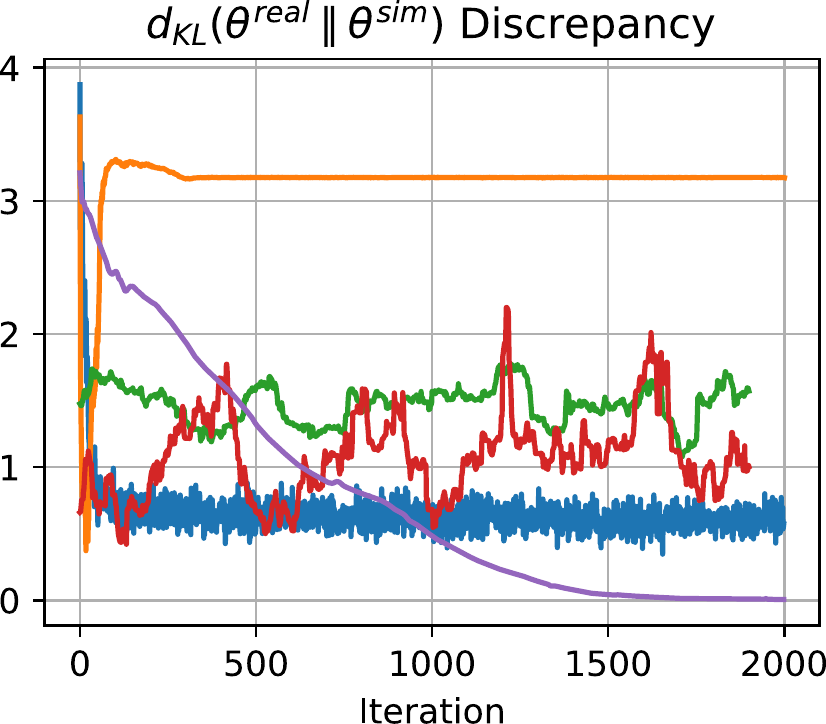}
        \caption{}
        \label{fig:exp-param-noise-metrics-kl-real-sim}
    \end{subfigure}
    \begin{subfigure}[b]{0.23\textwidth}
        \centering
        \includegraphics[height=\figheight]{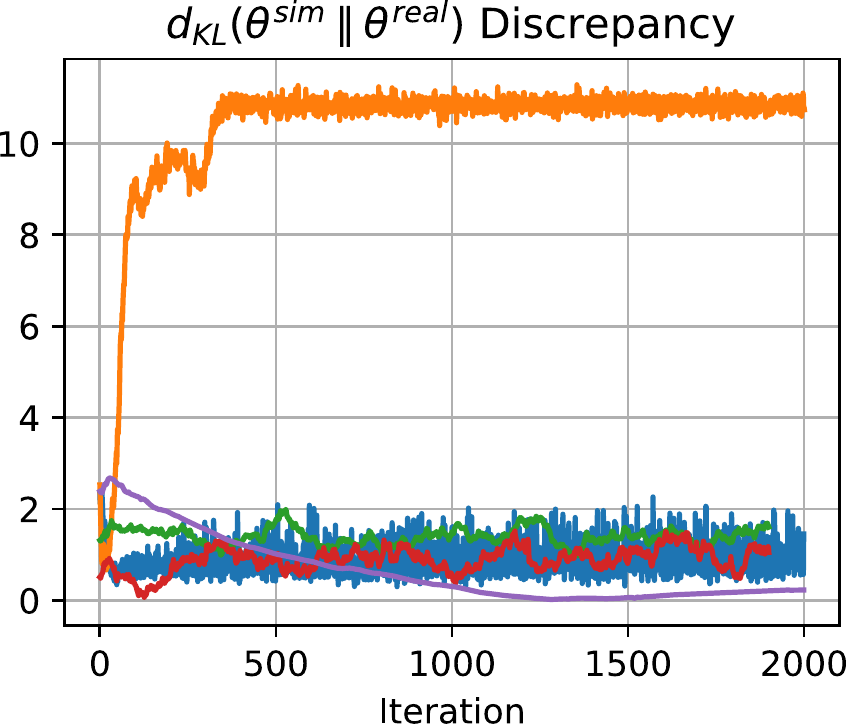}
        \caption{}
        \label{fig:exp-param-noise-metrics-kl-sim-real}
    \end{subfigure}
    \begin{subfigure}[b]{0.23\textwidth}
        \centering
        \includegraphics[height=\figheight]{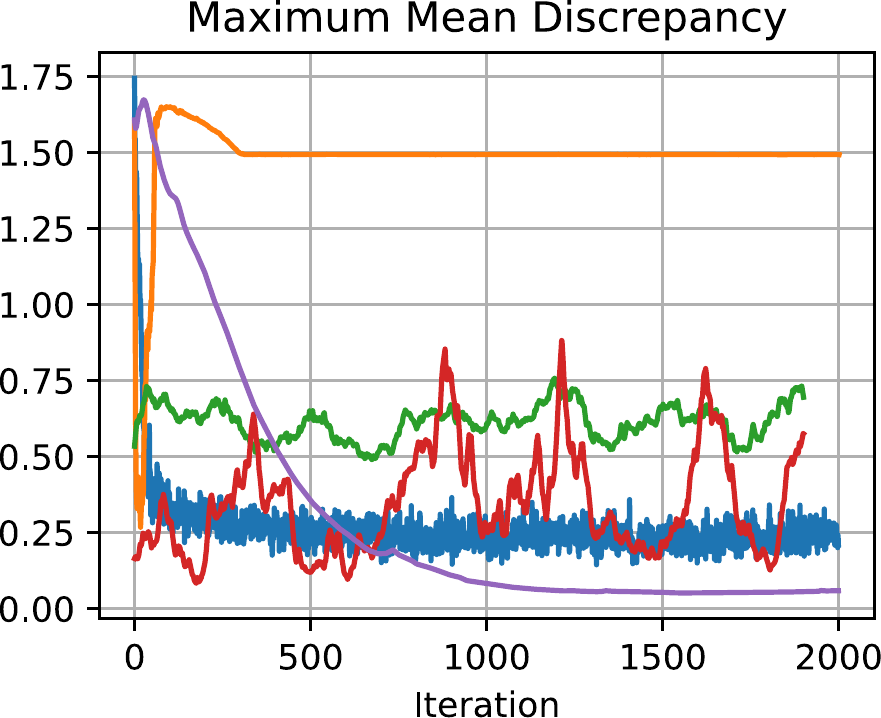}
        \caption{}
        \label{fig:exp-param-noise-metrics-mmd}
    \end{subfigure}
    \begin{subfigure}[b]{0.23\textwidth}
        \centering
        \includegraphics[height=\figheight]{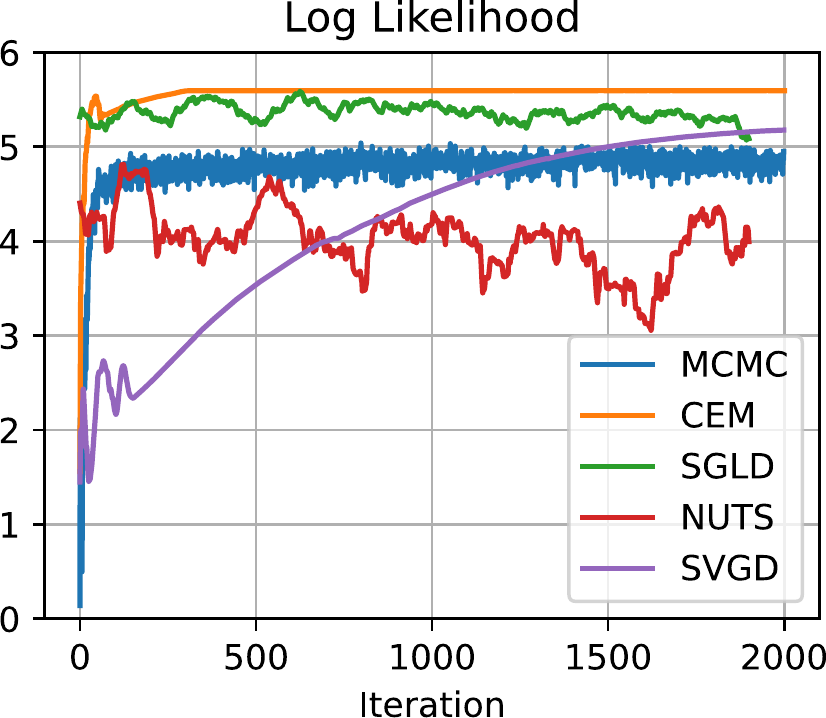}
        \caption{}
        \label{fig:exp-param-noise-metrics-ll}
    \end{subfigure}
    \caption{Accuracy metrics for the estimated parameter posteriors shown in \autoref{fig:exp-param-noise-posterior}.
        The estimations were done on the synthetic dataset of a multivariate Gaussian over parameters (\autoref{sec:exp-param-noise}).
        \autoref{fig:exp-param-noise-metrics-ll} shows the single-shooting likelihood using the equation described in \autoref{eq:ss_likelihood}.
\vspace*{-1.0em}
    }
    \label{fig:exp-param-noise-metrics}
\end{figure}

\vspace*{-0.5em}
\label{sec:exp-param-noise}
We create a synthetic dataset of 10 trajectories from a double pendulum which we simulate by varying its two link lengths. These two \rev{uniquely identifiable~\cite{fazeli2018identifiability}} parameters are drawn from a Gaussian distribution with a mean of $(\SI{1.5}{\meter}, \SI{2}{\meter})$ and a full covariance matrix (density visualized by the red contour lines in \autoref{fig:exp-param-noise-posterior}).
We show the evolution of the consistency metrics in \autoref{fig:exp-param-noise-metrics}. Trajectories generated by evaluating the posteriors of the compared methods are compared against 50 test trajectories rolled out from the ground-truth parameter distribution.

We find that SVGD produces a density very close to the density that matches the principal axes of the ground-truth posterior, and outperforms the other methods in all metrics except log likelihood.
CEM collapses to a single high-probability estimate but does not accurately represent the full posterior, as can be seen in \autoref{fig:exp-param-noise-metrics-ll} being maximal quickly but poor performance on the other metrics which measure the spread of the posterior.
Emcee and NUTS represent the spread of the posterior but do not sharply capture the high-likelihood areas, shown by their good performance in \autoref{fig:exp-param-noise-metrics-kl-sim-real}, \autoref{fig:exp-param-noise-metrics-kl-sim-real} and \autoref{fig:exp-param-noise-metrics-mmd}.
SGLD captures a small amount of the posterior around the high likelihood points but does not fully approximate the posterior.

\begin{table*}[]
    \centering
    \resizebox{0.8\textwidth}{!}{%
        \begin{tabular}{llrrrrrrr}
            \toprule
            \bf Experiment  & \bf Metric                                                          & \bf Emcee & \bf CEM   & \bf SGLD  & \bf NUTS  & \bf SVGD & \bf BayesSim  & \bf CSVGD (Ours) \\
            \midrule
            Double Pendulum & $d_{\text{KL}} (\trajectoryset\real \parallel \trajectoryset\simu)$ & 8542.2466 & 8911.1798 & 8788.0962 & 9196.7461 & 8803.5683  & 8818.1830 & \bf 5204.5336    \\
                            & $d_{\text{KL}} (\trajectoryset\simu \parallel \trajectoryset\real)$ & 4060.6312 & 8549.5927 & 7876.0310 & 6432.2131 & 10283.6659 & 3794.9873 & \bf 2773.1751    \\
                            & MMD                                                                 & 1.1365    & 0.9687    & 2.1220    & 0.5371    & 0.7177     & 0.6110    & \bf 0.0366       \\ \midrule
            Panda Arm       & $\log\pobs(\trajectoryset\real \parallel \trajectoryset\simu)$      & -16.1185  & -17.3331  & -17.3869  & -17.9809  & -17.7611   & -17.6395  & \bf -15.1671     \\
            \bottomrule                                                                                                                                                           \\
        \end{tabular}
    }\vspace*{-0.5em}
    \caption{Consistency metrics of the posterior distributions approximated by the different estimation algorithms. Each metric is calculated across simulated and real trajectories. Lower is better on all metrics except $\log\pobs(\trajectoryset\real \parallel \trajectoryset\simu)$.}
    \label{tab:system-accuracy}
    \vspace*{-1.5em}
\end{table*}

\vspace*{-0.5em}
\subsection{Identify Real-world Double Pendulum}
\vspace*{-0.5em}
\label{sec:exp-ibm-pendulum}

We leverage the dataset from~\cite{asseman2018learning} containing trajectories from a physical double pendulum. While in our previous synthetic data experiment the parameter space was reduced to only the two link lengths, we now \rev{define 11 parameters} to estimate.
The parameters for each link are the mass, inertia $I_{xx}$, the center of mass in the $x$ and $y$ direction, and the joint friction. We also estimate the length of the second link.
Note that parameters, such as the length of the first link, are not explicitly included since they are captured by the remaining parameters, which we validated through sensitivity analysis.
Like before, the state space is completely measurable, i.e. $\statevec \approx \observationvec$, except for observation noise.

\begin{figure}
    \centering
    \newcommand{\figwidth}{1.05\textwidth}
    \begin{subfigure}[b]{0.49\columnwidth}
        \centering
        \includegraphics[width=\figwidth]{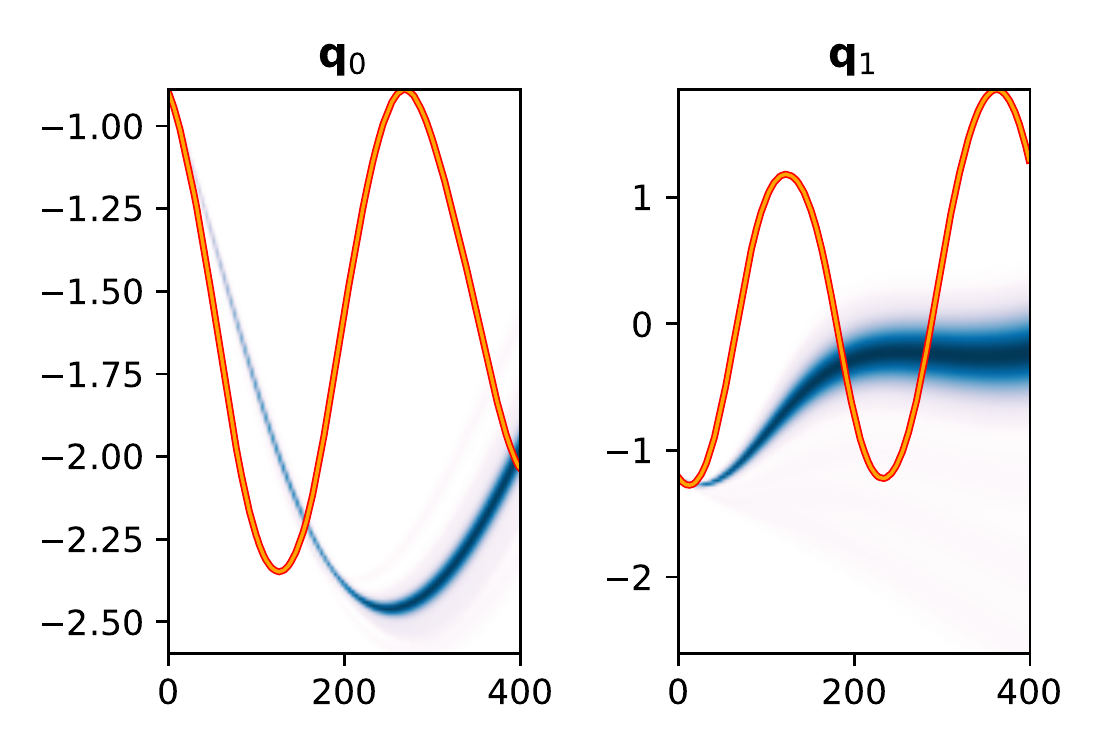}
        \caption{SVGD}
        \label{fig:svgd_ibm_ss}
    \end{subfigure}\hfill
    \begin{subfigure}[b]{0.49\columnwidth}
        \centering
        \includegraphics[width=\figwidth]{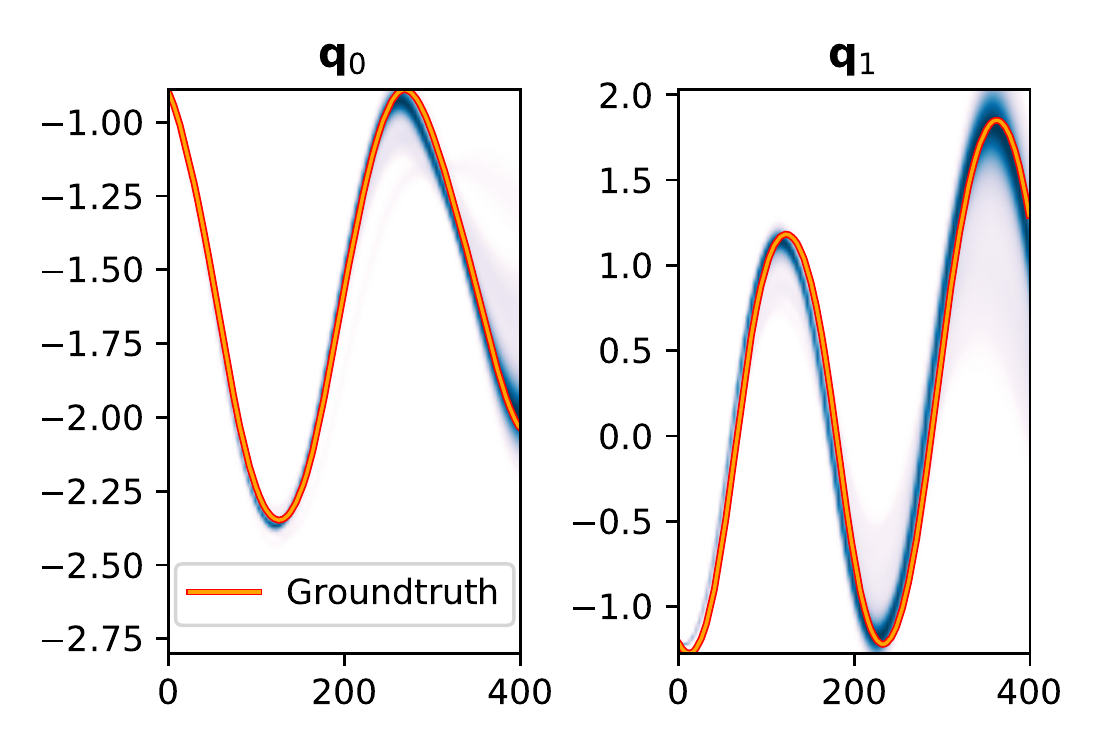}
        \caption{CSVGD}
        \label{fig:svgd_ibm_ms}
    \end{subfigure}
    \caption{Trajectory density plots (only joint positions $\jointpos$ are shown) obtained by simulating the particle distribution found by SVGD with the single-shooting likelihood (\autoref{fig:svgd_ibm_ss}) and the multiple-shooting likelihood (\autoref{fig:svgd_ibm_ms}) on the real-world double pendulum (\autoref{sec:exp-ibm-pendulum}).}
    \label{fig:ibm-pendulum-traj-density}
\end{figure}

In this experiment we find that CSVGD outperforms all other methods in KL divergence (both ways) as well as MMD, shown in \autoref{tab:system-accuracy}.
We believe this is because of the complex \revtwo{relationship between the} parameters of each link and the \revtwo{resulting} observations. \revtwo{This} introduces many local minima which are hard to escape from (see \autoref{fig:ibm-pendulum-traj-density}).
The multiple-shooting likelihood improves the convergence significantly by simplifying the optimization landscape.

\vspace*{-0.5em}
\subsection{Identify Inertia of an Articulated Rigid Object}
\vspace*{-0.5em}
\label{sec:panda-box}

In our final experiment, we investigate a more complicated physical system which is underactuated. Through an uncontrollable universal joint, we attach an acrylic box to the end-effector of a physical 7-DOF Franka Emika Panda robot arm. We fix two \SI{500}{\gram} weights at the bottom inside \rev{the} box, and prescribe a trajectory which the robot executes through a PD controller. By tracking the motion of the box via a VICON motion capture system, we aim to identify the 2D locations of the two weights (see \autoref{fig:panda-sim2real}).
The system only has access to the proprioceptive measurements from the robot arm (seven joint positions and velocities), as well as the 3D pose of the box at each time step.
In the first phase, we identify the inertia properties of the empty box, as well as the joint friction parameters of the universal joint from real-robot trajectories of shaking an empty box. Given our best estimate, in the actual parameter estimation setup we infer the two planar positions of the weights.

The particles from SVGD and CSVGD quickly converge in a way that the two weights are aligned opposed to each other. If the weights were not at locations symmetrical about the center, the box would tilt and yield a large discrepancy to the real observations. MCMC, on the other hand, even after more than ten times the number of iterations, only rarely approaches configurations in which the box remains balanced.
In \autoref{fig:panda-sim2real} \revtwo{(right)} we visualize the posterior over weight locations found by CSVGD (blue shade). The found symmetries are clearly visible when we draw lines (orange) between the inferred positions of both weights (yellow, green), while the true weight locations (red) are contained in the approximated distribution.
As can be seen in \autoref{tab:system-accuracy}, the log likelihood is maximized by CSVGD.
The results indicate the ability of our method to accurately model difficult posteriors over complex trajectories because of the symmetries underlying the simulation parameters.


\section{Conclusion}
We have presented Constrained Stein Variational Gradient Descent (CSVGD), a new method for estimating the distribution over simulation parameters that leverages Bayesian inference and parallel, differentiable simulators. By segmenting the trajectory into multiple shooting windows via hard defect constraints, and effectively using the likelihood gradient, CSVGD produces more accurate posteriors and exhibits improved convergence over previous estimation algorithms.

In future work, we plan to leverage the probabilistic predictions from our simulator for uncertainty-aware control applications. Similar to~\cite{lambert_stein_2020}, the particle-based uncertainty information can be leveraged by a model-predictive controller that takes into account the multi-modality of future outcomes.




\printbibliography

\end{refsection}


\begin{refsection}

\newpage
\appendices

\section{Technical Details}
In this section we provide further technical details on our approach.

\subsection{Initializing the Estimators}
For each experiment, we initialize the particles for the estimators via the deterministic Sobol sequence on the intervals specified through the parameter limits. Our code uses the Sobol sequence implementation from~\cite{burkardt2021sobol} which is based on a Fortran77 implementation by~\cite{bennett1986sobol}.
For the MCMC methods that sample a single Markov chain, we used the center point between the parameter limits as initial guess.

\subsection{\rev{Likelihood Model for Sets of Trajectories}}

\begin{figure}[H]
    \centering
    \newcommand{\figheight}{4.3cm}
    \begin{subfigure}[b]{0.48\columnwidth}
        \centering
        \hspace*{-1em}
        \includegraphics[height=\figheight]{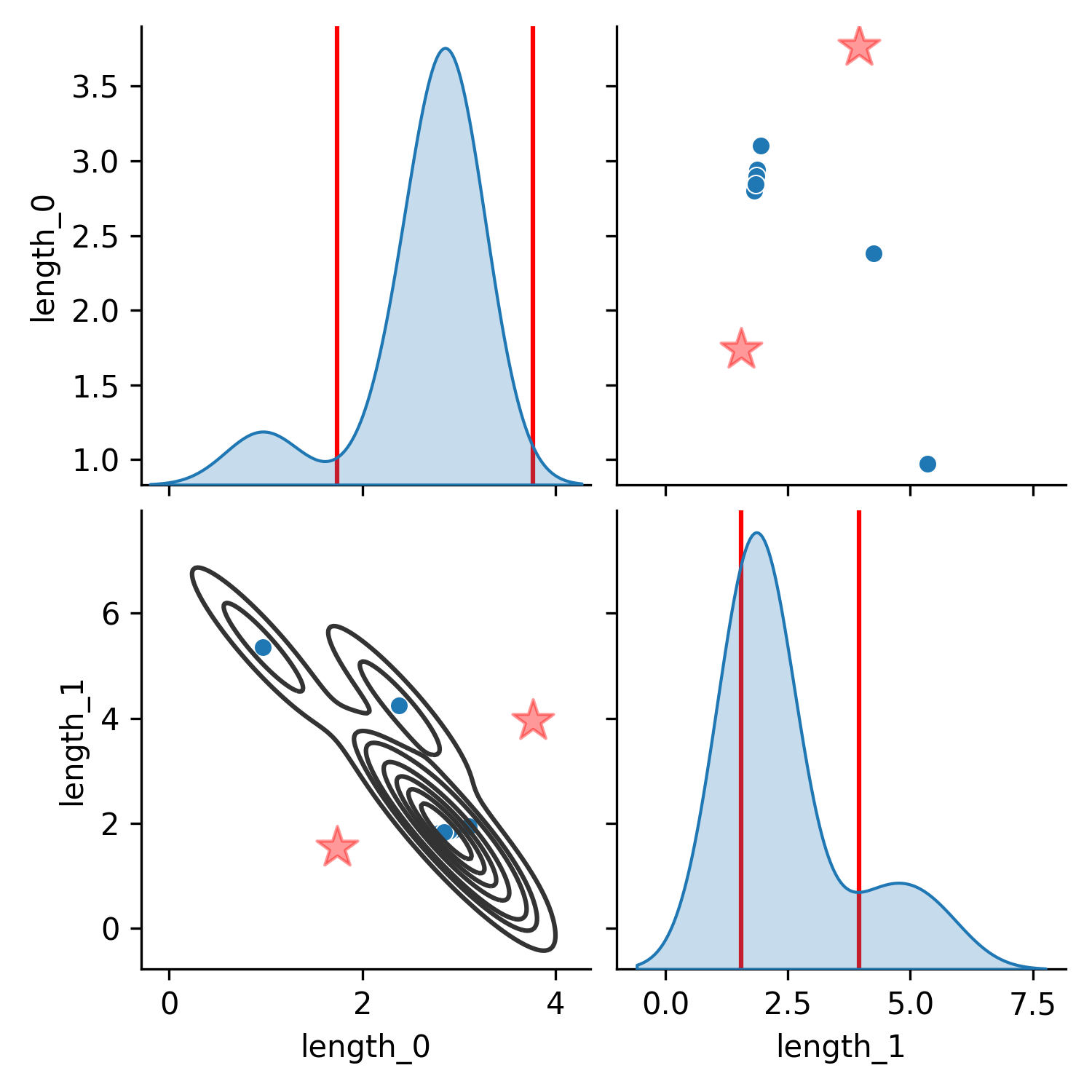}
        \caption{Product}
        \label{fig:double-pendulum-prod}
    \end{subfigure}
    \begin{subfigure}[b]{0.48\columnwidth}
        \centering
        \includegraphics[height=\figheight]{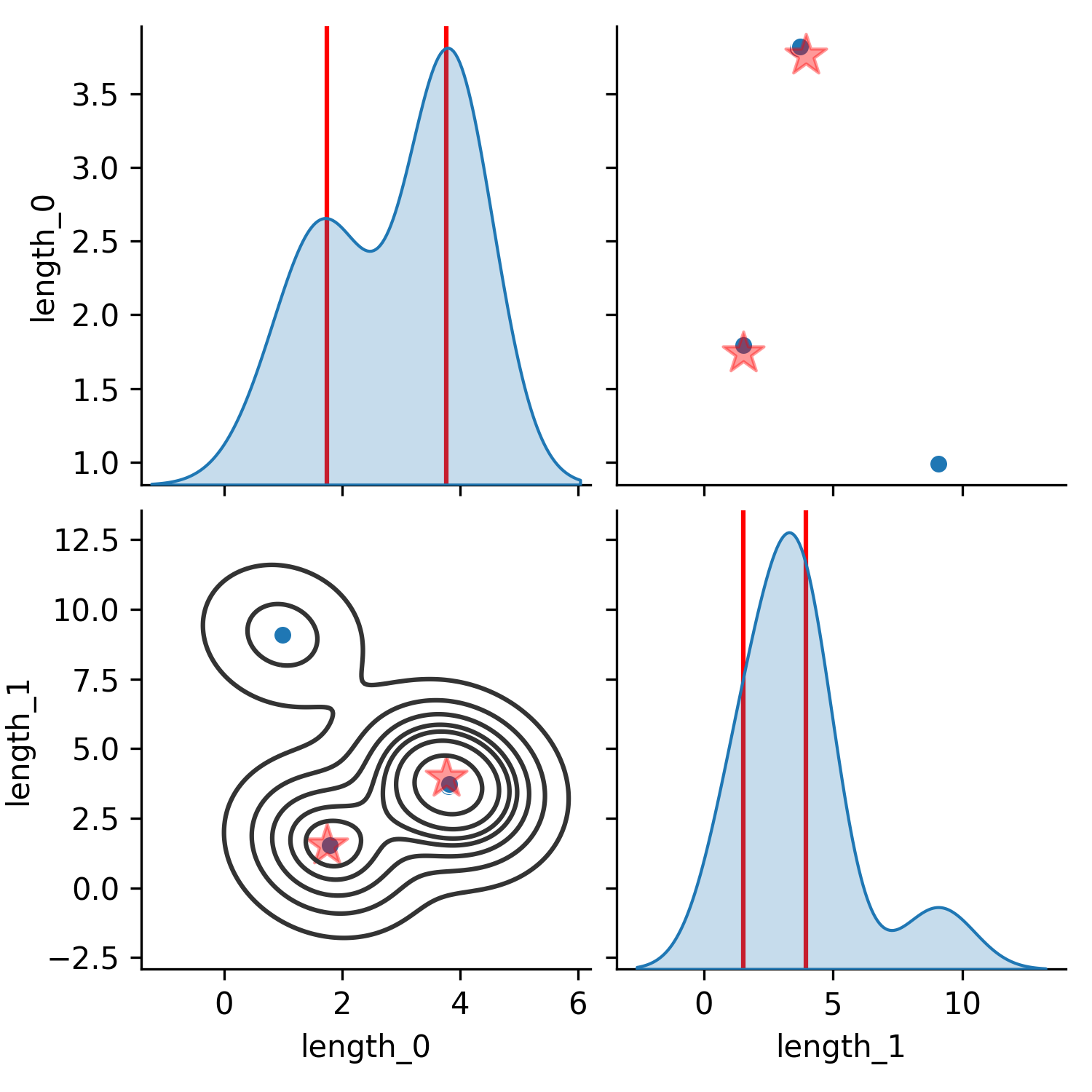}
        \caption{Sum}
        \label{fig:double-pendulum-sum}
    \end{subfigure}
    \caption{Comparison of posterior parameter distributions obtained from fitting the parameters to two ground-truth trajectories generated from different link lengths of a simulated double pendulum (units of the axes in meters). The trajectories were 300 steps long (which corresponds to a length of \SI{3}{\second}) and contain the 2 joint positions and 2 joint velocities of the uncontrolled double pendulum which starts from a zero-velocity initial configuration where the first angle is at $90^\circ$ (sideways) and the other at $0^\circ$. In (a), the product of the individual per-trajectory likelihoods is maximized (\autoref{eq:traj_likelihood_prod}). In (b), the sum of the likelihoods is maximized (\autoref{eq:traj_likelihood_sum}).}
    \label{fig:combination}
\end{figure}

In this work we assume that trajectories may have been generated by a distribution over parameters.
In the case of a replicable experimental setup, this could be a point distribution at the only true parameters.
\rev{However, when trajectories are collected from multiple robots}, or with slightly different experimental setups between \rev{experiments}, there may be a multimodal distribution over parameters which generated \rev{the set of} trajectories.

\rev{Note, that irrespective of the choice of likelihood function we do not make any assumption about the shape of the posterior distribution by leveraging SVGD which is a non-parametric inference algorithm. In trajectory space, the Gaussian likelihood function is a common choice as it corresponds to the typical least-squares estimation methodology. Other likelihood distributions may be integrated with our method, which we leave up to future work.}

The likelihood which we use is a mixture of equally-weighted Gaussians centered at each reference trajectory $\trajectory\real$:
\begin{equation}\label{eq:traj_likelihood_sum}
    p_{sum} (\trajectoryset\real | \params) = \sum_{\trajectory\real \in \trajectoryset\real} p_{ss}(\trajectory\real | \params).  
\end{equation}

\rev{If we were to consider each trajectory as an independent sample from the same trajectory distribution (the product), the likelihood function would be}
\begin{equation}\label{eq:traj_likelihood_prod}
    p_{product} (\trajectoryset\real | \params) = \prod_{\trajectory\real \in \trajectoryset\real} p_{ss}(\trajectory\real | \params).  
\end{equation}

\rev{While both Eqs.~\eqref{eq:traj_likelihood_sum} and~\eqref{eq:traj_likelihood_prod} define the likelihood for a combination of single-shooting likelihood functions $p_{ss}$ for a set of real trajectories $\trajectoryset\real$, the same combination operators (sum or product) apply to the combination of multiple-shooting likelihood functions $p_{ms}$ analogously.}

The consequence of using these likelihoods can be seen in \autoref{fig:combination} where \autoref{fig:double-pendulum-prod} shows the \rev{resulting posterior distribution (in parameter space)} when treating a set of trajectories as independent and taking the product of their likelihoods (\autoref{eq:traj_likelihood_prod}), while \autoref{fig:double-pendulum-sum} shows the result of treating them as a sum of Gaussian likelihoods (\autoref{eq:traj_likelihood_sum}).
In \autoref{fig:double-pendulum-prod} the posterior becomes the average of the two distributions since that is the most likely position that generated both of the distinct trajectories. \rev{In contrast, the posterior approximated by the same algorithm (CSVGD) but using the sum of Gaussian likelihoods, successfully captures the multimodality in the trajectory space since most particles have aligned near the two modes of the true distribution in parameter space.}

\subsection{State and Parameter Normalization}
\label{sec:normalization}
The parameters we are estimating are valid over only particular ranges of values. These ranges are often widely different - in the case of our real-world pendulum experiment, the center of mass of a link in a pendulum may be in the order of centimeters, while the angular velocity at the beginning of the recorded motion can reach values on the orders of meters per second. It is therefore important to scale the parameters to a common range to avoid any dimension to dominate smaller parameter ranges during the estimation.

Similarly, the state dimensions are of different units - for example, we typically include velocities and positions in the recorded data over which we compute the likelihood. Therefore, we also normalize the range over the state dimensions. Given the state vector, respective parameter vector, $w$, we normalize each dimension $i$ by its statistical variance $\sigma^2$, i.e. $\nicefrac{w_i}{\sigma_i^2}$.


\subsection{KNN-based Approximation for KL Divergence}
In this work, we compare a set of parameter guesses (particles) to the ground-truth parameters, or a set of trajectories generated by simulating trajectories from each parameter in the particle distribution to a set of trajectories created on a physical system.
To compare these distributions, we use the KL divergence to determine how the two distributions differ from each other.
Formally, the KL divergence is the expected value of the log likelihood ratio between two distributions, \rev{and is an asymmetric divergence that} does not satisfy the triangle inequality.

The KL divergence is easily computable in the case of discrete distributions or simple parametric distributions, but is not easily calculable for samples from non-parametric distributions such as those over trajectories.
Instead, we use an approximation to the KL divergence which uses the relative distances between samples in a set to estimate the KL divergence between particle distributions.
This method has been used to compare particle distributions over robot poses to asses the performance of particle filter distributions~\cite{chou_performance_2011}.
To estimate the KL divergence between particle distributions over trajectories $\trajectoryset^{p}$ and $\trajectoryset^{q}$ we adopt the formulation from~\cite{wang_divergence_2009, chou_performance_2011}:
\begin{align}\label{eq:knn-kl}
    \tilde{d}_{\text{KL}} (\trajectoryset^{p} \parallel \trajectoryset^{q}) &= \frac{\observationdim}{|\trajectoryset^p|} \sum_{i=1}^{|\trajectoryset^p|} \log\frac{\operatorname{KNN}^p_{k_i}(i)}{\operatorname{KNN}^q_{l_i}(i)} \\ \nonumber
    &+   \frac{1}{|\trajectoryset^p|} \sum_{i=1}^{|\trajectoryset^p|} [\psi(l_i) - \psi(k_i)] \\\nonumber
    &+ \log\frac{|\trajectoryset^q|}{|\trajectoryset^p|-1},
\end{align}
where $\observationdim$ is the dimensionality of the trajectories,  $|\trajectoryset^p|$ is the number of trajectories in the $\trajectoryset^{p}$ dataset, $|\trajectoryset^q|$ is the number of particles in the $\trajectoryset^{q}$ dataset, $\operatorname{KNN}^p_{k_i}(i)$ is the distance from trajectory $\trajectory_i \in \trajectoryset^{p}$ to its $k_i$-th nearest neighbor in $\trajectoryset^{q}$, $\operatorname{KNN}^q_{l_i}(i)$ is the distance from trajectory $\trajectory_i \in \trajectoryset^{p}$ to its $l_i$-th nearest neighbor in $\trajectoryset^{p} \backslash \trajectory_i$, and $\psi$ is the digamma function. Note that this approximation of KL divergence can also be applied to compare parameter distributions, as we show in the synthetic data experiment from \autoref{sec:exp-param-noise} (cf. \autoref{fig:exp-param-noise-metrics-kl-real-sim} and \autoref{fig:exp-param-noise-metrics-kl-sim-real}) where the ground-truth parameter distribution is known.

Throughout this work, we set $k_i$ and $l_i$ to 3 as this reduces the bias in the approximation, but does not require a large amount of samples from the ground-truth distribution.

\section{Experiments}
In the following, we provide further technical details and results from the experiments we present in the main paper.

\subsection{Differentiable Simulator}
\label{sec:dynamics}
Other than requiring a differentiable forward dynamics model which allows to simulate the system in its entirety following the Markov assumption, our proposed algorithm does not rely on a particular choice of dynamical system or simulator for which its parameters need to be estimated.
For our experiments, we use the Tiny Differentiable Simulator~\cite{heiden2021neuralsim} that implements end-to-end differentiable contact models and the Articulated Body Algorithm (ABA)~\cite{featherstone2007rbda} to compute the forward dynamics (FD) for articulated rigid-body mechanisms. Given joint positions $\jointpos$, velocities $\jointvel$, torques $\jointtorque$ in generalized coordinates, and external forces $\externalforce$, ABA calculates the joint accelerations $\jointacc$.
We use semi-implicit Euler integration to advance the system dynamics in time for a time step $\Delta t$:
\begin{align}
    \label{eq:dynamics}
    \jointacc_{t+1} &= \operatorname{ABA}(\jointpos_t, \jointvel_t, \jointtorque_t, \externalforce_t; \params), \\ \nonumber
    \jointvel_{t+1} &= \jointvel_t + \jointacc_{t+1} \timestep,   \\ \nonumber
    \jointpos_{t+1} &= \jointpos_t + \jointvel_{t+1} \timestep.
\end{align}

The second-order ODE described by \autoref{eq:dynamics} is lowered to a first-order system, with state $\statevec_t = \bmat{\jointpos_t & \jointvel_t}$. Furthermore, we deal primarily with the discrete time-stepped dynamics function
$\statevec_{t+1} = \fstep(\statevec_t, t, \params)$,
assuming that $\timestep$ is constant. The function $\fsim(\params, \statevec_0)$ uses $\fstep$ iteratively to produce a trajectory of states $[\statevec]_{t=1}^T$ given an initial state $\statevec_0$ and the parameters $\params$. Many systems of practical interest for robotics are controlled by an external input. In our formulation for parameter estimation, we include controls as explicit dependencies on the time parameter $t$.

For an articulated rigid body system, the parameters $\params$ may include (but are not limited to) the masses, inertial properties and geometrical properties of the bodies in the mechanism, as well as joint and contact friction coefficients. Given $\frac{\partial \fstep}{\partial\params}$ and $\frac{\partial\fstep}{\partial\statevec}$, gradients of simulation parameters with respect to the state trajectories can be computed directly through the chain rule, or via the adjoint sensitivity method~\cite{pontryagin2018mathematical}.


\begin{figure}[t]
    \centering

    \newcommand{\subfigwidth}{7cm}
    \begin{subfigure}[b]{\subfigwidth}
        \centering
        \resizebox{7cm}{!}{
            \begin{tabular}{llS[table-format=5.3]@{\,}s[table-unit-alignment = left]S[table-format=5.3]@{\,}s[table-unit-alignment = left]}
                \toprule
                \bf Link & \bf Parameter  & \multicolumn{2}{c}{\bf Minimum} & \multicolumn{2}{c}{\bf Maximum}                           \\
                \midrule
                Link 1   & Mass           & 0.05                            & \si{\kg}                        & 0.5 & \si{\kg}          \\
                         & $I_{xx}$       & 0.002                           & \si{\kg.\meter^2}               & 1.0 & \si{\kg.\meter^2} \\
                         & COM $x$        & -0.2                            & \si{\meter}                     & 0.2 & \si{\meter}       \\
                         & COM $y$        & -0.2                            & \si{\meter}                     & 0.2 & \si{\meter}       \\
                         & Joint friction & 0.0                             &                                 & 0.5                     \\
                \midrule\addlinespace[.5em]
                Link 2   & Length         & 0.08                            & \si{\meter}                     & 0.3 & \si{\meter}       \\
                         & Mass           & 0.05                            & \si{\kg}                        & 0.5 & \si{\kg}          \\
                         & $I_{xx}$       & 0.002                           & \si{\kg.\meter^2}               & 1.0 & \si{\kg.\meter^2} \\
                         & COM $x$        & -0.2                            & \si{\meter}                     & 0.2 & \si{\meter}       \\
                         & COM $y$        & -0.2                            & \si{\meter}                     & 0.2 & \si{\meter}       \\
                         & Joint friction & 0.0                             &                                 & 0.5                     \\
                \bottomrule                                                                                                             \\
            \end{tabular}
        }
        \caption{Parameters to be estimated. $I$ refers to the $3\times3$ inertia matrix, COM stands for center of mass.}
        \label{tab:params-ibm-pendulum}
    \end{subfigure}\hfill
    \renewcommand{\subfigwidth}{5cm}
    \begin{subfigure}[b]{\subfigwidth}
        \centering
        \includegraphics[width=\subfigwidth,trim=0 0 0 4.5cm,clip]{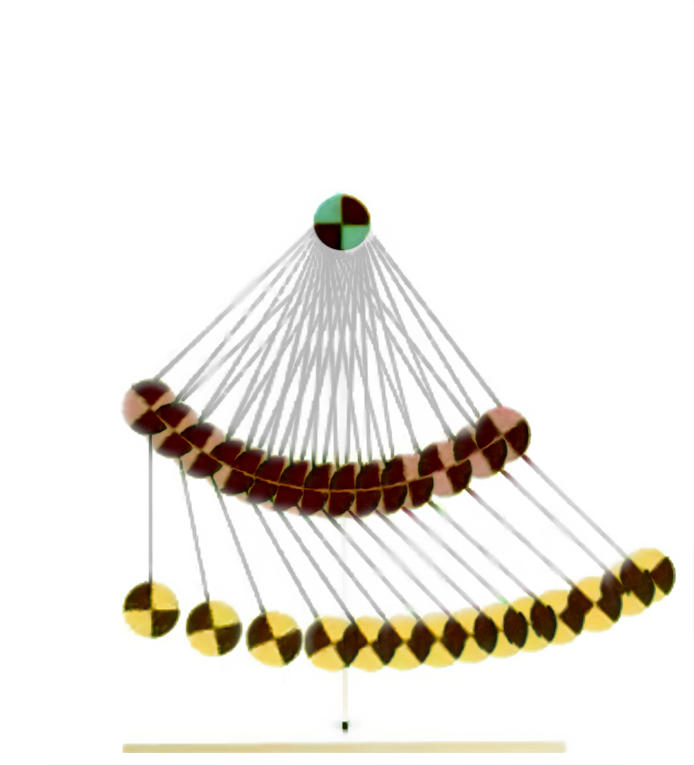}
        \caption{Time lapse of a double pendulum trajectory from the IBM dataset~\cite{asseman2018learning}.}
        \label{tab:dataset-ibm-pendulum}
    \end{subfigure}
    \caption{Physical double pendulum experiment from \autoref{sec:ibm-pendulum}.}
\end{figure}

\subsection{Identify Real-world Double Pendulum}
\label{sec:ibm-pendulum}
We set up the double pendulum estimation experiment with the 11 parameters shown in \autoref{tab:params-ibm-pendulum} to be estimated. The state space consists of the two positions and velocities of both joints: $\statevec = \bmat{\mathbf{q}_{0:1} & \mathbf{\dot{q}}_{0:1}}$. The dataset of trajectories contains image sequences (see time lapse of an excerpt from a trajectory in \autoref{tab:dataset-ibm-pendulum}) and annotated pixel coordinates of the three vertices in the double pendulum, from which we extracted joint positions and velocities (via finite differencing given the known recording frequency of \SI{400}{\Hz}).

\begin{figure}[t]
    \centering
    \newcommand{\figwidth}{0.8\columnwidth}
    \begin{tabular}{m{1.2cm}m{\figwidth}}
        Emcee &
        \includegraphics[width=\figwidth]{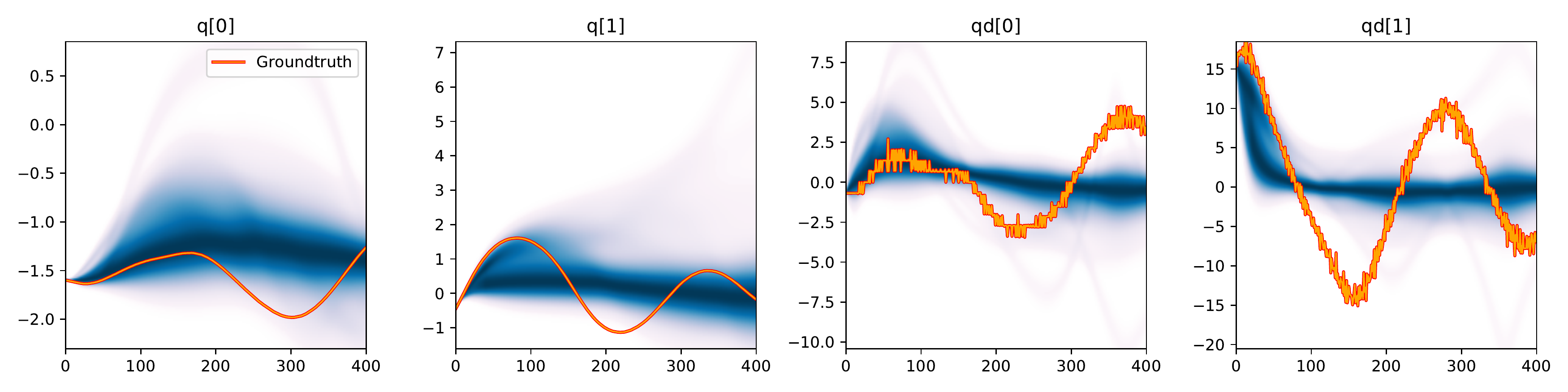}      \\
        CEM   &
        \includegraphics[width=\figwidth]{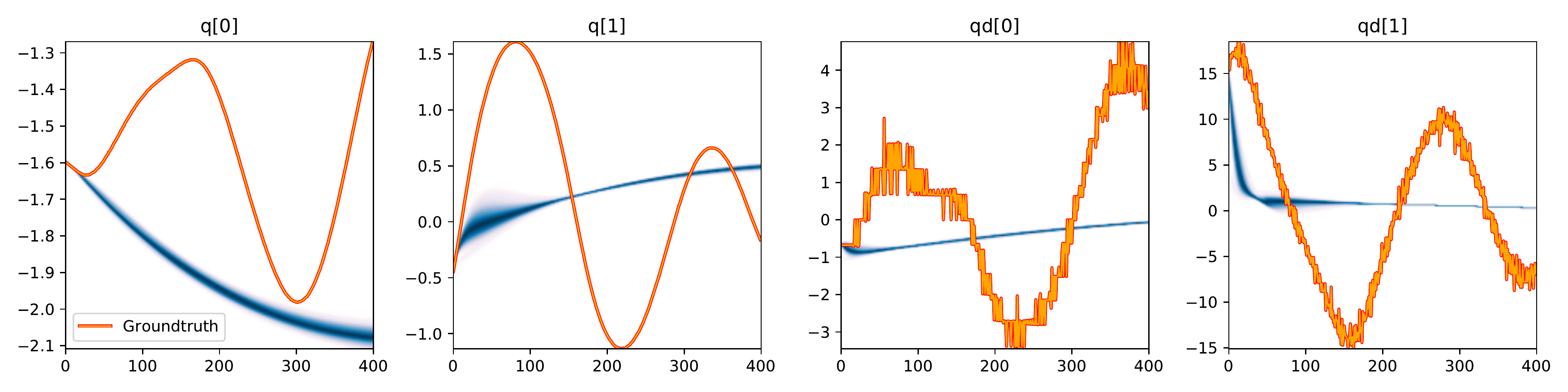}       \\
        NUTS  &
        \includegraphics[width=\figwidth]{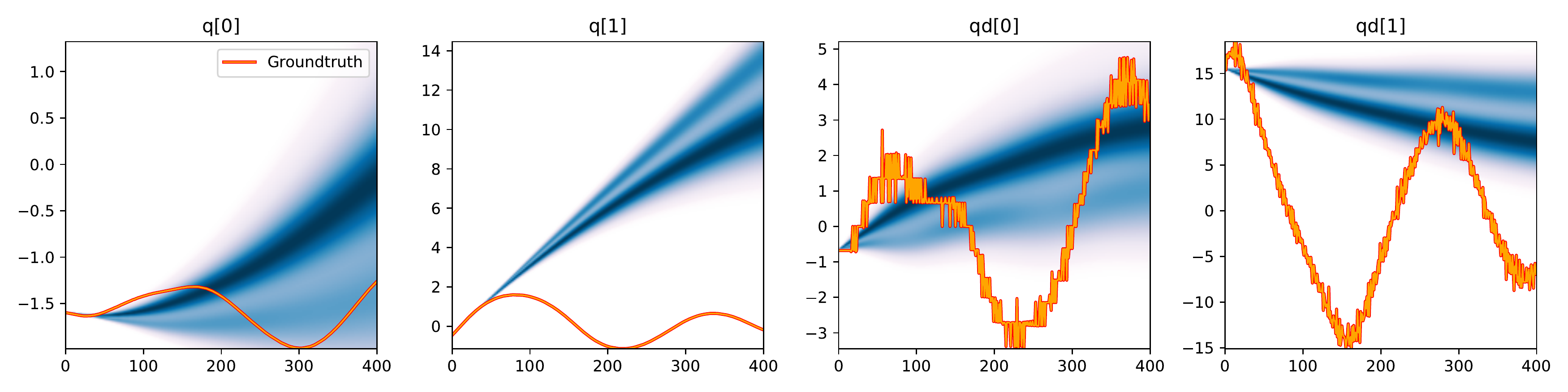}      \\
        SGLD  &
        \includegraphics[width=\figwidth]{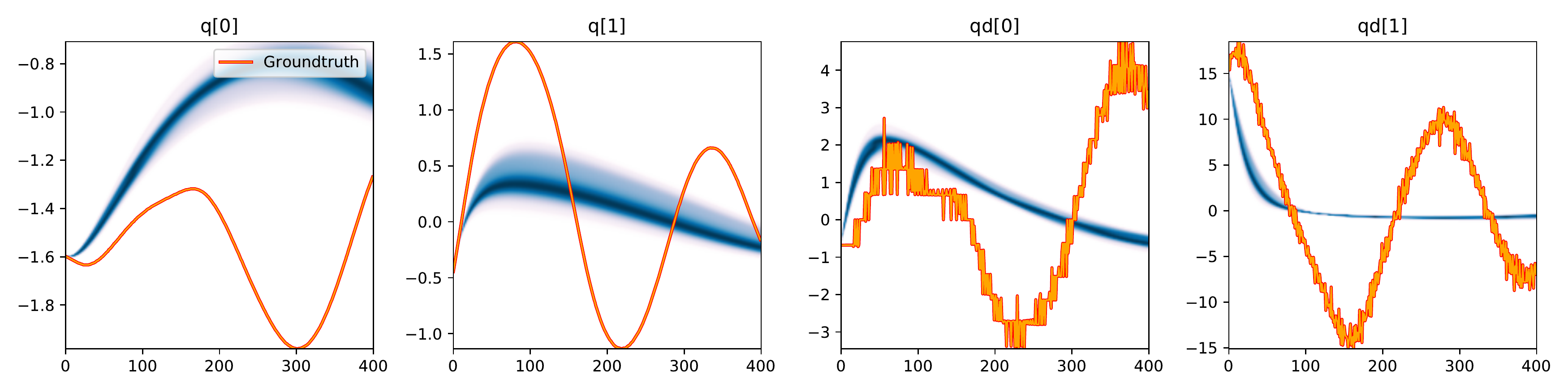}      \\
        SVGD  &
        \includegraphics[width=\figwidth]{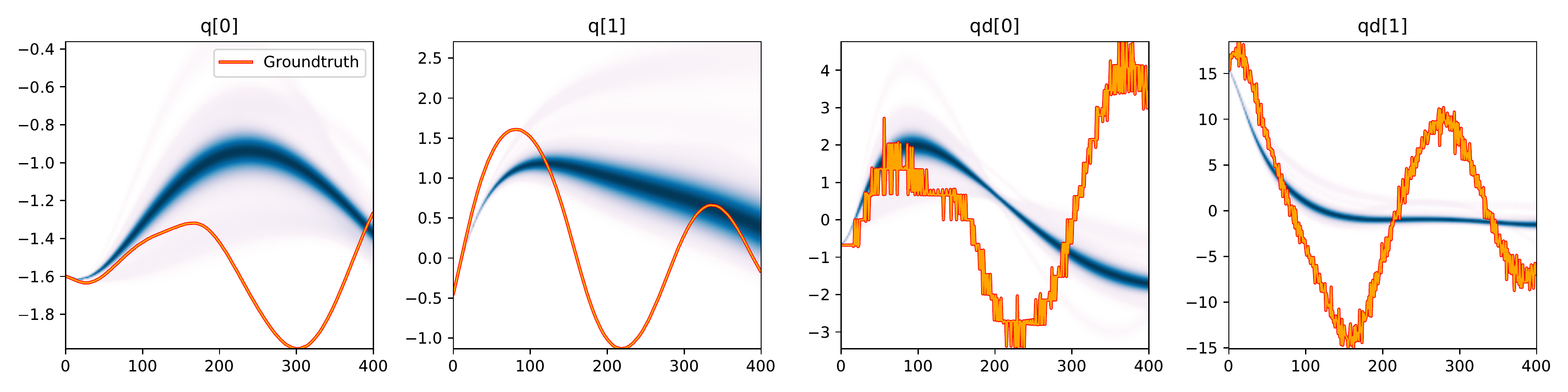}   \\
        CSVGD &
        \includegraphics[width=\figwidth]{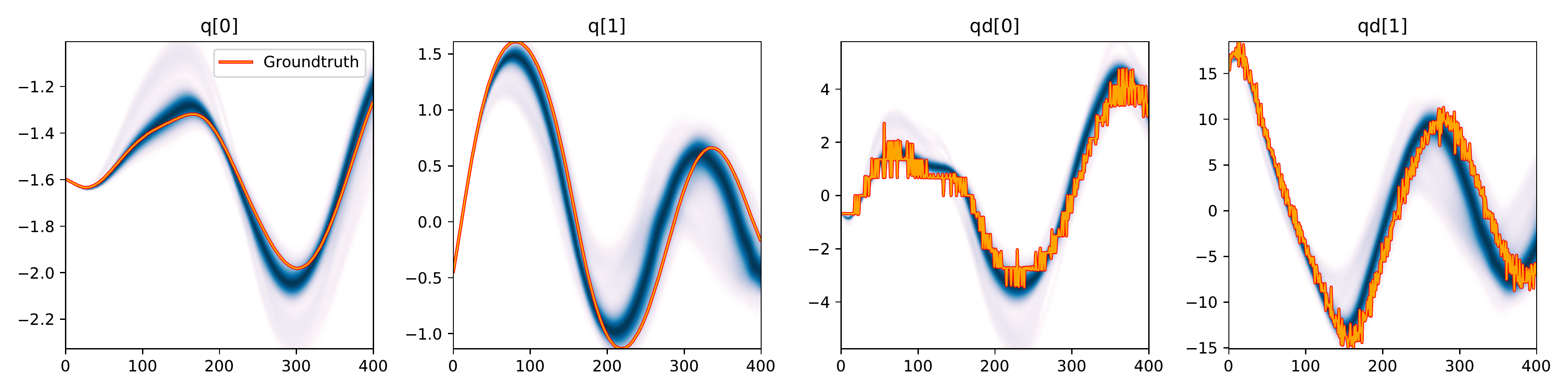} \\
    \end{tabular}
    \caption{Kernel density estimation over trajectory roll-outs from the last estimated 100 parameter guesses of each method, applied to the physical double pendulum dataset from \autoref{sec:ibm-pendulum}. The ground-truth trajectory here stems from the test dataset of 10 trajectories that were held out during training. The particle-based approaches (CEM, SVGD, CSVGD) use 100 particles.}
    \label{fig:ibm-pendulum-rollouts}
\end{figure}

Since we know that all trajectories in this dataset stem from the same double pendulum~\cite{asseman2018learning}, we only used a single reference trajectory as target trajectory $\trajectory\real$ during the estimation. We let each estimator run for 2000 iterations. For evaluation, we calculate the consistency metrics from \autoref{tab:system-accuracy} over 10 held-out trajectories from a test dataset. For comparison, we visualize the trajectory density over simulations rolled out from the last 100 Markov samples (or 100 particles in the case of particle-based approaches) in \autoref{fig:ibm-pendulum-rollouts}. The ground-truth shown in these plots again stems from the unseen test dataset. This experiment further demonstrates the generalizability of simulation-based inference, which, when an adequate model has been implemented and its parameters identified, can predict outcomes under various novel conditions even though the training dataset consisted of only a single trajectory in this example.

\subsection{Ablation Study on Multiple Shooting}
\label{sec:exp-multiple-shooting}

We evaluate the baseline estimation algorithms with our proposed multiple-shooting likelihood function (using 10 shooting windows) on the physical double pendulum dataset from before. To make the constrained optimization problem amenable to the MCMC samplers, we formulate the defect constraint through the likelihood defined in~\autoref{eq:likelihood-defect}
where we tune $\sigma^2_{\text{def}}$ to a small value (on the order of $10^{-2}$) such that the defects are minimized during the estimation. As we describe in \autoref{sec:multiple-shooting}, the parameter space is augmented by the shooting variables $\statevec_t^s$.

As shown in \autoref{tab:ms-baselines}, the MCMC approaches Emcee and NUTS do not benefit meaningfully from the multiple-shooting approach. Emcee often yields unstable simulations from which we are not able to compute some of the metrics. The increased dimensionality of the parameter space \rev{appears to add} a significant challenge to these methods, \rev{which are known to scale poorly to higher dimensions}. \rev{Despite being configured to use a Gaussian mixture model of 3 kernels, the} CEM posterior immediately collapses to a single point such that the KL divergence of simulated against real trajectories cannot be computed.

\begin{figure}[t]
    \centering
    \includegraphics[width=\columnwidth]{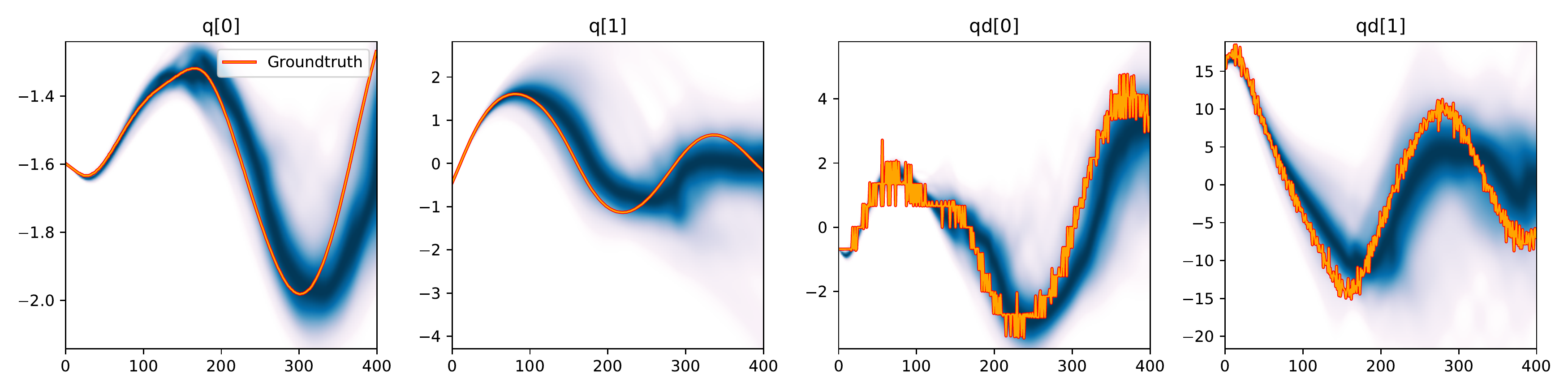}
    \caption{Kernel density estimation over trajectory roll-outs from the last estimated 100 parameter guesses of SGLD with the multiple-shooting likelihood model (see \autoref{sec:exp-multiple-shooting}), applied to the physical double pendulum dataset from \autoref{sec:ibm-pendulum}. Similarly to SVGD, SGLD benefits significantly from the smoother likelihood function while being able to cope with the augmented parameter space thanks to its gradient-based approach.}
    \label{fig:sgld-ms}
\end{figure}

We observe a significant improvement in estimation accuracy on SGLD, where the multiple-shooting approach allowed it to converge to closely matching trajectories, as shown in \autoref{fig:sgld-ms}. As with SVGD, the availability of gradients allows this method to scale better to the higher dimensional parameter space, while the smoothed likelihood landscape further helps the approach to find better fitting parameters.

\begin{table*}[]
    \centering
    \resizebox{0.65\textwidth}{!}{%
        \begin{tabular}{l|rr|rr|rr}
            \toprule
            & \multicolumn{2}{c|}{$d_{\text{KL}} (\trajectoryset\real \parallel \trajectoryset\simu)$}
            & \multicolumn{2}{c|}{$d_{\text{KL}} (\trajectoryset\simu \parallel \trajectoryset\real)$}
            & \multicolumn{2}{c}{\bf MMD} \\[0.5em]
            \bf Algorithm & \multicolumn{1}{c|}{\bf SS} & \multicolumn{1}{c|}{\bf MS} & \multicolumn{1}{c|}{\bf SS} & \multicolumn{1}{c|}{\bf MS} & \multicolumn{1}{c|}{\bf SS} & \multicolumn{1}{c}{\bf MS} \\\midrule
            \bf Emcee & \bf 8542.2466 & 8950.4574 & 4060.6312 & N/A & 1.1365 & N/A \\
            \bf CEM & 8911.1798 & \bf 8860.5115 & 8549.5927 & N/A & 0.9687 & \bf 0.5682 \\
            \bf SGLD & 8788.0962 & \bf 5863.2728 & 7876.0310 & \bf 2187.2825 & 2.1220 & \bf 0.0759 \\
            \bf NUTS & 9196.7461 & \bf 8785.5326 & 6432.2131 & \bf 4935.8983 & \bf 0.5371 & 1.1642 \\
            \bf (C)SVGD & 8803.5683  & \bf 5204.5336 & 10283.6659 & \bf 2773.1751 & 0.7177 & \bf 0.0366 \\
            \bottomrule 
        \end{tabular}
    }\vspace*{1em}
    \caption{Consistency metrics of the posterior distributions approximated from the physical double pendulum dataset (\autoref{sec:ibm-pendulum}) by the different estimation algorithms using the single-shooting likelihood $p_{ss}(\trajectory\real | \params)$ (column ``SS'') and the multiple-shooting likelihood $p_{ms}(\trajectory\real | \params)$ (column ``MS'') with 10 shooting windows. Note that SVGD with multiple-shooting corresponds to CSVGD.}
    \label{tab:ms-baselines}\vspace*{-1em}
\end{table*}

\subsection{Comparison to Likelihood-free Inference}
\label{sec:likelihoodfree}
Our \revtwo{Bayesian inference} approach \revtwo{leverages the simulator as part of the likelihood model to approximate posterior distributions over simulation parameters, which means the simulator is indispensable in our estimation process}. In the following, we compare our approach against the likelihood-free inference approach BayesSim~\cite{ramos_bayessim_2019} that leverages approximate Bayesian computation (ABC) which is the most popular family of algorithms within likelihood-free methods.

Likelihood-free methods assume the simulator is a black box that can generate a trajectory given a parameter setting. Instead of querying the simulator to evaluate the likelihood (as in our approach), a conditional density $q(\params | \trajectoryset)$ is learned directly from a dataset of simulation parameters and their corresponding rolled-out trajectories via supervised learning to approximate the posterior. A common choice of model for such density is a mixture density network~\cite{bishop1994mixture}, which parameterizes a Gaussian mixture model. This is in contrast to \revtwo{our} (C)SVGD \revtwo{algorithm} which can approximate any shape of posterior by being a nonparametric inference algorithm.

For our experiments we collect a dataset of 10,000 simulated trajectories of parameters randomly sampled from the prior distribution. We train the density network via the Adam optimizer with a learning rate of $10^{-3}$ for 3000 epochs, after which we observed no meaningful improvement to the calculated log-likelihood loss during training. In the following, we provide further details on the likelihood-free inference pipeline we consider, by describing the input data processing and the model used for approximating the posterior.

\subsubsection{Input Data Processing}
\label{sec:bayessim-input}
\begin{figure*}[]
\newcommand{\subfigwidth}{3.4cm}
\newcommand{\subfigimgwidth}{3.8cm}
\centering
    \begin{subfigure}[b]{\subfigwidth}
        \centering
        \hspace*{-0.8cm}
        \includegraphics[width=\subfigimgwidth]{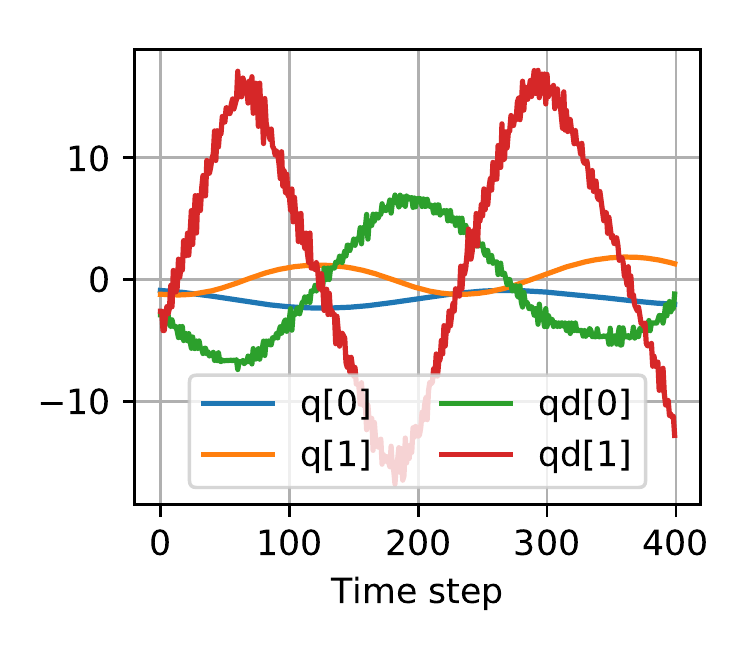}
        \caption{Raw Input}
    \end{subfigure}
    \begin{subfigure}[b]{\subfigwidth}
        \centering
        \hspace*{-2em}
        \includegraphics[width=\subfigimgwidth]{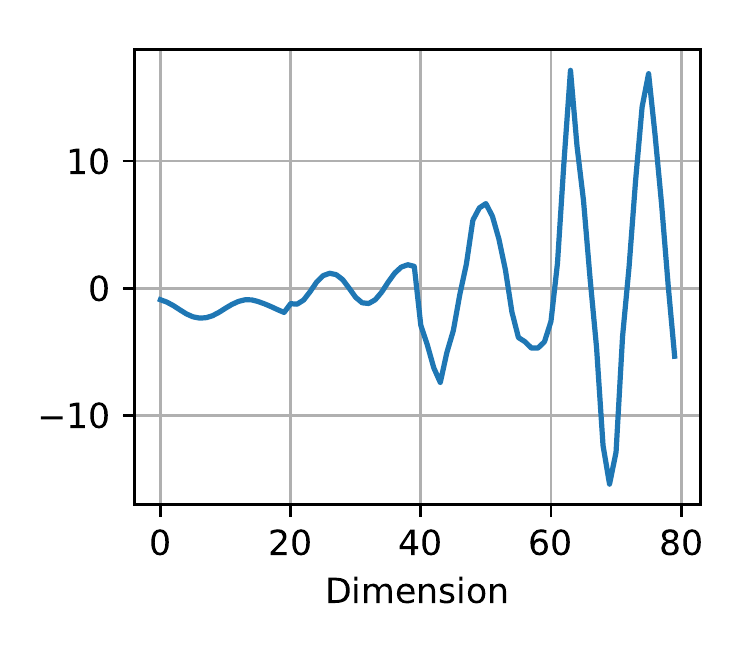}
        \caption{Downsampled}
    \end{subfigure}
    \begin{subfigure}[b]{\subfigwidth}
        \centering
        \hspace*{-1.5em}
        \includegraphics[width=\subfigimgwidth]{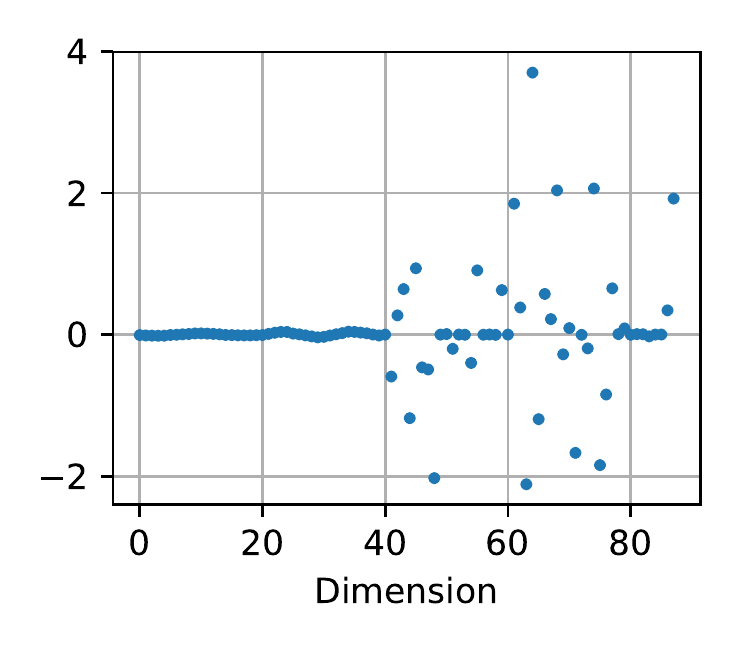}
        \caption{Difference}
    \end{subfigure}
    \begin{subfigure}[b]{\subfigwidth}
        \centering
        \hspace*{-1em}
        \includegraphics[width=\subfigimgwidth]{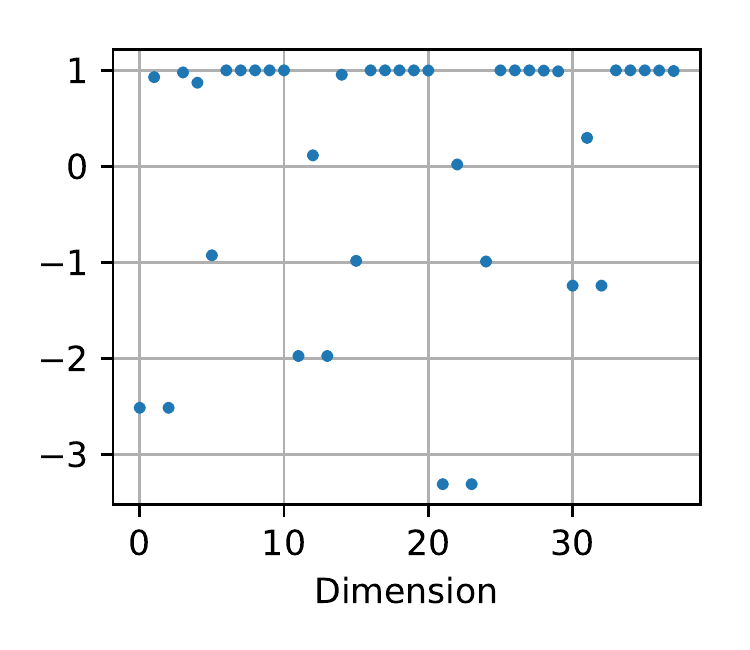}
        \caption{Summary}
    \end{subfigure}
    \begin{subfigure}[b]{\subfigwidth}
        \centering
        \includegraphics[width=\subfigimgwidth]{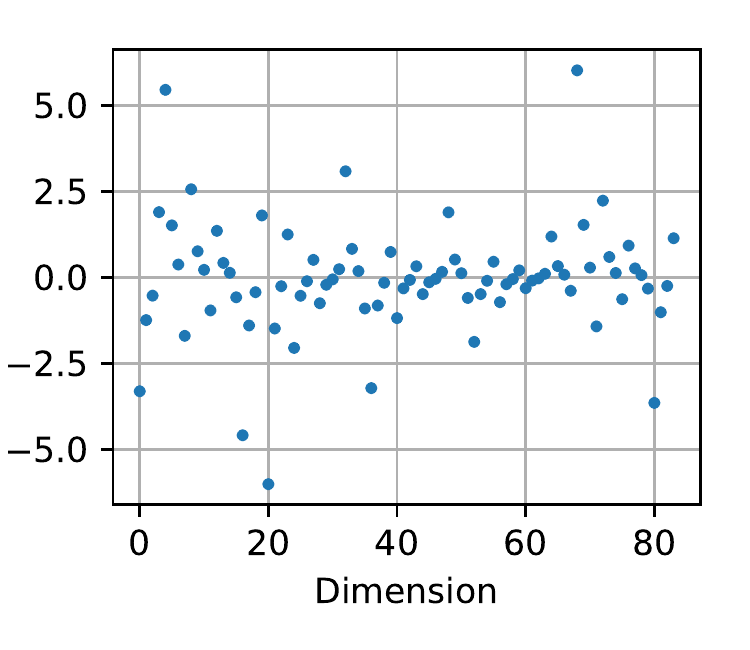}
        \caption{Signature}
    \end{subfigure}
    \caption{Exemplary visualization of the input processing methods for the likelihood-free baselines from \autoref{sec:likelihoodfree} applied to a trajectory from the double pendulum experiment in \autoref{sec:ibm-pendulum}.}
    \label{fig:bayessim-inputs}
\end{figure*}

The input to the learned density model has to be a sufficient statistic of the underlying data, while being low-dimensional in order to keep the learning problem computationally tractable. We consider the following \revtwo{four} methods of processing the trajectories that are the input to the likelihood-free methods, as visualized for an example trajectory in \autoref{fig:bayessim-inputs}. Note that we configured the following input processing methods to generate a one-dimensional input vector that has a reasonable length to be computationally feasible to train on (given the 10,000 trajectories from the training dataset), while achieving acceptable performance which we validated through testing various settings.

\paragraph{Downsampled} we down-sample the trajectory so that for the double pendulum experiment (\autoref{sec:ibm-pendulum}) we use only every 20th state, for the Panda arm experiment (\autoref{sec:panda-box}) only every 200-th state of the trajectory. Finally, the state dimensions per trajectory are concatenated to a one-dimensional vector.

\revtwo{\paragraph{Difference} we adapt the input statistic from the original BayesSim approach in \cite[Eq. (22)]{ramos_bayessim_2019} where the differences of two consecutive states along the trajectory are used in concatenation with their mean and variance:
\begin{align*}
\psi(\trajectory) &= (\operatorname{downsample}(\tau), \mathbb{E}[\tau], \operatorname{Var}[\tau]) \\
&\text{where}~~ 
\tau = \{\statevec_t - \statevec_{t-1}\}_{t=1}^T
\end{align*}
As before, we down-sample these state differences and concatenate them to a vector.}

\paragraph{Summary} for each state dimension of the trajectory, we compute the following statistics typical for time series: mean, variance, cross correlation between state dimensions of the trajectory, as well as auto-correlations for each dimension at 5 different time delays: [5, 10, 20, 50, 100] \revtwo{time steps}. These numbers \revtwo{are concatenated} for all state dimensions to a one-dimensional vector per input trajectory.
\paragraph{Signature} we compute the signature transform from the signatory package~\cite{kidger2021signatory} over the input trajectory. Such so-called path signatures have been recently introduced to extract information about order and area, thereby preserving features inherent to nonlinear trajectories. We select a depth for the signature transform of 3 for the double pendulum experiment, and 2 for the Panda arm experiment, \revtwo{to obtain feature vectors of comparable size to the aforementioned input techniques}.

\subsubsection{Density Model}
\label{sec:bayessim-model}
As the density model for the learned posterior $q(\params | \trajectoryset)$, we select the following commonly used representations.

\paragraph{Mixture density network (\textbf{MDN})} uses neural network features from a feed-forward neural network using two hidden layers with 24 units each.
\paragraph{Mixture density random Fourier features (\textbf{MDRFF})} this density model uses Fourier features and a kernel. We evaluate the MDRFF with the following common choices for the kernel:
    \begin{itemize}
        \item Radial Basis Function (\textbf{RBF}) kernel
        \item \textbf{Matérn} kernel~\cite[Equation (4.14)]{rasmussen2005gp} with $\nu=\nicefrac{5}{2}$
    \end{itemize}

\subsubsection{Evaluation}
Note that instead of action generation, which is part of the proposed BayesSim pipeline~\cite{ramos_bayessim_2019}, we only focus on the inference of the posterior density over simulation parameters in order to compare such likelihood-free inference approach against our method.

Finally, to evaluate the metrics shown in \autoref{tab:likelihoodfree-results} for each BayesSim instantiation (input method \revtwo{and} density model), we sample 100 parameter vectors from the learned posterior $q(\params | \trajectoryset)$ and simulate them to obtain 100 trajectories which are compared against the reference trajectory sets, as we did in the comparison for the other Bayesian inference methods in \autoref{tab:system-accuracy}. 

\begin{table*}[]
    \centering
    \resizebox{0.8\textwidth}{!}{%
        \begin{tabular}{cc||ccc|c}
            \toprule
            & & \multicolumn{3}{c|}{\bf Double Pendulum Experiment} & \bf Panda Arm Experiment \\
            \bf Input  & \bf Model & $d_{\text{KL}} (\trajectoryset\real \parallel \trajectoryset\simu)$ 
            & $d_{\text{KL}} (\trajectoryset\simu \parallel \trajectoryset\real)$
            & MMD
            & $\log\pobs(\trajectoryset\real \parallel \trajectoryset\simu)$ \\ \midrule
            Downsampled & MDN & 8817.9222 & 4050.4666 & 0.6748 & -17.4039 \\
            \revtwo{Difference} & \revtwo{MDN} & \revtwo{8919.2463} & \revtwo{4633.2637} & \revtwo{0.6285} & \revtwo{-17.1646} \\
            Summary & MDN & 9092.5575 & 5093.8851 & 0.5664 & -18.3481   \\
            Signature & MDN & 8985.8056 & 4610.5438 & 0.5807 & -19.3432    \\
            Downsampled & MDRFF (RBF) & 9027.9474 & 5091.5283 & 0.5593 & -17.2335 \\
            \revtwo{Difference} & \revtwo{MDRFF (RBF)} & \revtwo{8936.3823} & \revtwo{4282.8599} & \revtwo{0.5988} & \revtwo{-18.4892} \\
            Summary & MDRFF (RBF) & 9063.1753 & 4884.1398 & 0.5672 & -19.5430   \\
            Signature & MDRFF (RBF) & 8980.9080 & 4081.1160 & 0.6016 & -18.3458   \\
            Downsampled & MDRFF (Matérn) & 8818.1830 & 3794.9873 & 0.6110 & -17.6395   \\
            \revtwo{Difference} & \revtwo{MDRFF (Matérn)} & \revtwo{8859.2156} & \revtwo{4349.9971} & \revtwo{0.6176} & \revtwo{-17.2752} \\
            Summary & MDRFF (Matérn) & 8962.0501 & 4241.4551 & 0.5999 & -19.6672  \\
            Signature & MDRFF (Matérn) & 9036.9626 & 4620.9517 & 0.5715 & -18.1652  \\ \midrule
            \multicolumn{2}{c||}{\bf CSVGD} & \bf 5204.5336 & \bf 2773.1751 & \bf 0.0366 & \bf -15.1671 \\ 
            \bottomrule\\
        \end{tabular}
    }
    \caption{Consistency metrics of the posterior distributions approximated by the different BayesSim instantiations, where the input method and model name (including the kernel type for the MDRFF model) are given. Each metric is calculated across simulated and real trajectories. Lower is better on all metrics except the log-likelihood $\log\pobs(\trajectoryset\real \parallel \trajectoryset\simu)$ from the Panda arm experiment. For comparison, in the last row, we reproduce the numbers from CSVGD shown in \autoref{tab:system-accuracy}.}
    \label{tab:likelihoodfree-results}
\end{table*}

\begin{figure*}[]
    \centering
    \newcommand{\figwidth}{0.52\textwidth}
    \hspace*{-0.8cm}
    \begin{tabular}{cc}
        \includegraphics[width=\figwidth]{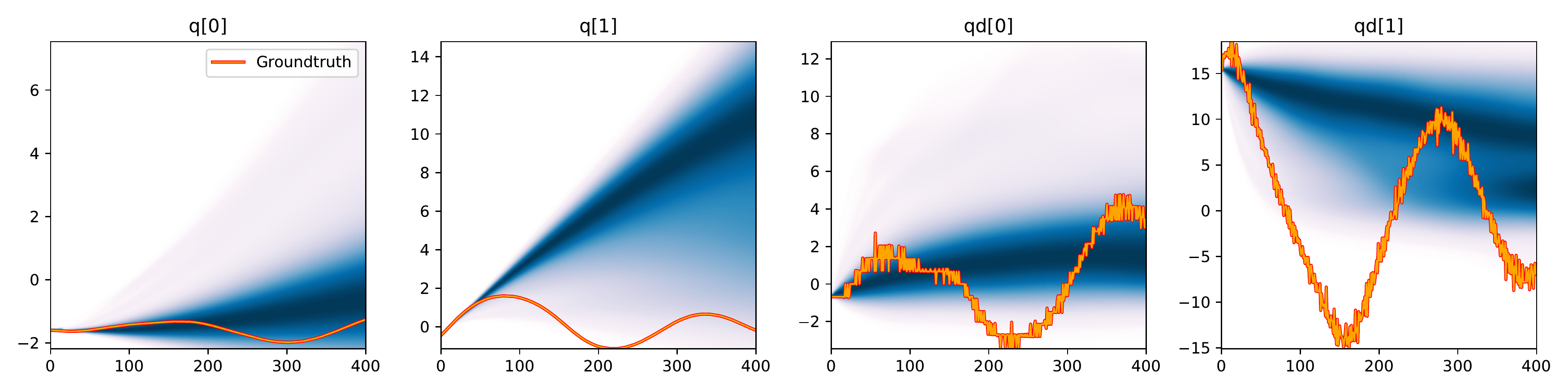}      &
        \includegraphics[width=\figwidth]{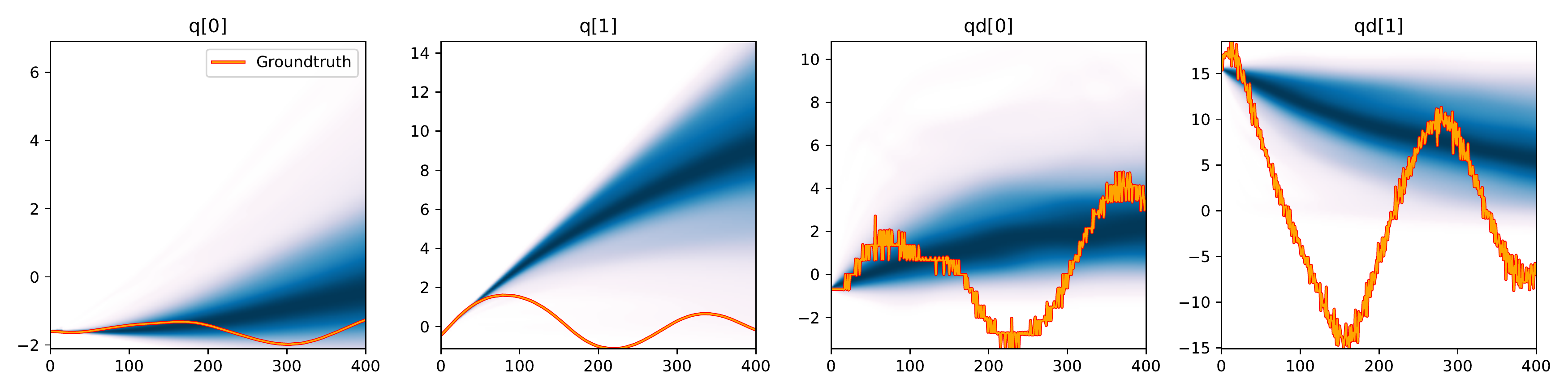}      \\
        \rev{\textbf{MDN} -- Downsampled} &
        \revtwo{\textbf{MDN} -- Difference} \\[1em]
        \includegraphics[width=\figwidth]{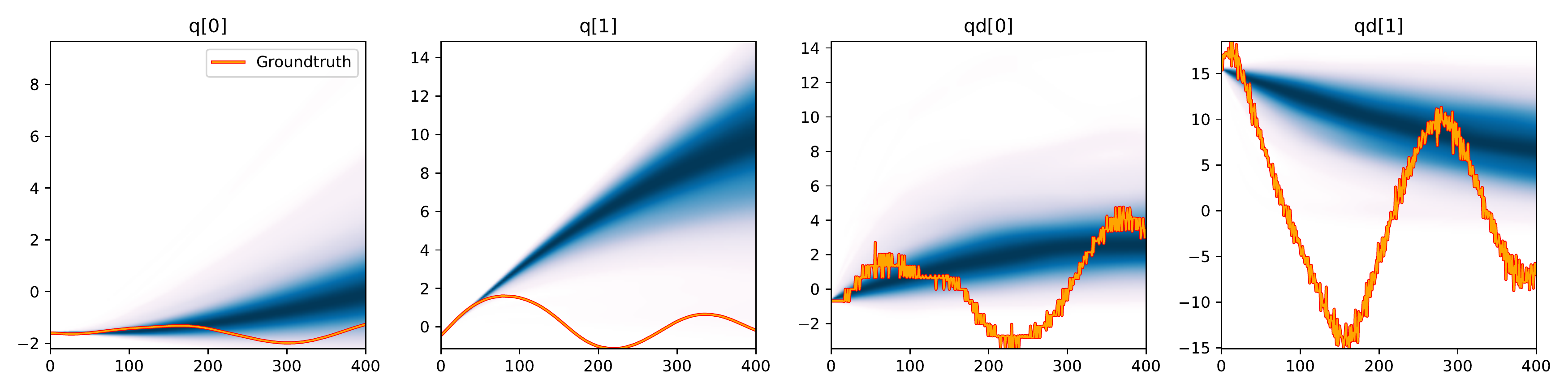}      &
        \includegraphics[width=\figwidth]{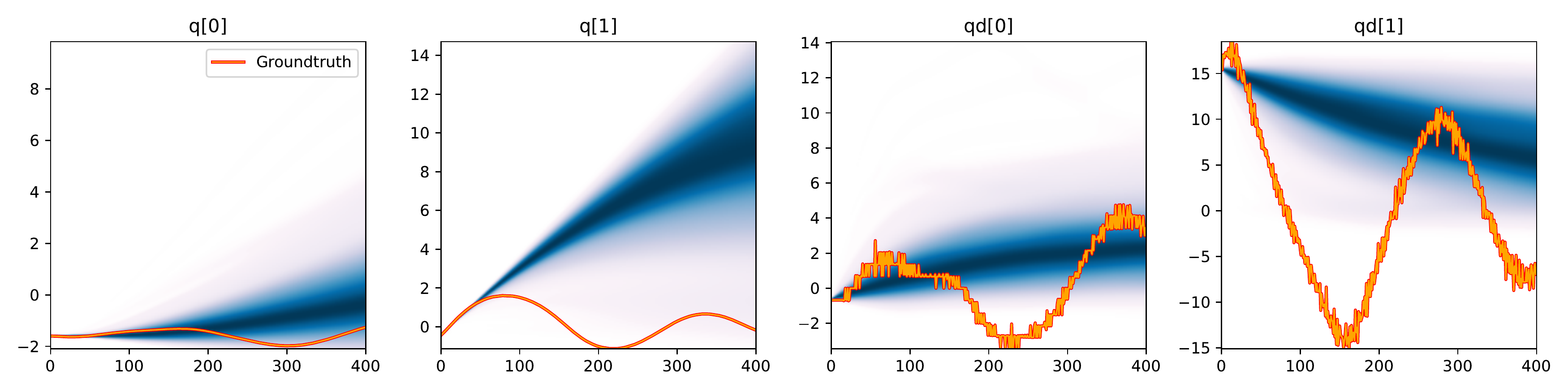}      \\
        \rev{\textbf{MDN} -- Summary} &
        \rev{\textbf{MDN} -- Signature} \\[1em]
        \includegraphics[width=\figwidth]{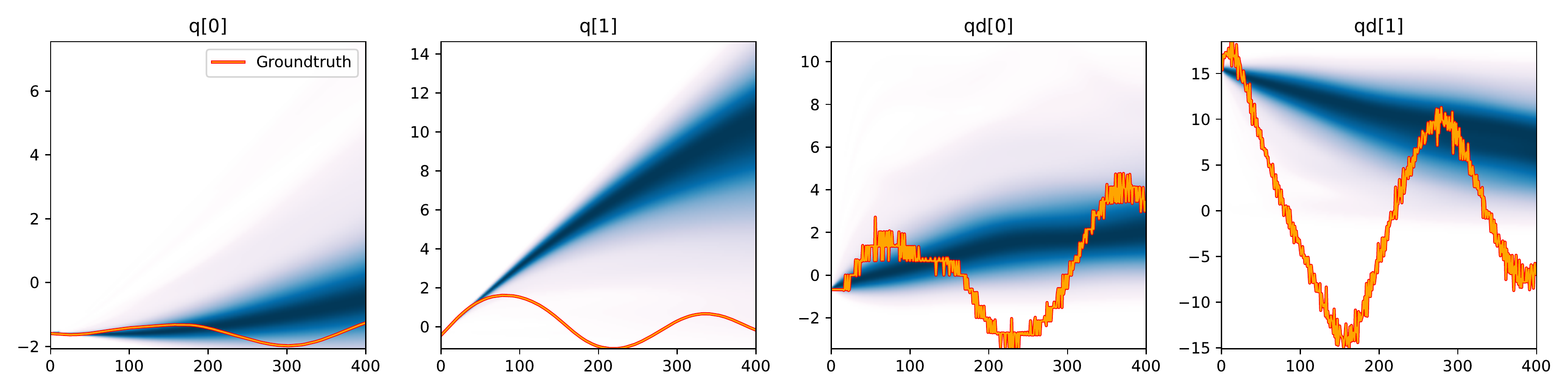}      &
        \includegraphics[width=\figwidth]{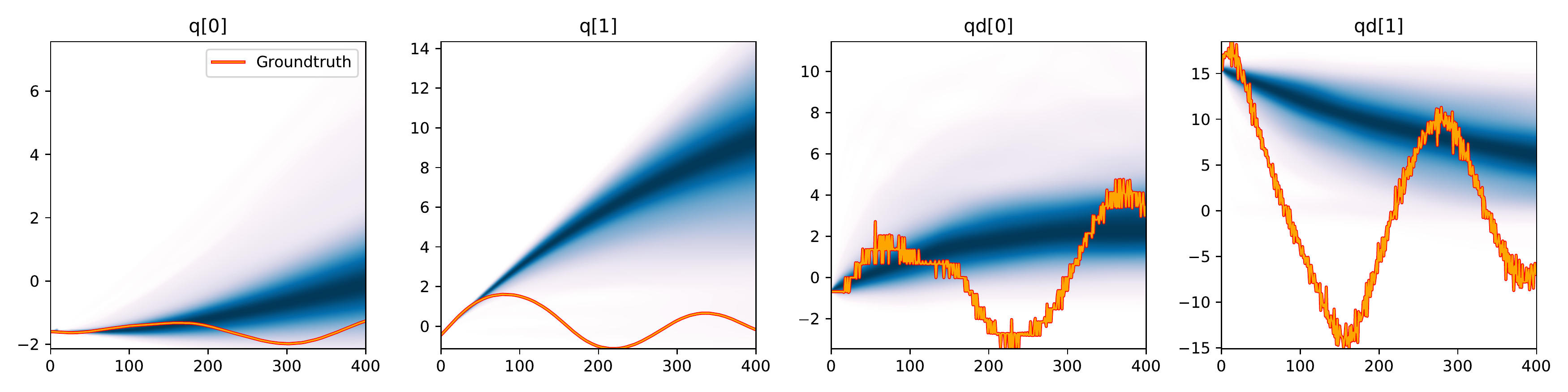}      \\
        \rev{\textbf{MDRFF (RBF)} -- Downsampled} &
        \revtwo{\textbf{MDRFF (RBF)} -- Difference} \\[1em]
        \includegraphics[width=\figwidth]{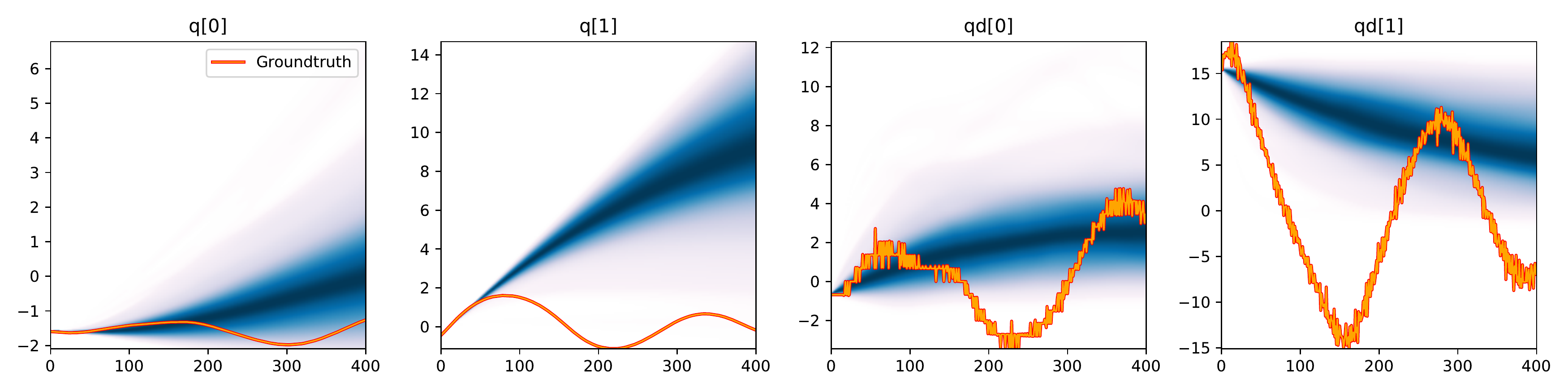}      &
        \includegraphics[width=\figwidth]{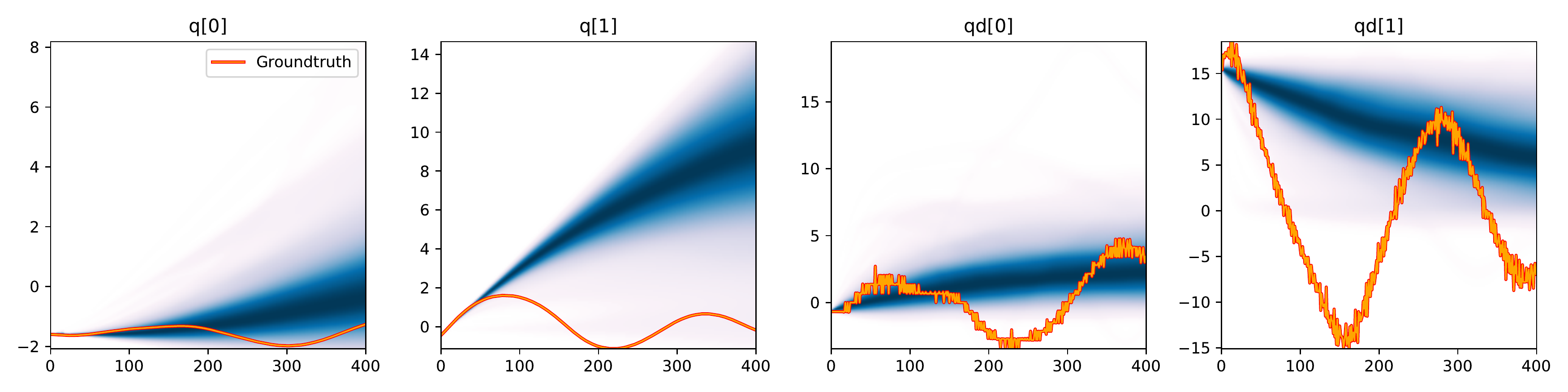}      \\
        \rev{\textbf{MDRFF (RBF)} -- Summary} &
        \rev{\textbf{MDRFF (RBF)} -- Signature} \\[1em]
        \includegraphics[width=\figwidth]{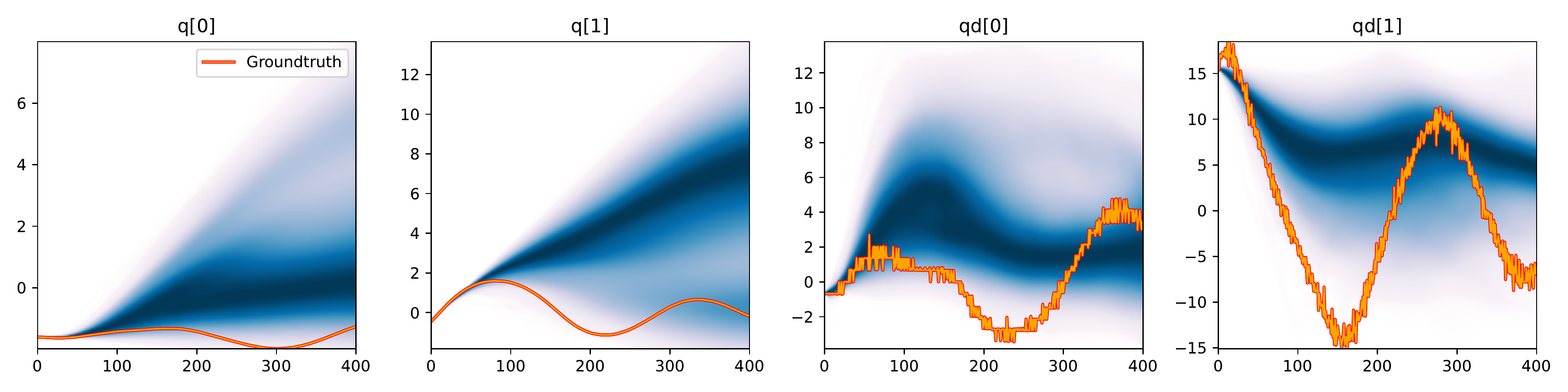}    &
        \includegraphics[width=\figwidth]{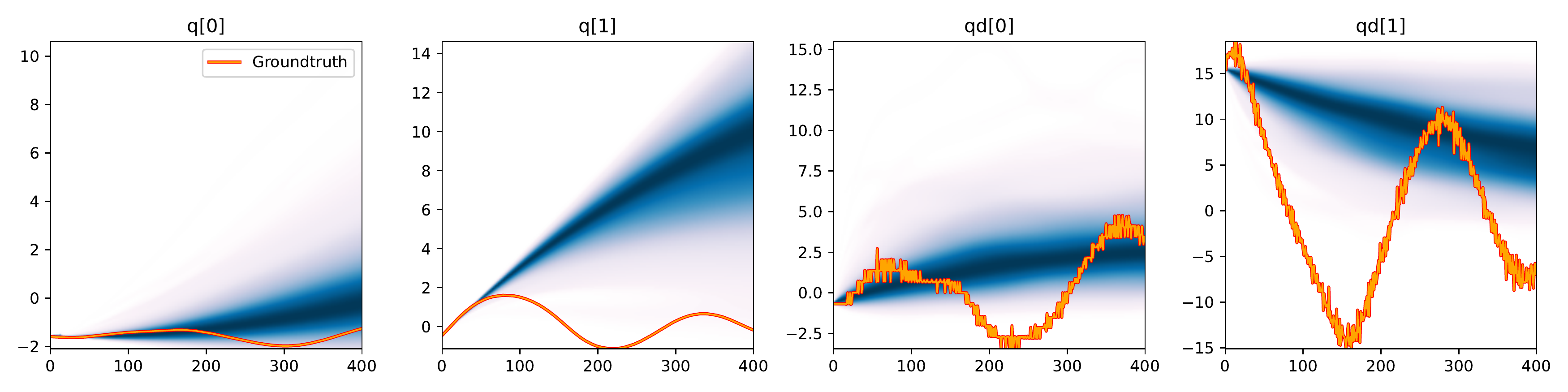}      \\
        \rev{\textbf{MDRFF (Matérn)} -- Downsampled} &
        \revtwo{\textbf{MDRFF (Matérn)} -- Difference} \\[1em]
        \includegraphics[width=\figwidth]{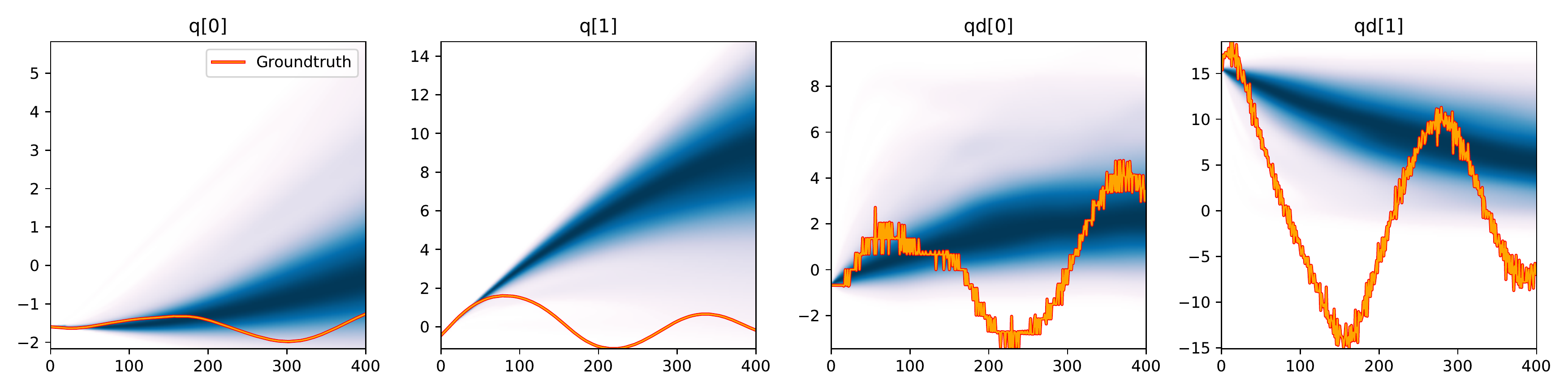}      &
        \includegraphics[width=\figwidth]{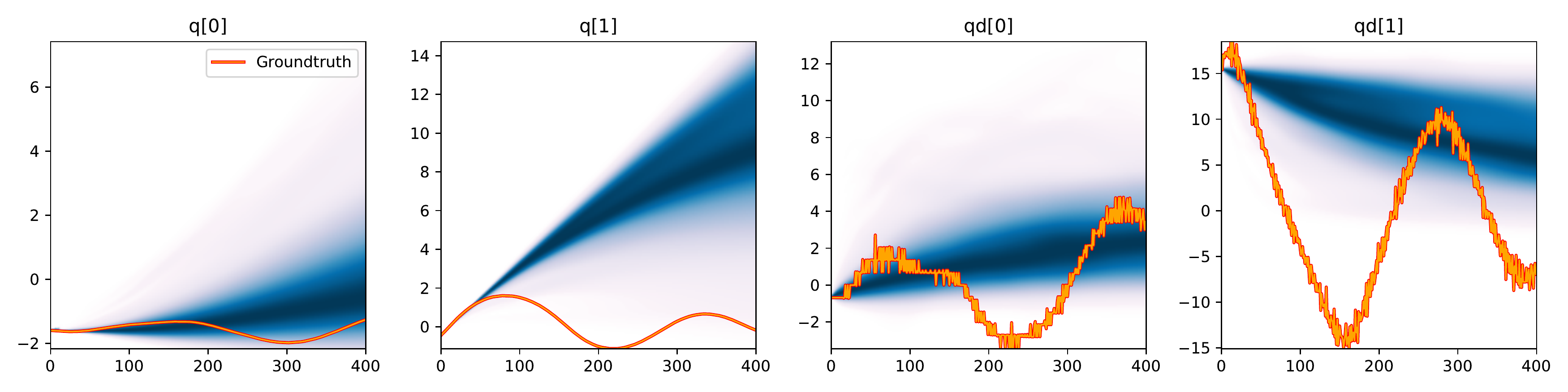}      \\
        \rev{\textbf{MDRFF (Matérn)} -- Summary} &
        \rev{\textbf{MDRFF (Matérn)} -- Signature}
    \end{tabular}\\[1em]
    \caption{Kernel density estimation over trajectory roll-outs from 100 parameter samples drawn from the posterior of each BayesSim method (model name with kernel choice in bold font + input method, see \autoref{sec:likelihoodfree}), applied to the physical double pendulum dataset from \autoref{sec:ibm-pendulum}. The ground-truth trajectory here stems from the test dataset of 10 trajectories that were held out during training.}
    \label{fig:ibm-pendulum-rollouts-bayessim}
\end{figure*}

\subsubsection{Discussion}
The results from our experiments with the various likelihood-free approaches in \autoref{tab:likelihoodfree-results} indicate that, among the tested pipelines, the MDRFF model with Matérn kernel and downsampled trajectory input overall performed the strongest, followed by the MDN with downsampled input. In comparison to the likelihood-based algorithms from \autoref{tab:system-accuracy}, these results are comparable on the double pendulum experiment. However, in comparison to CSVGD, the estimated likelihood-free posteriors are significantly less accurate, which can also be clearly seen in the density plots over the rolled out trajectories from such learned densities in \autoref{fig:ibm-pendulum-rollouts-bayessim}. On the Panda arm experiment, the likelihood-free methods are outperformed by the likelihood-based algorithms (such as the Emcee sampler) more often on the likelihood of the learned parameter densities. CSVGD again achieves a much more accurate posterior in this experiment than any likelihood-free approach.

\begin{figure*}
    \centering
    \textbf{\revtwo{BayesSim synthetic 2D inference experiment}}\\
    \newcommand{\figheight}{6.5cm}
    \begin{subfigure}[t]{0.36\textwidth}
        \centering
        \includegraphics[height=6cm,trim=0 0.5cm 0 0.5cm]{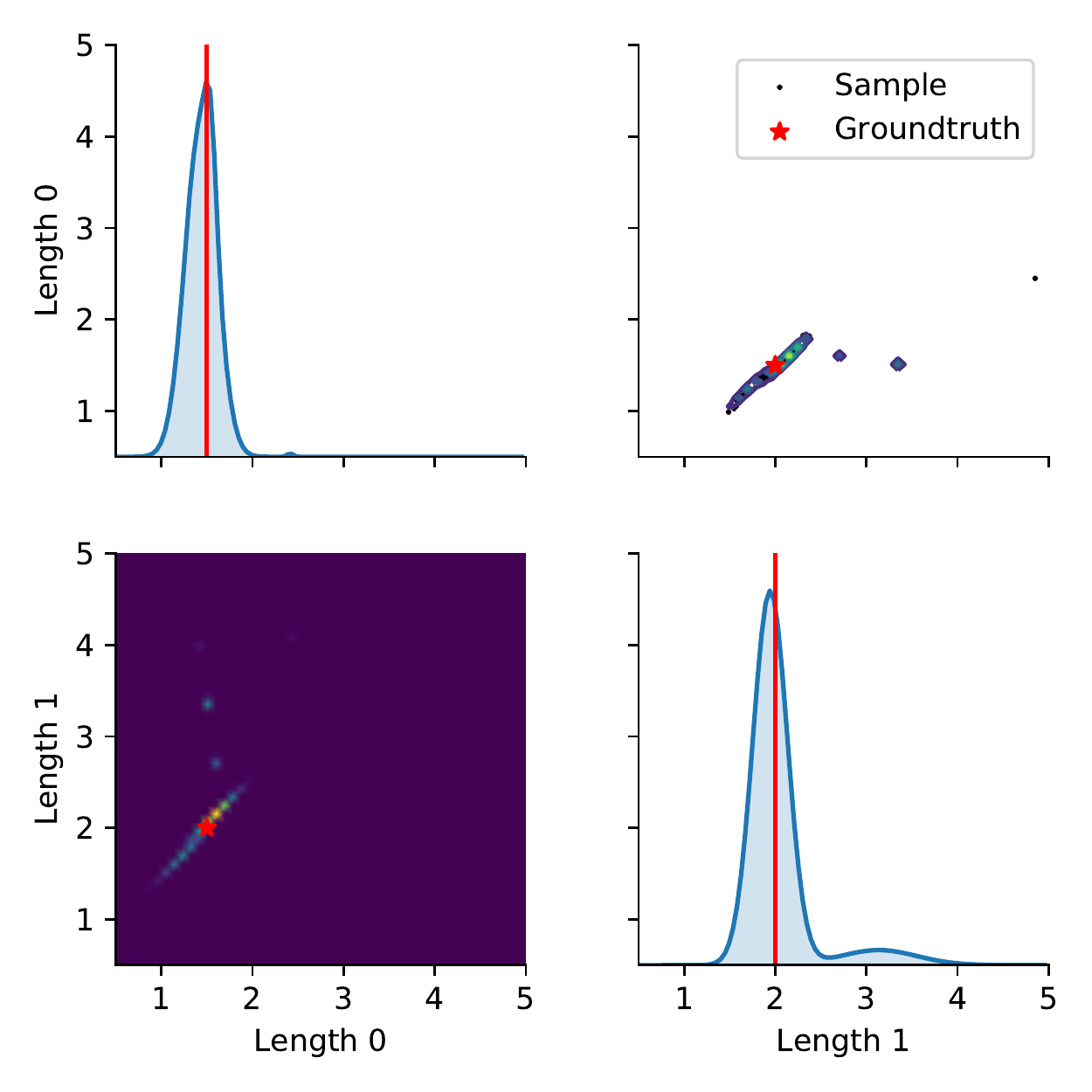}
        \caption{\revtwo{Posterior}}
        \label{fig:bayessim-twoparam-posterior}
    \end{subfigure}
    \begin{subfigure}[t]{0.36\textwidth}
        \centering
        \includegraphics[height=\figheight]{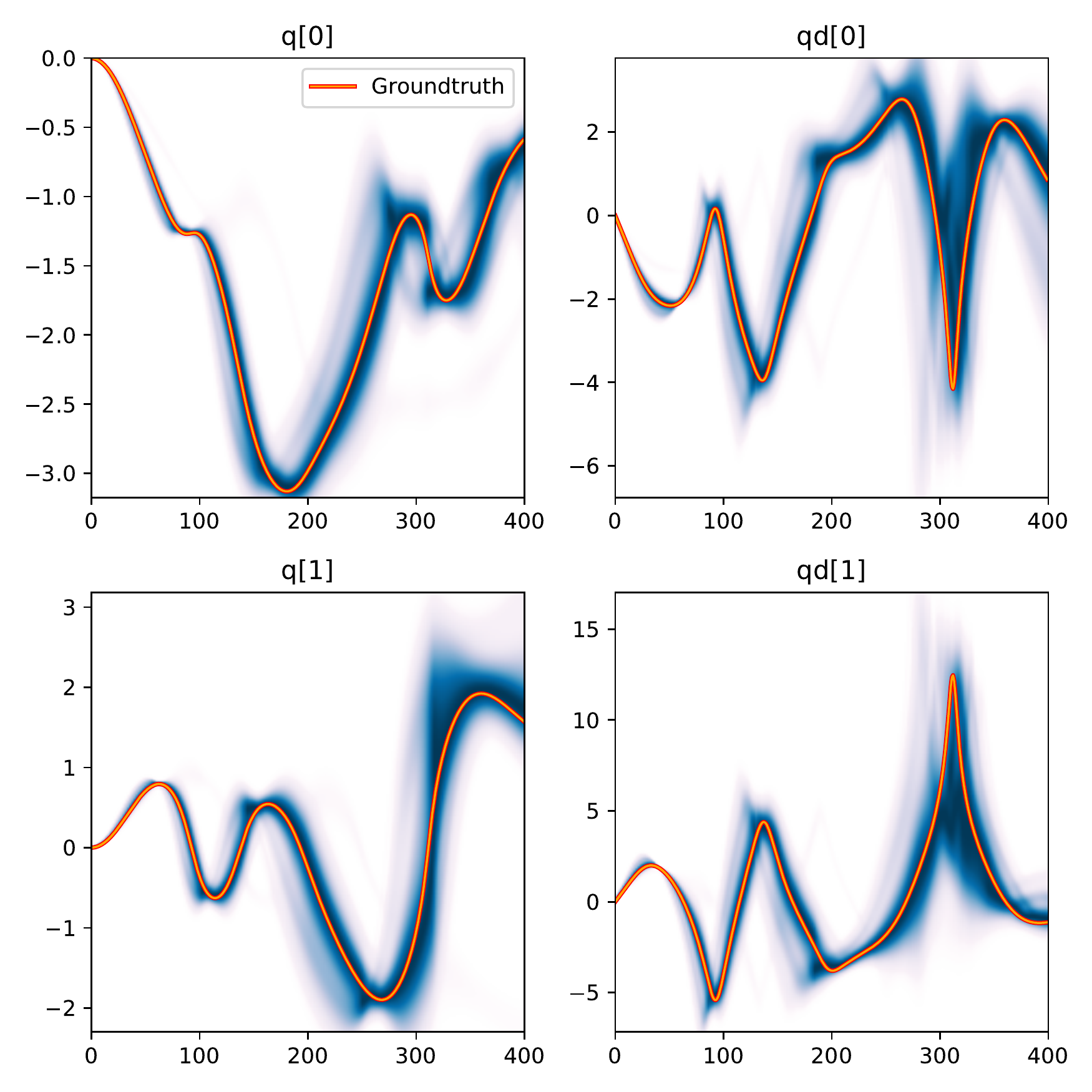}
        \caption{\revtwo{Trajectory density}}
        \label{fig:bayessim-twoparam-trajectories}
    \end{subfigure}
    \caption{\revtwo{Results from BayesSim on the simplified double pendulum experiment where only the two link lengths need to be inferred. (a) shows the approximated posterior distribution by the MDN model and ``downsampled'' input statistics. The diagonal plots show the marginal parameter distributions, the bottom-left heatmap and the top-right contour plot show the 2D posterior where the ground-truth parameters at (\SI{1.5}{\meter}, \SI{2}{\meter}) are indicated by a red star. The black dots in the top-right plot are 100 parameters sampled from the posterior which are rolled out to generate trajectories for the trajectory density plot in (b). (b) shows a kernel density estimation over these 100 trajectories for the four state dimensions $(\jointpos_{[0:1]},\jointvel_{[0:1]})$ of the double pendulum.}}
    \label{fig:bayessim-twoparam}
\end{figure*}

Why do these likelihood-free methods perform so poorly on a seemingly simple double pendulum? One would expect that this kind of dynamical system poses no greater challenge to BayesSim when it was shown to identify a cartpole's link length and cart mass successfully~\cite{ramos_bayessim_2019}. To investigate this problem, we revisit the simplified double pendulum estimation experiment from \autoref{sec:exp-ibm-pendulum} of our main paper, where only the two link lengths need to be estimated from simulated trajectories. As before, we create a dataset with 10,000 trajectories of 400 time steps based on the two simulation parameters sampled from a uniform distribution ranging between \SI{0.5}{\meter} and \SI{5}{\meter}. While keeping all parameters the same as in our previous double-pendulum experiment where eleven parameters had to be inferred, all of the density models in combination with both the ``difference'' and ``downsampled'' input statistic infer a highly accurate parameter distribution, as shown in \autoref{fig:bayessim-twoparam-posterior}. The trajectories produced by sampling from the BayesSim posterior (\autoref{fig:bayessim-twoparam-trajectories}) also match the reference observations much more closely than any of the BayesSim models on the previous 11-parameter double pendulum~(\autoref{fig:ibm-pendulum-rollouts-bayessim}). These results suggest that BayesSim and potentially other likelihood-free method have problems in inferring higher dimensional parameter distributions. The experiments in \cite{ramos_bayessim_2019} demonstrated as many as four parameters being estimated (for the acrobot), while showing inference results for simulated systems only. While our double pendulum system from \autoref{sec:ibm-pendulum} is basic in principle, the higher dimensional parameter space (see parameters in \autoref{tab:params-ibm-pendulum}) and the fact that we need to fit against real-world data makes it a significantly harder problem for most state-of-the-art inference algorithms. CSVGD is able to achieve a close fit thanks to the multiple-shooting segmentation of the trajectory which improves the convergence (see more ablation results for multiple-shooting on this experiment in \autoref{sec:exp-multiple-shooting}).

\subsection{Identify Inertia of an Articulated Rigid Object}
\label{sec:panda-box-appendix}

\begin{figure}[H]
    \centering
    \includegraphics[width=\columnwidth,trim=4cm 8cm 8cm 4cm,clip]{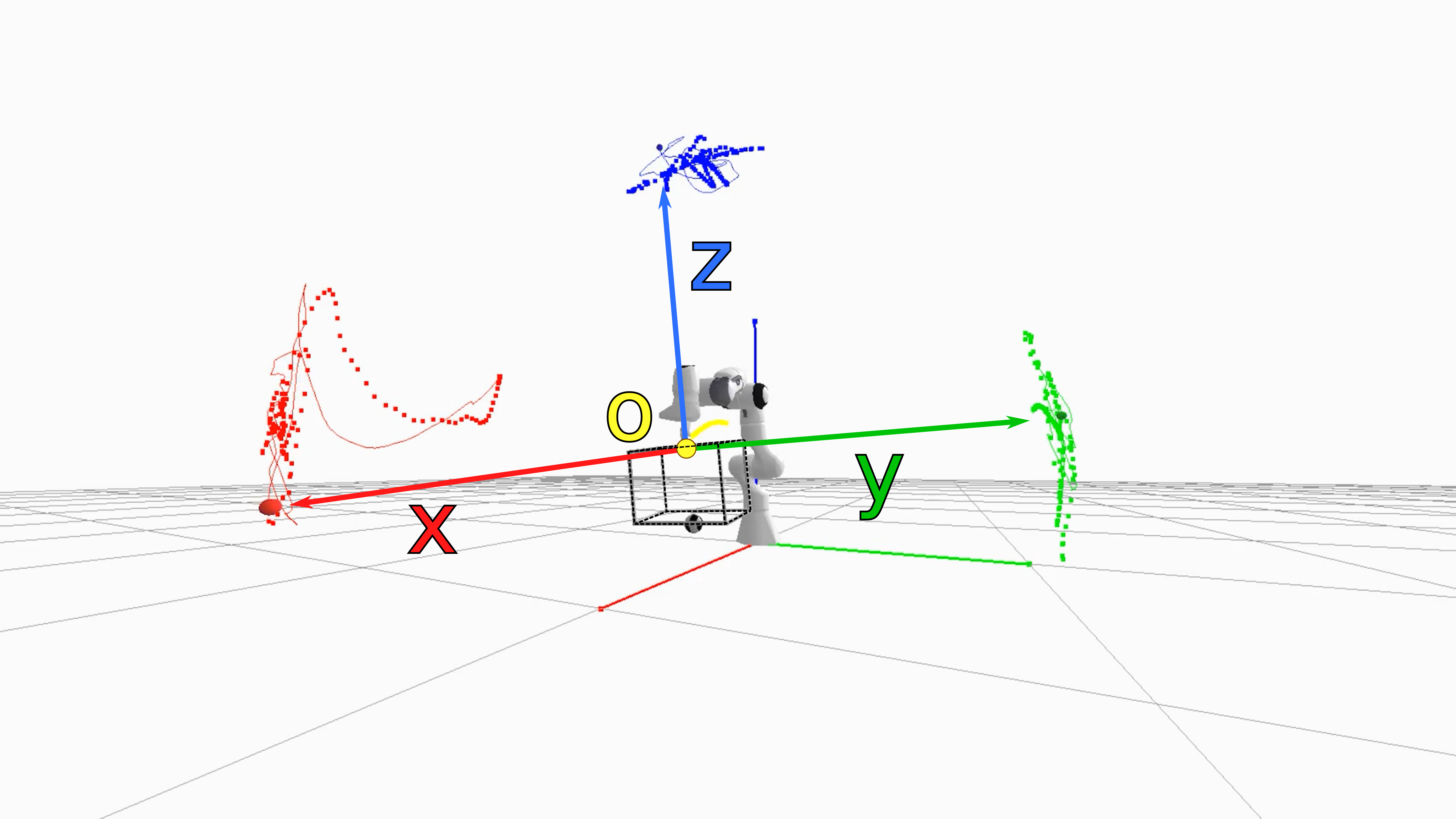}
    \caption{Rendering of the simulation for the system from \autoref{sec:panda-box}, where the four reference points for the origin, unit x, y, and z vectors are shown. The trace of the simulated trajectory is visualized by the solid lines, the ground-truth trajectories of the markers are shown as dotted lines.}
    \label{fig:panda-markers}
\end{figure}

The state space consists of the positions and velocities of the seven degrees of freedom of the robot arm and the two degrees of freedom in the universal joint, resulting in a 20-dimensional state vector $\statevec=\bmat{\mathbf{q}_{0:8} & \mathbf{\dot{q}}_{0:8} & \mathbf{q}^d_{0:6} & \mathbf{\dot{q}}^d_{0:6}}$ consisting of nine joint positions and velocities, plus the PD control position and velocity targets, $\mathbf{q}^d$ and $\mathbf{\dot{q}}^d$, for the actuated joints of the robot arm. We control the arm using the \emph{MoveIt!} motion planning framework~\cite{coleman_david_reducing_2014} by moving joints 6 and 7 to predefined joint-space offsets of $0.1$ and $-0.1$ radians, in sequence. We use the default Iterative Parabolic Time Parameterization algorithm with a velocity scaling factor of $0.1$. We track the motion of the acrylic box via a Vicon motion capture system and derive four Cartesian coordinates as observation $\observationvec=\bmat{\mathbf{p}_{o} & \mathbf{p}_{x} & \mathbf{p}_{y} & \mathbf{p}_{z}}$ to represent the frame of the box (shown in \autoref{fig:panda-markers}): a point of origin located at the center of the upper lid of the box, and three points located \SI{1}{\meter} away from the origin into the x, y, and z direction (transformed by the reference frame of the box). We chose this state representation to ease the computation of the likelihood, since we only need to compute differences between 3D points instead of computing the distances over 3D rotations which requires special treatment~\cite{huynh2009metrics}.

\begin{figure}[t]
    \centering
    \newcommand{\subfigwidth}{\columnwidth}
    \begin{subfigure}[t]{\subfigwidth}
        \centering
        \includegraphics[width=0.8\textwidth,trim=2cm 2cm 2cm 2cm]{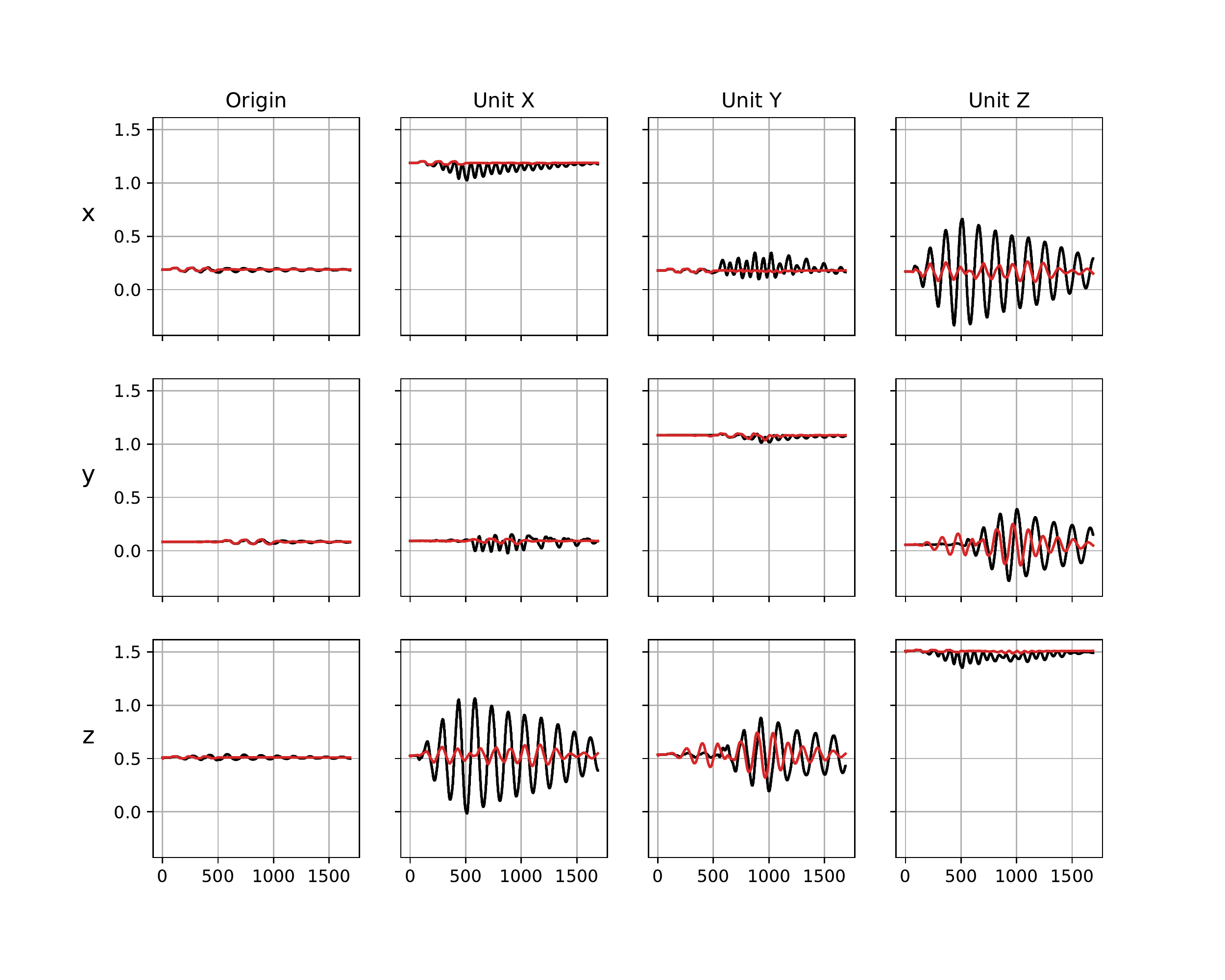}
        \caption{\revtwo{Before identification of empty box}}
        \label{fig:panda-box-trajectory-before}
    \end{subfigure}
    \begin{subfigure}[t]{\subfigwidth}
        \centering \vspace*{1em}
        \includegraphics[width=0.8\textwidth,trim=2cm 2cm 2cm 2cm]{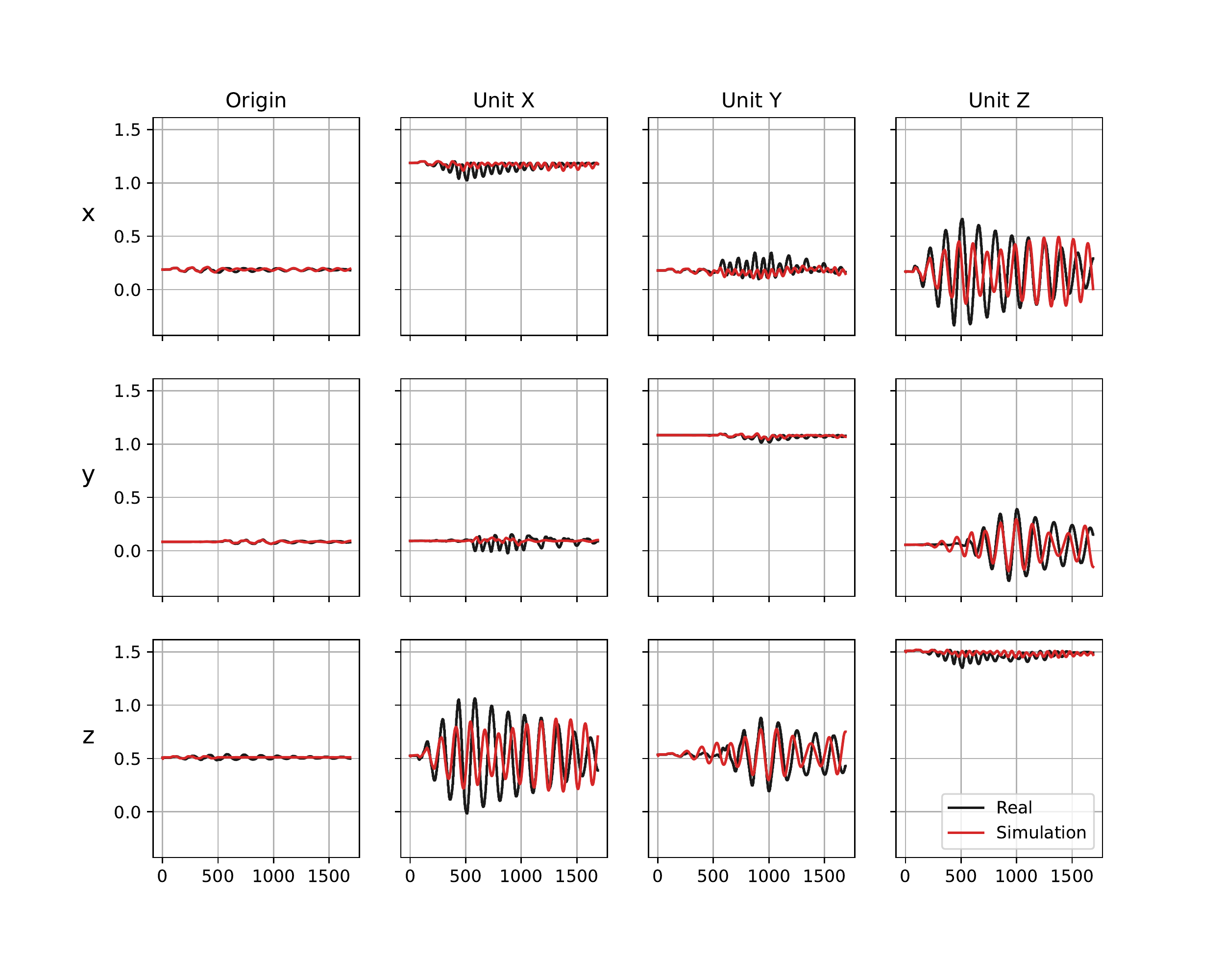}
        \caption{\revtwo{After identification of empty box}}
        \label{fig:panda-box-trajectory-after}
    \end{subfigure}
    \caption{Trajectories from the Panda robot arm shaking an empty box. Visualized are the simulated (red) and real (black) observations \revtwo{before (a) and after (b)} the inertial parameters of the empty box and the friction from the universal joint (\autoref{tab:params-panda-box-phase1}) have been identified. The columns correspond to the four reference points in the frame of the box (see a rendering of them in \autoref{fig:panda-markers}), the rows show the $x$, $y$, and $z$ axes of these reference points in meters. The horizontal axes show the time step.}
    \label{fig:panda-box-trajectory}
\end{figure}

We first identify the simulation parameters pertaining to the inertial properties of the box and the friction parameters of the universal joint. As shown in \autoref{tab:params-panda-box-phase1}, the symmetric inertia matrix of the box is fully determined by the first six parameters, followed by the 3D center of mass. We have measured the mass to be \SI{920}{\gram}, so we do not need to estimate it. We simulate the universal joint with velocity-dependent damping, with possibly different friction coefficients for both degrees of freedom. The simulation parameters yielding the most accurate fit to a ground-truth trajectory from the physical robot shaking an empty box is shown in \autoref{fig:panda-box-trajectory}. We were able to find such parameters via SVGD, CSVGD and Emcee (shown is a parameter configuration from the particle distribution estimated by CSVGD with the highest likelihood).

\revtwo{While the simulated trajectory matches the real data significantly better after the inertial parameters of the empty box have been identified (\autoref{fig:panda-box-trajectory-after}) than before (\autoref{fig:panda-box-trajectory-before}), a reality gap remains. We believe this to be a result from a slight modeling error that the rigid body simulator cannot capture, e.g. the top of the box where the universal joint is attached bends slightly while the box is moving, and there may be slight geometric offsets between the real system and the model of it we use in the simulator. The latter parameters could have been further identified with our approach, nonetheless the simulation given the identified parameters is sufficient to be used in the next phase of the inference experiment.}

Given the parameters found in the first phase, we now investigate how well the various approaches can cope with dependent variables. By fixing two \SI{500}{\gram} to the bottom of the acrylic box, the 2D locations of such weights need to be inferred. Naturally, such assignment is symmetric, i.e. weight 1 and 2 can swap locations without affecting the dynamics. What would significantly alter the dynamics, however, is an unbalanced configuration of the weights which would cause the box to tilt.

\begin{table}[t]
    \centering
    \newcommand{\subfigwidth}{\columnwidth}
    \begin{subfigure}[b]{\subfigwidth}
        \centering
        \resizebox{\subfigwidth}{!}{
            \begin{tabular}{llS[table-format=5.3]@{\,}s[table-unit-alignment = left]S[table-format=5.3]@{\,}s[table-unit-alignment = left]}
                \toprule
                        & \bf Parameter  & \multicolumn{2}{c}{\bf Minimum} & \multicolumn{2}{c}{\bf Maximum}                             \\
                \midrule
                Box
                        & $I_{xx}$       & 0.05                            & \si{\kg.\meter^2}               & 0.1   & \si{\kg.\meter^2} \\
                        & $I_{yy}$       & 0.05                            & \si{\kg.\meter^2}               & 0.1   & \si{\kg.\meter^2} \\
                        & $I_{zz}$       & 0.05                            & \si{\kg.\meter^2}               & 0.1   & \si{\kg.\meter^2} \\
                        & $I_{xy}$       & -0.01                           & \si{\kg.\meter^2}               & 0.01  & \si{\kg.\meter^2} \\
                        & $I_{xz}$       & -0.01                           & \si{\kg.\meter^2}               & 0.01  & \si{\kg.\meter^2} \\
                        & $I_{yz}$       & -0.01                           & \si{\kg.\meter^2}               & 0.01  & \si{\kg.\meter^2} \\
                        & COM $x$        & -0.005                          & \si{\meter}                     & 0.005 & \si{\meter}       \\
                        & COM $y$        & -0.005                          & \si{\meter}                     & 0.005 & \si{\meter}       \\
                        & COM $z$        & 0.1                             & \si{\meter}                     & 0.4   & \si{\meter}       \\
                \midrule\addlinespace[.5em]
                U-Joint & Friction DOF 1 & 0.0                             &                                 & 0.15  &                   \\
                        & Friction DOF 2 & 0.0                             &                                 & 0.15  &                   \\
                \bottomrule                                                                                                              \\
            \end{tabular}
        }\vspace*{-1em}
        \caption{Phase I}
        \label{tab:params-panda-box-phase1}
    \end{subfigure}\\[2em]
    \begin{subfigure}[b]{\subfigwidth}
        \centering
        \resizebox{\subfigwidth}{!}{
            \begin{tabular}{llS[table-format=5.3]@{\,}s[table-unit-alignment = left]S[table-format=5.3]@{\,}s[table-unit-alignment = left]}
                \toprule
                 & \bf Parameter & \multicolumn{2}{c}{\bf Minimum} & \multicolumn{2}{c}{\bf Maximum}                      \\
                \midrule
                Weight 1   & Position $x$  & -0.14                           & \si{\meter}                     & 0.14 & \si{\meter} \\
                           & Position $y$  & -0.08                           & \si{\meter}                     & 0.08 & \si{\meter} \\
                \midrule\addlinespace[.5em]
                Weight 2   & Position $x$  & -0.14                           & \si{\meter}                     & 0.14 & \si{\meter} \\
                           & Position $y$  & -0.08                           & \si{\meter}                     & 0.08 & \si{\meter} \\
                \bottomrule                                                                                                         \\
            \end{tabular}
        }\vspace*{-1em}
        \caption{Phase II}
        \label{tab:params-panda-box-phase2}
    \end{subfigure}

    \caption{Parameters to be estimated and their ranges for the two estimation phases of the underactuated mechanism experiment from \autoref{sec:panda-box}.}
    \label{tab:params-panda-box}
\end{table}

\begin{figure*}[t]
    \centering
    \newcommand{\figheight}{4.2cm}
    \newcommand{\subfigwidth}{.45\textwidth}
    \begin{subfigure}[b]{\subfigwidth}
        \centering
        \includegraphics[height=\figheight]{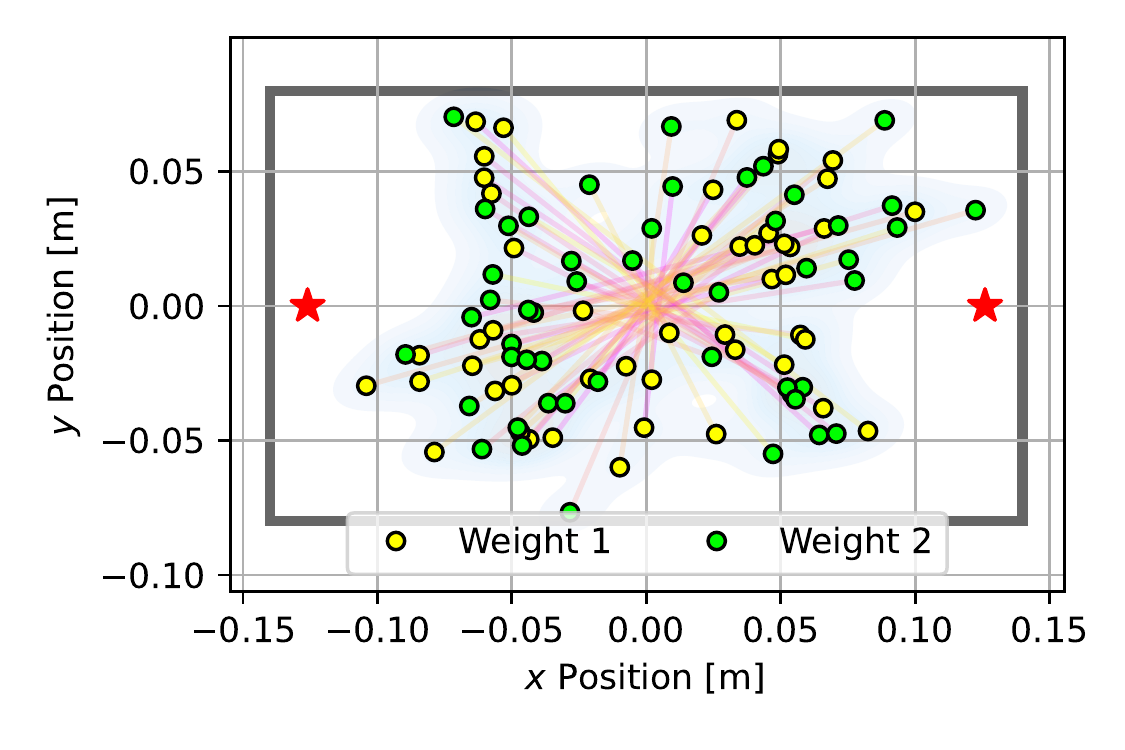}
        \caption{Emcee}
        \label{fig:panda-box-posterior-emcee}
    \end{subfigure}
    \begin{subfigure}[b]{\subfigwidth}
        \centering
        \includegraphics[height=\figheight]{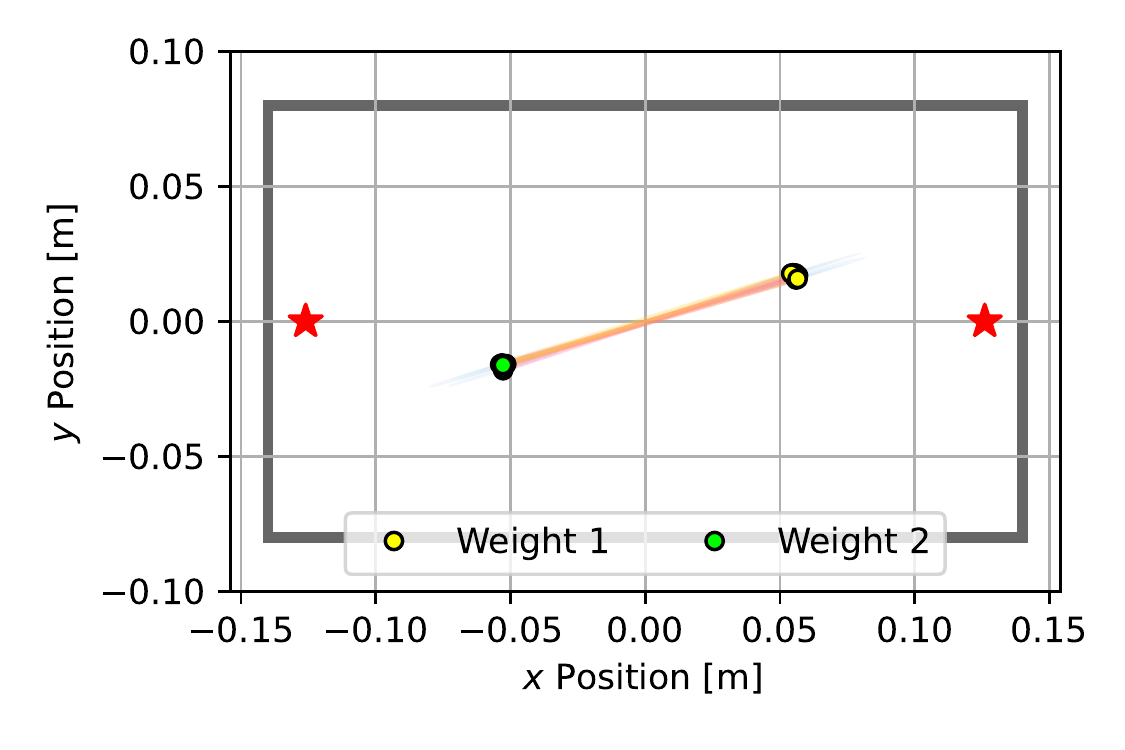}
        \caption{CEM}
        \label{fig:panda-box-posterior-cem}
    \end{subfigure}
    \begin{subfigure}[b]{\subfigwidth}
        \centering
        \includegraphics[height=\figheight]{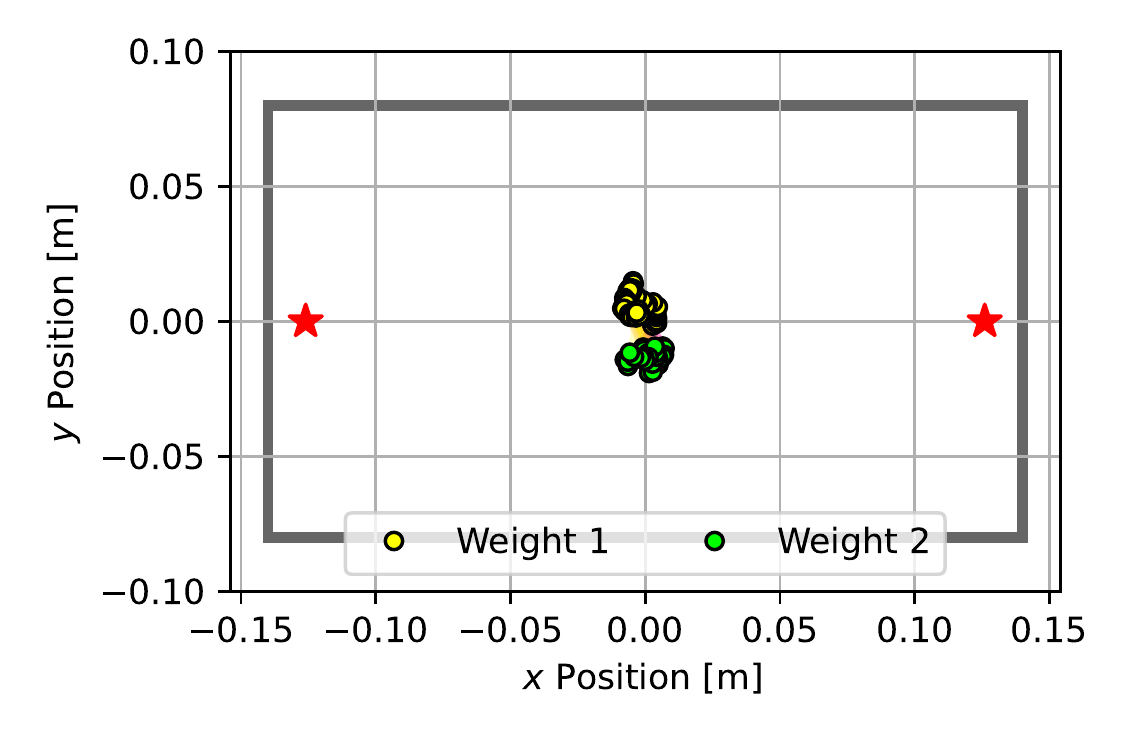}
        \caption{SGLD}
        \label{fig:panda-box-posterior-sgld}
    \end{subfigure}
    \begin{subfigure}[b]{\subfigwidth}
        \centering
        \includegraphics[height=\figheight]{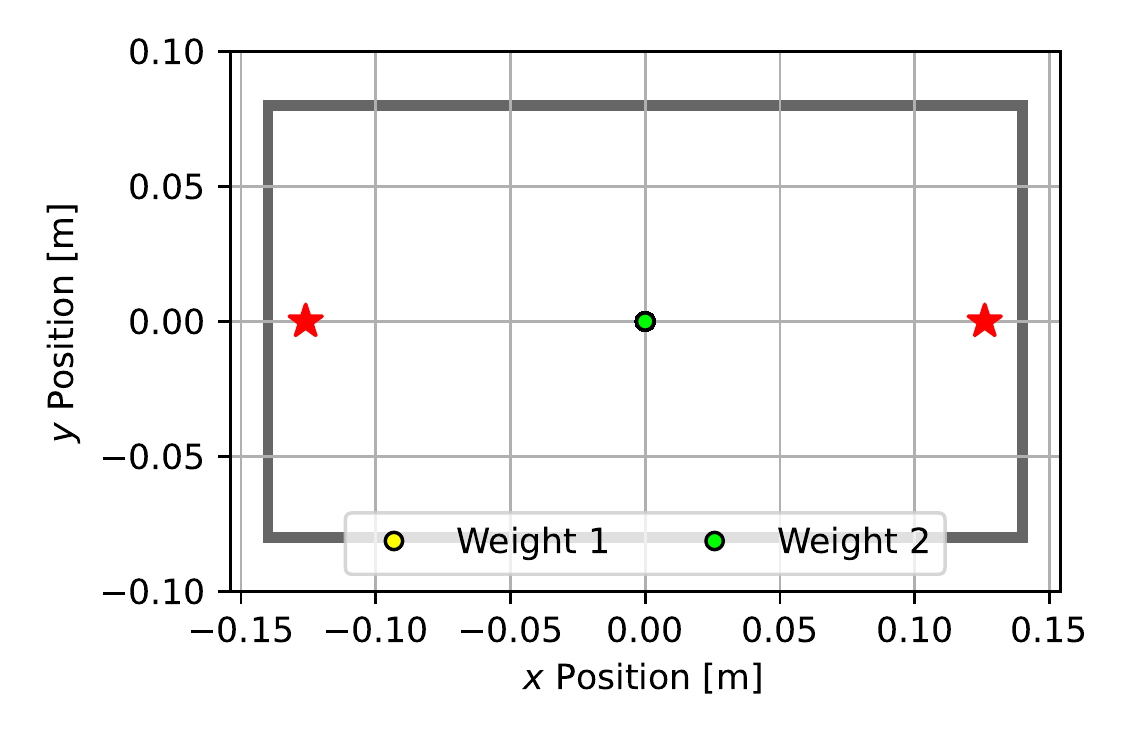}
        \caption{NUTS}
        \label{fig:panda-box-posterior-nuts}
    \end{subfigure}
    \begin{subfigure}[b]{\subfigwidth}
        \centering
        \includegraphics[height=\figheight]{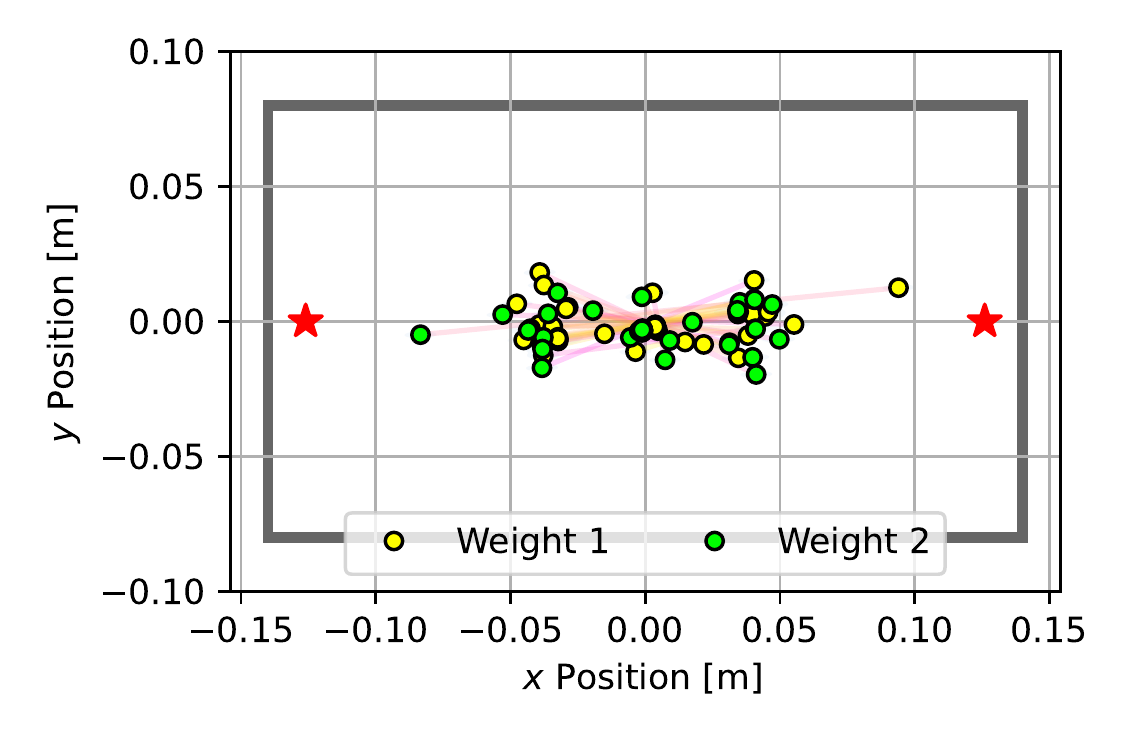}
        \caption{SVGD}
        \label{fig:panda-box-posterior-svgd}
    \end{subfigure}
    \begin{subfigure}[b]{\subfigwidth}
        \centering
        \includegraphics[height=\figheight]{fig/panda/box_position_svgd_ms.pdf}
        \caption{CSVGD}
        \label{fig:panda-box-posterior-csvgd}
    \end{subfigure}
    \caption{Posterior plots over the 2D weight locations approximated by the estimation algorithms applied to the underactuated mechanism experiment from \autoref{sec:panda-box}. Blue shades indicate a Gaussian kernel density estimation computed over the inferred parameter samples. \rev{Since it is an unbounded kernel density estimate, the blue shades cross the parameter boundaries in certain areas (e.g. for CSVGD), while in reality none of the estimated particles violate the parameter limits.}}
    \label{fig:panda-box-posterior}
\end{figure*}

We use 50 particles and run each estimation algorithm for 500 iterations. For each baseline method, we carefully tuned the hyper parameters to facilitate a fair comparison. Such tuning included selecting an appropriate measurement noise variance, which, as we observed on Emcee and SGLD in particular, had a significant influence on the exploration behavior of these algorithms. With a larger observation noise variance the resulting posterior distribution became wider, however we were unable to attain such behavior with the NUTS estimator whose iterates quickly collapsed to a single point at the center of the box (see \autoref{fig:panda-box-posterior-nuts}). Similarly, CEM immediately became stuck in the suboptimal configuration shown in \autoref{fig:panda-box-posterior-cem}. Nonetheless, after 500 iterations, all methods predicted weight positions that were aligned opposite to one another to balance the box.

As can be seen in \autoref{fig:panda-box-posterior-svgd}, SVGD achieves a fairly predictive posterior approximation, with many particles aligned close to the true vertical position at $y=0$. With the introduction of the multiple-shooting constraints, Constrained SVGD (CSVGD) converges significantly faster to a posterior distribution that accurately captures the true locations of the box, while retaining the exploration performance of SVGD that spreads out the particles over multiple modes, as shown in \autoref{fig:panda-box-posterior-csvgd}.



\newpage

\printbibliography
\end{refsection}

\end{document}